\documentclass{article}

\usepackage{microtype}
\usepackage{graphicx}
\usepackage{subfigure}
\usepackage{booktabs} % for professional tables

\usepackage{hyperref}

\usepackage[accepted]{icml2024}

\usepackage{amsmath}
\usepackage{amssymb}
\usepackage{amsfonts}
\usepackage{mathtools}
\usepackage{amsthm}
\usepackage{float}

\usepackage{multirow}
\usepackage{booktabs}
\usepackage{xcolor}
\usepackage{enumitem}

\usepackage[capitalize,noabbrev]{cleveref}

\theoremstyle{plain}
\newtheorem{theorem}{Theorem}[section]

\newtheorem{corollary}[theorem]{Corollary}
\theoremstyle{definition}
\newtheorem{definition}[theorem]{Definition}
\newtheorem{assumption}[theorem]{Assumption}
\theoremstyle{remark}

\usepackage{macros}

\icmltitlerunning{Convex and Bilevel Optimization for Neural-Symbolic Inference and Learning}

\begin{document}

\twocolumn[
\icmltitle{Convex and Bilevel Optimization for\\
Neural-Symbolic Inference and Learning}

\begin{icmlauthorlist}
\icmlauthor{Charles Dickens}{ucsc}
\icmlauthor{Changyu Gao}{uwisc}
\icmlauthor{Connor Pryor}{ucsc}
\icmlauthor{Stephen Wright}{uwisc}
\icmlauthor{Lise Getoor}{ucsc}
\end{icmlauthorlist}

\icmlaffiliation{ucsc}{Department of Computer Science and Engineering, University of California, Santa Cruz, CA 95060}
\icmlaffiliation{uwisc}{Department of Computer Sciences, University of Wisconsin-Madison, Madison, WI 53706}

\icmlcorrespondingauthor{Charles Dickens}{cadicken@ucsc.edu}

\icmlkeywords{Neural-Symbolic AI, Optimization, Machine-Learning}

\vskip 0.3in
]

\printAffiliationsAndNotice{} 

\begin{abstract}
We leverage convex and bilevel optimization techniques to develop a general gradient-based parameter learning framework for neural-symbolic (NeSy) systems.
We demonstrate our framework with NeuPSL, a state-of-the-art NeSy architecture.
To achieve this, we propose a smooth primal and dual formulation of NeuPSL inference and show learning gradients are functions of the optimal dual variables.
Additionally, we develop a dual block coordinate descent algorithm for the new formulation that naturally exploits warm-starts. 
This leads to over $100 \times$ learning runtime improvements over the current best NeuPSL inference method.
Finally, we provide extensive empirical evaluations across $8$ datasets covering a range of tasks and demonstrate our learning framework achieves up to a $16\%$ point prediction performance improvement over alternative learning methods.
\end{abstract}

\commentout{
\begin{abstract}
We address a key challenge for neural-symbolic (NeSy) systems by leveraging convex and bilevel optimization techniques to develop a general 
% first-order 
gradient-based framework for end-to-end neural and symbolic parameter learning.
% Specifically, we formulate NeSy learning as a bilevel program, and employ Moreau smoothing and a graduated value-function approach to support learning with a constrained lower-level inference problem.
The applicability of our framework is demonstrated with NeuPSL, a state-of-the-art NeSy architecture.
To achieve this, we propose a smooth primal and dual formulation of NeuPSL inference
% as a convex linearly constrained quadratic program 
and show learning gradients are functions of the optimal dual variables.
Additionally, we develop a dual block coordinate descent algorithm for the new formulation that naturally exploits warm-starts. 
This leads to over $100 \times$ learning runtime improvements over the current best NeuPSL inference method.
Finally, we provide extensive empirical evaluations across $8$ datasets covering a range of tasks and demonstrate our learning framework achieves up to a $16\%$ point prediction performance improvement over alternative learning methods.
\end{abstract}
}

\section{Introduction}
\label{sec:intro}
The goal of neural-symbolic (NeSy) AI is a seamless integration of neural models for processing low-level data with symbolic frameworks to reason over high-level symbolic structures \citep{garcez:book02, garcez:book09, garcez:jal19}.
In this paper, we introduce a principled and general NeSy learning framework that supports end-to-end neural and symbolic parameter learning.
Further, we propose a novel inference algorithm and establish theoretical properties for a state-of-the-art NeSy system that are crucial for learning.

Our proposed learning framework builds upon NeSy energy-based models (NeSy-EBMs) \citep{pryor:ijcai23}, a general class of NeSy systems that encompasses a variety of existing NeSy methods, including DeepProblog \citep{manhaeve:neurips18, manhaeve:ai21}, SATNet \citep{wang:icml19}, logic tensor networks \citep{badreddine:ai22}, and NeuPSL \citep{pryor:ijcai23}.
NeSy-EBMs use neural network outputs to parameterize an energy function and formulate an inference problem that may be non-smooth and constrained.
Thus, predictions are not guaranteed to be a function of the inputs and parameters with an explicit form or to be differentiable, and traditional deep learning techniques are not directly applicable.
Therefore, we equivalently formulate NeSy-EBM learning as a bilevel problem.
Moreover, to support first-order gradient-based optimization, we propose a smoothing strategy that is novel to NeSy learning.
Specifically, we replace the constrained NeSy energy function with its Moreau envelope.
The augmented Lagrangian method for equality-constrained minimization is then applied with the new formulation. 

We demonstrate the effectiveness of our proposed learning framework with NeuPSL.
To ensure differentiability and provide principled forms of gradients for learning, we present a new formulation and regularization of NeuPSL inference as a quadratic program.
Moreover, we introduce a dual block coordinate descent (dual BCD) inference algorithm for the quadratic program.
The dual BCD algorithm is the first NeuPSL inference method that produces optimal dual variables for producing both optimal primal variables and gradients for learning.
Additionally, empirical results demonstrate that dual BCD is able to leverage warm starts effectively, thus improving learning runtime.

Our key contributions are: 
(1) A bilevel formulation of the NeSy-EBM learning problem that establishes a foundation for applying smooth first-order gradient-based optimization techniques;
(2) A reformulation of NeuPSL inference that is used to prove continuity properties and obtain explicit forms of gradients for learning;
(3) A dual BCD algorithm for NeuPSL inference that naturally produces statistics necessary for computing gradients for learning and that fully leverages warm-starts to improve learning runtime;
(4) Two parallelization strategies for dual BCD inference; and
(5) A thorough empirical evaluation demonstrating prediction performance improvements on $8$ different datasets and a learning runtime speedup of up to $100 \times$.

\section{Related Work}
NeSy AI is an active area of research that incorporates symbolic (commonly logical and arithmetic) reasoning with neural networks 
% CAD: Longer version
% \citep{garcez:book02,bader:wwst05,garcez:book09,serafini:aiia16,besold:arxiv17,donadello:ijcai17,yang:neurips17,evans:jair18,manhaeve:ai21,garcez:jal19,deraedt:ijcai20,lamb:ijcai20,badreddine:ai22}.
\citep{bader:wwst05, garcez:book09, besold:arxiv17, deraedt:ijcai20, lamb:ijcai20, giunchiglia:ijcai22}.
% For a thorough introduction to NeSy, we recommend the surveys by \citenoun{besold:arxiv17} and \citenoun{deraedt:ijcai20}.
We will show that learning for a general class of NeSy systems is naturally formulated as bilevel optimization \citep{bracken:or71, colson:aor07, bard:book13}.
% CAD: Longer version
% \citep{bracken:or71, colson:aor07, bard:book13, dempe:book20}.
In other words, the NeSy learning objective is a function of predictions obtained by solving a lower-level inference problem that is symbolic reasoning.
In this work, we focus on a general setting where the lower-level problem is an expressive and complex program capable of representing cyclic dependencies and ensuring the satisfaction of constraints during both learning and inference \citep{wang:icml19, badreddine:ai22, dasartha:ijcai23, pryor:ijcai23, cornelio:iclr23}.
One prominent and tangential subgroup of such NeSy systems we would like to acknowledge enforces constraints on the structure of the symbolic model, and hence the lower-level problem, to ensure the final prediction has an explicit gradient with respect to the parameters \citep{xu:icml18, manhaeve:ai21, ahmed:neurips22}.
% CAD: Longer version
% In the deep learning community, bilevel optimization notably also arises in hyperparameter optimization \citep{pedregosa:icml16}, meta-learning \citep{franceschi:icml18, rajeswaran:neurips19}, generative adversarial networks \citep{goodfellow:neurips14}, and reinforcement learning \citep{sutton:book18}. 
In the deep learning community, bilevel optimization also arises in hyperparameter optimization and meta-learning \citep{pedregosa:icml16, franceschi:icml18}, generative adversarial networks \citep{goodfellow:neurips14}, and reinforcement learning \citep{sutton:book18}.

Researchers typically take one of three approaches to bilevel optimization:
(1) \emph{Implicit differentiation} methods compute or approximate the Hessian matrix at the lower-level problem solution to derive an analytic expression for the gradient of the upper-level objective called a hypergradient \citep{do:neurips07, pedregosa:icml16, ghadimi:arxiv18, rajeswaran:neurips19, giovannelli:arxiv22, khanduri:icml23}.
(2) \emph{Automatic differentiation} methods unroll inference into a differentiable computational graph \citep{stoyanov:aistats11, domke:aistats12, belanger:icml17, ji:icml21}.
(3) \emph{Value-Function approaches} reformulate the bilevel problem as a single-level constrained program using the optimal value of the lower-level objective (the \emph{value-function}) to develop principled first-order gradient-based algorithms that do not require the calculation of Hessian matrices for the lower-level problem 
% CAD: Longer version
% \citep{outrata:zor90, ye:opt95, liu:icml21, sow:arxiv22, liu:neurips22, liu:arxiv23, kwon:icml23}.
\citep{outrata:zor90, ye:opt95, liu:icml21, sow:arxiv22, liu:neurips22, liu:arxiv23, kwon:icml23}.

Note that standard algorithms for all three approaches to bilevel optimization suggest solving the lower-level problem to derive gradients.
Principled techniques for using approximate lower-level solutions to make progress on the bilevel program is an open research direction \citep{pedregosa:icml16, liu:icml21}.
Further, the lower-level problem for NeSy learning (inference) is commonly constrained.  
Current approaches to working with lower-level constraints are based on implicit differentiation \citep{giovannelli:arxiv22, khanduri:icml23}.
In this work, we introduce a value-function approach for bilevel optimization with lower-level constraints.

% CAD: (09/25/23)
\commentout{
% cp - So, in general, the related work looks great. If you are strapped for space, I think you can remove implicit differentiation and automatic differentiation and discuss bilevel value function approaches. When doing that, throw in one sentence for each acknowledging their existence.
NeSy AI is an active area of research that incorporates symbolic (commonly logical and arithmetic) reasoning with neural networks 
% CAD: Longer version
% \citep{garcez:book02,bader:wwst05,garcez:book09,serafini:aiia16,besold:arxiv17,donadello:ijcai17,yang:neurips17,evans:jair18,manhaeve:ai21,garcez:jal19,deraedt:ijcai20,lamb:ijcai20,badreddine:ai22}.
\citep{bader:wwst05,garcez:book09,besold:arxiv17,yang:neurips17,manhaeve:ai21,deraedt:ijcai20,lamb:ijcai20,badreddine:ai22}.
% For a thorough introduction to NeSy, we recommend the surveys by \citenoun{besold:arxiv17} and \citenoun{deraedt:ijcai20}.
% cp - ss: Do you need this sentence? You just stated this in the introduction.
In this work, we develop a learning framework for NeSy-EBMs, and demonstrate its applicability and performance with NeuPSL~\cite{pryor:ijcai23}.
NeSy-EBM learning objectives are functions of an $argmin$ operation representing inference and is naturally formulated as bilevel optimization \citep{bracken:or71, colson:aor07, bard:book13}.
% CAD: Longer version
% \citep{bracken:or71, colson:aor07, bard:book13, dempe:book20}.
% CAD: Longer version
% In the deep learning community, bilevel optimization notably also arises in hyperparameter optimization \citep{pedregosa:icml16}, meta-learning \citep{franceschi:icml18, rajeswaran:neurips19}, generative adversarial networks \citep{goodfellow:neurips14}, and reinforcement learning \citep{sutton:book18}. 
In the deep learning community, bilevel optimization notably also arises in hyperparameter optimization and meta-learning \citep{pedregosa:icml16, franceschi:icml18}, generative adversarial networks \citep{goodfellow:neurips14}, and reinforcement learning \citep{sutton:book18}. 
Researchers typically take one of the following three approaches to bilevel optimization.

\textbf{Implicit Differentiation} 
There is a long history of research on analyzing the stability of solutions to optimization problems using implicit differentiation \citep{fiacco:book68, robinson:mor80, bonnans:book00}.
Implicit differentiation methods compute or approximate the Hessian matrix of the lower-level problem to derive a so-called hypergradient.
Bilevel algorithms in this category have recently been proposed and make varying assumptions about the problem structure \cite{do:neurips07, pedregosa:icml16, ghadimi:arxiv18, rajeswaran:neurips19, giovannelli:arxiv22, khanduri:icml23}.
% cp - ss: This is the other sentence we can maybe leave out? I don't know how fundamental is the discussion around the deep learning community.
Related to this approach, the deep learning community has developed layers that are functions of convex $argmin$ operations with analytic expressions for gradients derived from implicit differentiation \citep{amos:icml17, agrawal:neurips19, agrawal:jano19, wang:icml19}. 

\textbf{Automatic Differentiation}
Another direction unrolls inference into a differentiable computational graph \citep{stoyanov:aistats11, domke:aistats12, belanger:icml17, ji:icml21}.
This approach leverages widely applied automatic differentiation techniques \citep{griewank:book08}.
However, unrolling the inference computation creates a large, complex computational graph that can accumulate numerical errors dependent on the solver.
% CAD (5/15): Potential cut
% Moreover, some operations of the optimization algorithms for inference may be non-smooth or discontinuous and difficult to unroll.

\textbf{Bilevel Value-Function Approach}
Finally, an increasingly popular approach is to reformulate the problem as single-level constrained program using the optimal value-function of the $argmin$ objective \citep{outrata:zor90, ye:opt95, liu:icml21, sow:arxiv22, liu:neurips22, liu:arxiv23, kwon:icml23}.
Value-function approaches yield practical first-order gradient-based algorithms that do not require calculation of the lower-level Hessian.

Existing value-function approaches are not directly applicable to NeSy-EBMs as they typically assume the lower-level problem is unconstrained and the objective is continuously differentiable.
More generally, bilevel optimization with lower-level constraints is an open area of research and, until now, implicit differentiation methods are applied with strong assumptions about the structure of the lower-level problem \citep{giovannelli:arxiv22, khanduri:icml23}.
Our framework is, to the best of our knowledge, the first value-function approach to work with lower-level problem constraints.
}

\section{NeSy Energy-Based Models}
\label{sec:nesy-ebms}
In this work, we use \emph{NeSy energy-based models (NeSy-EBMs)} \citep{pryor:ijcai23} to develop a generally applicable NeSy learning framework.
Here, we provide background on NeSy-EBMs and introduce a classification of losses that motivates the need for general learning algorithms.\footnote{See \appref{appendix:notation} for a table of notation.}

NeSy-EBMs are a family of EBMs \citep{lecun:book06} that use neural model predictions to define potential functions with symbolic interpretations.
NeSy-EBM energy functions are parameterized by a set of neural and symbolic weights from the domains $\mathcal{W}_{nn}$ and $\mathcal{W}_{sy}$, respectively, and quantify the compatibility of a target variable from a domain $\mathcal{Y}$ and neural and symbolic inputs from the domains $\mathcal{X}_{nn}$ and $\mathcal{X}_{sy}$: $E: \mathcal{Y} \times \mathcal{X}_{sy} \times \mathcal{X}_{nn} \times \mathcal{W}_{sy} \times \mathcal{W}_{nn} \to \mathbb{R}$.
NeSy-EBM inference requires first computing the output of the neural networks, \emph{neural inference}, and then minimizing the energy function over the targets, \emph{symbolic inference}: 
{
% \small
\begin{align}
\argmin_{\mathbf{y} \in \mathcal{Y}} E(\mathbf{y}, \mathbf{x}_{sy}, \mathbf{x}_{nn}, \mathbf{w}_{sy}, \mathbf{w}_{nn}).
\end{align}
}%
NeSy-EBM learning is finding weights to create an energy function that associates lower energies to target values near their truth in a set of training data.
The training data consists of $P$ samples that are tuples of symbolic variables and neural network inputs: $\mathcal{S} := \{ S_{1} := (\mathbf{y}_{1}, \mathbf{x}_{1, sy}, \mathbf{x}_{1, nn}), \cdots, S_{P} := (\mathbf{y}_{P}, \mathbf{x}_{P, sy}, \mathbf{x}_{P, nn})\} $.
Moreover, targets $\mathbf{y}_{i}$ from a training sample $S_{i}$ are partitioned into \emph{labeled variables}, $\mathbf{t}_{i}$ from a domain $\mathcal{T}_{\mathcal{Y}}$, value, and \emph{latent variables}, $\mathbf{z}_{i}$ from a domain $\mathcal{Z}$.
Without loss of generality, we write $\mathbf{y}_{i} = (\mathbf{t}_{i}, \mathbf{z}_{i})$.
NeSy-EBM learning losses are defined using the \emph{latent minimizer} and \emph{full minimizers}, 
{
% \small
\begin{align}
    \mathbf{z}_{i}^{*} & \in \argmin_{\mathbf{z} \in \mathcal{Z}}E((\mathbf{t}_{i}, \mathbf{z}), \mathbf{x}_{i, sy}, \mathbf{x}_{i, nn}, \mathbf{w}_{sy}, \mathbf{w}_{nn}) \\
    \mathbf{y}_{i}^{*} & \in \argmin_{\mathbf{y} \in \mathcal{Y}} E(\mathbf{y}, \mathbf{x}_{i, sy}, \mathbf{x}_{i, nn}, \mathbf{w}_{sy}, \mathbf{w}_{nn}),
\end{align}
}%
and the \emph{latent} and \emph{full optimal value-functions}:
{
% \small
\begin{align}
    V_{\mathbf{z}^{*}_{i}}(\mathbf{w}_{sy}, \mathbf{w}_{nn}) &:= E((\mathbf{t}_{i}, \mathbf{z}^{*}_{i}), \mathbf{x}_{i, sy}, \mathbf{x}_{i, nn}, \mathbf{w}_{sy}, \mathbf{w}_{nn}), \\
    V_{\mathbf{y}^{*}_{i}}(\mathbf{w}_{sy}, \mathbf{w}_{nn}) &:= E(\mathbf{y}^{*}_{i}, \mathbf{x}_{i, sy}, \mathbf{x}_{i, nn}, \mathbf{w}_{sy}, \mathbf{w}_{nn}).
\end{align}
}%
Note the optimal values-functions are functions of the parameters, inputs, and symbolic variables; however, to simplify notation, we only write the parameters as arguments.

\emph{Value-based} learning losses depend on the model weights strictly via the optimal value-functions.
Two common value-based losses for NeSy-EBMs are the latent optimal value-function (\emph{energy loss}), and the difference between the latent and full optimal value-functions (\emph{structured perceptron loss}) \citep{lecun:ieee98, collins:emnlp02}:
{
% \small
\begin{align}
    & L_{Energy}(E(\cdot, \cdot, \cdot, \mathbf{w}_{sy}, \mathbf{w}_{nn}), S_{i}) \\
    & \quad \quad :=  V_{\mathbf{z}^{*}_{i}}(\mathbf{w}_{sy}, \mathbf{w}_{nn}), \nonumber \\
    & L_{SP}(E(\cdot, \cdot, \cdot, \mathbf{w}_{sy}, \mathbf{w}_{nn}), S_{i}) \\ 
    & \quad \quad := V_{\mathbf{z}^{*}_{i}}(\mathbf{w}_{sy}, \mathbf{w}_{nn}) - V_{\mathbf{y}^{*}_{i}}(\mathbf{w}_{sy}, \mathbf{w}_{nn}). \nonumber
\end{align}
}%
A principled first-order gradient-based method for optimizing a value-based objective only requires differentiability of the value-functions.
However, performance metrics are not always aligned with value-based losses.
Moreover, they are known to have degenerate solutions, e.g., weights minimizing the loss but producing a collapsed energy function \citep{lecun:book06, pryor:ijcai23}.

Alternatively, \emph{minimizer-based} learning losses assume the minimizer of the energy function is unique.
With this assumption, energy minimization is a vector-valued function from the weight space $\mathcal{W}_{sy} \times \mathcal{W}_{nn}$ to the target space $\mathcal{Y}$, $\mathbf{y}_{i}^{*}(\mathbf{w}_{sy}, \mathbf{w}_{nn}): \mathcal{W}_{sy} \times \mathcal{W}_{nn} \to \mathcal{Y}$.
Then, minimizer-based losses are compositions of a differentiable supervised loss $d: \mathcal{Y} \times \mathcal{Y} \to \mathbb{R}$, and the minimizer:
{
\small
\begin{align}
    L_{d}(E(\cdot, \cdot, \cdot, \mathbf{w}_{sy}, \mathbf{w}_{nn}), S_{i}) := d(\mathbf{y}_{i}^{*}(\mathbf{w}_{sy}, \mathbf{w}_{nn}), \mathbf{t}_{i}).
\end{align}
}

Minimizer-based losses are general and allow learning with objectives aligned with evaluation metrics.
However, a direct application of a first-order gradient based method for minimizer-based learning requires the Jacobian at the minimizer.
NeSy-EBM predictions are not necessarily differentiable. 
Even if they are differentiable, the computation of the Jacobian is often too expensive to be practical.

\section{A Bilevel NeSy Learning Framework}
\label{sec:learning_framework}

In this section, we introduce a novel and general framework for gradient-based NeSy learning.
Our framework begins with the following formulation of learning as a bilevel optimization problem with an objective that is a combination of minimizer and value-based losses, denoted by $d$ and $L_{Val}$, respectively:
{
% \small
\begin{align}
    & \argmin_{\substack{(\mathbf{w}_{sy}, \mathbf{w}_{nn}) \\ (\mathbf{y}_{1}, \cdots , \mathbf{y}_{P})}} \sum_{i = 1}^{P} \left ( d(\mathbf{y}_{i}, \mathbf{t}_{i}) + L_{Val}(E(\cdot, \cdot, \cdot, \mathbf{w}_{sy}, \mathbf{w}_{nn}), S_{i}) \right ) 
    % + \mathcal{R}(\mathbf{w}_{sy}, \mathbf{w}_{nn}) 
    \nonumber \\
    & \quad \textrm{s.t.} \quad \mathbf{y}_{i} \in \argmin_{\mathbf{y} \in \mathcal{Y}} E(\mathbf{y}, \mathbf{x}_{i, sy}, \mathbf{x}_{i, nn}, \mathbf{w}_{sy}, \mathbf{w}_{nn}), \nonumber \\
    & \quad \quad \quad \quad \quad \forall i \in \{1, \cdots, P\} 
    , \label{eq:bilevel_learning}
\end{align}
}%
A regularizer, denoted by $\mathcal{R}: \mathcal{W}_{sy} \times \mathcal{W}_{nn} \to \mathbb{R}$ is typically added to the objective but is omitted in this discussion to simplify notation.
We make the following (standard) lower-level singleton assumption.
\begin{assumption}
    \label{assumption:lower_level_singleton}
    % $E$ is minimized over the target variables, $\mathbf{y} \in \mathcal{Y}$, at a single point for every setting of the weights $(\mathbf{w}_{sy}, \mathbf{w}_{nn}) \in \mathcal{W}_{sy} \times \mathcal{W}_{nn}$. 
    $E$ is minimized over $\mathbf{y} \in \mathcal{Y}$ at a single point for every $(\mathbf{w}_{sy}, \mathbf{w}_{nn}) \in \mathcal{W}_{sy} \times \mathcal{W}_{nn}$. 
\end{assumption}
Under \assumptionref{assumption:lower_level_singleton}, and regardless of the continuity and curvature properties of the upper and lower level objectives, \eqref{eq:bilevel_learning} is equivalent to the following:
{
% \small
\begin{align}
    & \argmin_{\substack{(\mathbf{w}_{sy}, \mathbf{w}_{nn}) \\ (\mathbf{y}_{1}, \cdots , \mathbf{y}_{P})}} \sum_{i = 1}^{P} \left ( d(\mathbf{y}_{i}, \mathbf{t}_{i}) + L_{Val}(E(\cdot, \cdot, \cdot, \mathbf{w}_{sy}, \mathbf{w}_{nn}), S_{i}) \right ) \nonumber \\
    & \quad \quad \textrm{s.t.} \quad \quad E(\mathbf{y}_{i}, \mathbf{x}_{i, sy}, \mathbf{x}_{i, nn}, \mathbf{w}_{sy}, \mathbf{w}_{nn}) \label{eq:bilevel_learning_value_function_reformulation} \\
    & \quad \quad \quad \quad \quad - V_{\mathbf{y}^{*}_{i}}(\mathbf{w}_{sy}, \mathbf{w}_{nn}) \leq 0, \quad \forall i \in \{1, \cdots, P\} 
    . \nonumber
\end{align}
}%
The formulation in \eqref{eq:bilevel_learning_value_function_reformulation} is referred to as a \emph{value-function} approach in bilevel literature \citep{outrata:zor90, liu:icml21, liu:neurips22, sow:arxiv22, kwon:icml23}.
Value-function approaches view the bilevel program as a constrained optimization problem by leveraging the value-function as a tight lower bound on the lower-level objective.

The inequality constraints in \eqref{eq:bilevel_learning_value_function_reformulation} do not satisfy any of the standard \emph{constraint qualifications} that ensure the feasible set near the optimal point is similar to its linearized approximation \citep{nocedal:book06}.
This raises a challenge for providing theoretical convergence guarantees for constrained optimization techniques.
Following a recent line of value-function approaches to bilevel programming \citep{liu:icml21, sow:arxiv22, liu:arxiv23}, we overcome this challenge by allowing at most an $\iota > 0$ violation in each constraint in \eqref{eq:bilevel_learning_value_function_reformulation}.
With this relaxation, strictly feasible points exist and, for instance, the linear independence constraint qualification (LICQ) can hold.

Another challenge that arises from \eqref{eq:bilevel_learning_value_function_reformulation} is that the energy function of NeSy-EBMs is typically non-differentiable with respect to the targets and even infinite-valued to implicitly represent constraints.
As a result, penalty or augmented Lagrangian functions derived from   \eqref{eq:bilevel_learning_value_function_reformulation} are intractable.
Therefore, we substitute each instance of the energy function in the constraints of \eqref{eq:bilevel_learning_value_function_reformulation} with the following function:
{
% \small
\begin{align}
    \label{eq:moreau_envelop_of_energy}
    & M_{i}(\mathbf{y}; \mathbf{w}_{sy}, \mathbf{w}_{nn}, \rho) \\ 
    & \, := \inf_{\hat{\mathbf{y}} \in \mathcal{Y}} \left( E(\hat{\mathbf{y}}, \mathbf{x}_{i, sy}, \mathbf{x}_{i, nn}, \mathbf{w}_{sy}, \mathbf{w}_{nn}) + \frac{1}{2 \rho} \Vert \hat{\mathbf{y}} - \mathbf{y} \Vert_{2}^{2} \right), \nonumber
\end{align}
}%
where $\rho$ is a positive scalar.
For convex $E$, \eqref{eq:moreau_envelop_of_energy} is the Moreau envelope of the energy function \cite{rockafellar:book70, boyd:ftml13}.
In general, even for non-convex energy functions, the smoothing in \eqref{eq:moreau_envelop_of_energy} preserves global minimizers and minimum values, i.e., $\mathbf{y}_{i}^{*}(\mathbf{w}_{sy}, \mathbf{w}_{nn}) = \argmin_{\mathbf{y}} M_{i}(\mathbf{y}; \mathbf{w}_{sy}, \mathbf{w}_{nn}, \rho)$ and $V_{\mathbf{y}_{i}^{*}}(\mathbf{w}_{sy}, \mathbf{w}_{nn}) = \min_{\mathbf{y}} M_{i}(\mathbf{y}; \mathbf{w}_{sy}, \mathbf{w}_{nn}, \rho)$.
Moreover, under \assumptionref{assumption:lower_level_singleton} each $M_{i}$ is finite for all $\mathbf{y} \in \mathcal{Y}$ even if the energy function is not.
When the energy function is a lower semi-continuous convex function, its Moreau envelope is convex, finite, and continuously differentiable, and its gradient with respect to $\mathbf{y}$ is: 
{
% \small
\begin{align}
    & \nabla_{\mathbf{y}} M_{i}(\mathbf{y}; \mathbf{w}_{sy}, \mathbf{w}_{nn}, \rho) = \frac{1}{\rho} \Bigg( \mathbf{y} - \\
    & \argmin_{\hat{\mathbf{y}} \in \mathcal{Y}} \Big ( \rho E(\hat{\mathbf{y}}, \mathbf{x}_{i, sy}, \mathbf{x}_{i, nn}, \mathbf{w}_{sy}, \mathbf{w}_{nn}) + \frac{1}{2} \Vert \hat{\mathbf{y}} - \mathbf{y} \Vert_{2}^{2} \Big ) \Bigg).  \nonumber
    % \\ 
    % & = \frac{1}{\rho} \left( \mathbf{y} - \textrm{prox}_{\rho E(\cdot, \mathbf{x}_{i}, \mathbf{w}_{sy}, \mathbf{w}_{nn})}(\mathbf{y}) \right).
\end{align}
}%
Convexity is a sufficient but not necessary condition to ensure each $M_{i}$ is differentiable with respect to the targets.
See \cite{bonnans:book00} for results regarding the sensitivity of optimal value-functions to perturbations.
The use of the Moreau envelope of the energy function is a novel method for ensuring smoothness in NeSy learning.

Altogether, we propose the following relaxed and smoothed value-function formulation of bilevel NeSy learning in \eqref{eq:bilevel_learning}:
{
% \small
\begin{align}
    & \argmin_{\substack{(\mathbf{w}_{sy}, \mathbf{w}_{nn}) \\ (\mathbf{y}_{1}, \cdots , \mathbf{y}_{P})}} 
    \sum_{i = 1}^{P} \left ( d(\mathbf{y}_{i}, \mathbf{t}_{i}) + L_{Val}(E(\cdot, \cdot, \cdot, \mathbf{w}_{sy}, \mathbf{w}_{nn}), S_{i}) \right ) \nonumber \\
    \label{eq:relaxed_smoothed_value_function_bound_constrained_approach}
    & \quad \textrm{s.t.} \quad M_{i}(\mathbf{y}_{i}; \mathbf{w}_{sy}, \mathbf{w}_{nn}, \rho) - V_{\mathbf{y}^{*}_{i}}(\mathbf{w}_{sy}, \mathbf{w}_{nn}) \leq \iota, \nonumber \\
    & \quad \quad \quad \quad \forall i \in \{ 1, \cdots, P\},
\end{align}
}% 
The formulation \eqref{eq:relaxed_smoothed_value_function_bound_constrained_approach} is the core of our proposed NeSy-EBM learning framework outlined in \algoref{alg:nesy_ebm_learning} below. 
The algorithm proceeds by approximately solving instances of \eqref{eq:relaxed_smoothed_value_function_bound_constrained_approach} in a sequence defined by a decreasing $\iota$.
This is a graduated approach to solving \eqref{eq:relaxed_smoothed_value_function_bound_constrained_approach} with instances of \eqref{eq:relaxed_smoothed_value_function_bound_constrained_approach} that are increasingly tighter approximations.

\begin{algorithm}[H]
\small
\caption{NeSy-EBM Learning Framework}
\label{alg:nesy_ebm_learning}
\begin{algorithmic}[1]
    \REQUIRE{Moreau Param.: $\rho$, \\ Starting weights: $(\mathbf{w}^{(0)}_{sy}, \mathbf{w}^{(0)}_{nn}) \in \mathcal{W}_{sy} \times \mathcal{W}_{nn}$}
    \STATE{$\mathbf{y}_{i}^{(0)} \gets (\mathbf{t}_{i}, \mathbf{z}_{i}^{*}), \, \forall{i = 1, \cdots, P};$}
    \STATE{$\iota^{(0)} \gets \max_{i} M_{i}(\mathbf{y}_{i}^{(0)}; \mathbf{w}_{sy}^{(0)}, \mathbf{w}_{nn}^{(0)}, \rho) - V_{\mathbf{y}_{i}^{*}}(\mathbf{w}_{sy}^{(0)}, \mathbf{w}_{nn}^{(0)})$;}
    \FOR{$t = 0, 1, 2, \cdots$}
        \STATE{Find $\mathbf{w}_{sy}^{(t + 1)}, \mathbf{w}_{nn}^{(t + 1)}, \mathbf{y}_{1}^{(t + 1)}, \cdots, \mathbf{y}_{P}^{(t + 1)}$ \\ \quad minimizing \eqref{eq:relaxed_smoothed_value_function_bound_constrained_approach} with $\iota^{(t)}$.}
        \IF{Stopping criterion satisified}
            \STATE{Stop with: $\mathbf{w}_{sy}^{(t + 1)}, \mathbf{w}_{nn}^{(t + 1)}, \mathbf{y}_{1}^{(t + 1)}, \cdots, \mathbf{y}_{P}^{(t + 1)}$;}
        \ENDIF
        \STATE{$\iota^{(t + 1)} \gets \frac{1}{2} \cdot \iota^{(t)};$}
    \ENDFOR
\end{algorithmic}
\end{algorithm}
We suggest starting points for each $\mathbf{y}_{i}$ to be the latent inference minimizer and $\iota$ to be the maximum difference in the value-function and the smooth energy function at $\mathbf{y}^{i}$, $M_{i}(\mathbf{y}_{i}^{(0)}; \mathbf{w}_{sy}^{(0)}, \mathbf{w}_{nn}^{(0)}, \rho)$. 
At this suggested starting point, the supervised loss is initially $0$, and the subproblem reduces to minimizing the learning objective without increasing the most violated constraint.
Then, the value for $\iota$ is halved every time an approximate solution to the subproblem, \eqref{eq:relaxed_smoothed_value_function_bound_constrained_approach}, is reached.
The outer loop of the NeSy-EBM learning framework may be stopped by either watching the progress of a training or validation evaluation metric or by specifying a final value for $\iota$.

Each instance of \eqref{eq:relaxed_smoothed_value_function_bound_constrained_approach} in \algoref{alg:nesy_ebm_learning} can be optimized using only first-order gradient-based methods.
Specifically, we employ the bound-constrained augmented Lagrangian algorithm, Algorithm 17.4 from \citenoun{nocedal:book06}, which finds approximate minimizers of the problem's augmented Lagrangian for a fixed setting of the penalty parameters using gradient descent.
To simplify notation, let the equality constraints in \eqref{eq:relaxed_smoothed_value_function_bound_constrained_approach} be denoted by:
{
% \small
\begin{align*}
    & c_{i}(\mathbf{y}_{i}, \mathbf{w}_{sy}, \mathbf{w}_{nn}; \iota) \\
    & \quad := M_{i}(\mathbf{y}_{i}; \mathbf{w}_{sy}, \mathbf{w}_{nn}, \rho) - V_{\mathbf{y}^{*}_{i}}(\mathbf{w}_{sy}, \mathbf{w}_{nn}) - \iota,
\end{align*}
}%
for each constraint indexed $i \in \{1, \cdots, P\}$.
% Moreover, let $c(\mathbf{y}_{1}, \cdots, \mathbf{y}_{P}, \mathbf{w}_{sy}, \mathbf{w}_{nn}; \iota) := [ c_{i}(\mathbf{y}_{i}, \mathbf{w}_{sy}, \mathbf{w}_{nn}; \iota) ]_{i = 1}^{P}.$
The augmented Lagrangian function corresponding to \eqref{eq:relaxed_smoothed_value_function_bound_constrained_approach} introduces a quadratic penalty parameter $\mu$ and $P$ linear penalty parameters $\mathbf{\lambda} := [ \lambda_{i} ]_{i = 1}^{P}$, as follows:
{
\small
\begin{align}
    \label{eq:augmented_lagrangian}
    & \mathcal{L}_{A}(\mathbf{w}_{sy}, \mathbf{w}_{nn}, \mathbf{y}_{1}, \cdots, \mathbf{y}_{p}, \mathcal{S}; \mathbf{\lambda}, \mu, \iota) \\
    & \quad :=  \sum_{i = 1}^{P} \left ( d(\mathbf{y}_{i}, \mathbf{t}_{i}) + L_{Val}(E(\cdot, \cdot, \cdot, \mathbf{w}_{sy}, \mathbf{w}_{nn}), S_{i}) \right ) \nonumber \\
    & \quad \quad + \frac{\mu}{2} \sum_{i = 1}^{P} (c_{i}(\mathbf{y}_{i}, \mathbf{w}_{sy}, \mathbf{w}_{nn}; \iota) + s_{i})^{2} \nonumber \\
    & \quad \quad + \sum_{i = 1}^{P} \lambda_{i} (c_{i}(\mathbf{y}_{i}, \mathbf{w}_{sy}, \mathbf{w}_{nn}; \iota) + s_{i}). \nonumber
\end{align}
}%
where we introduced $P$ slack variables, $\mathbf{s} = \left[s_{i} \right]_{i = 1}^{P}$, for each inequality constraint.
We make the following assumption to ensure the augmented Lagrangian function is differentiable:
% with respect to the weights and target variables:
\begin{assumption}
    \label{assumption:differentiable_value function}
    Every $V_{y^*_{i}}$, $V_{z^*_{i}}$, and $M_{i}$ is differentiable with respect to the weights. 
\end{assumption}

We employ the bound-constrained augmented Lagrangian algorithm to solve \eqref{eq:relaxed_smoothed_value_function_bound_constrained_approach} (see \appref{appendix:extended_bilevel_nesy_learning} for details).
This method provides a principled algorithm for updating the penalty parameters and ensures fundamental convergence properties of our learning framework.
Notably, we have that limit points of the iterate sequence are stationary points of $\Vert c(\mathbf{y}_{1}, \cdots, \mathbf{y}_{P}, \mathbf{w}_{sy}, \mathbf{w}_{nn}) + \mathbf{s} \Vert^{2}$ when the problem has no feasible points.
When the problem is feasible and  LICQ holds at the limits, they are KKT points of \eqref{eq:relaxed_smoothed_value_function_bound_constrained_approach} (Theorem 17.2 in \cite{nocedal:book06}).
Convergence rates and stronger guarantees are likely possible from analyzing the structure of the energy function for specific NeSy-EBMs and is a direction for future work.

\section{Deep Hinge-loss Markov Random Fields}
\label{sec:neupsl_deep_hlmrfs}

We demonstrate the applicability of our learning framework with Neural Probabilistic Soft Logic (NeuPSL), a general class of NeSy-EBMs designed for scalable joint reasoning \citep{pryor:ijcai23}.
In NeuPSL, relations and attributes are represented by \emph{atoms}, and dependencies between atoms are encoded with first-order logical clauses and linear arithmetic inequalities referred to as \emph{rules}.
Atom values can be target variables, observations, or outputs from a neural network. 
The rules and atoms are translated into potentials measuring rule satisfaction. 
Then, the potentials are aggregated to define the energy function for a member of a tractable class of graphical models: \emph{deep hinge-loss Markov random fields} (deep HL-MRF).\footnote{To simply exposition, a single deep HL-MRF energy function aggregated over training examples is presented.}
\begin{definition}
    Let $\mathbf{g} = [g_{i}]_{i = 1}^{n_{g}}$ be functions with corresponding weights $\mathbf{w}_{nn}=[\mathbf{w}_{nn, i}]_{i = 1}^{n_{g}}$ and inputs $\mathbf{x}_{nn}$ such that $g_{i}:(\mathbf{w}_{nn, i}, \mathbf{x}_{nn}) \mapsto [0, 1]$.
    Let $\mathbf{y} \in [0, 1]^{n_{y}}$ and $\mathbf{x}_{sy} \in [0, 1]^{n_{x}}$.
    A \textbf{deep hinge-loss potential} is a function of the form:
    {\small
    \begin{align*}
        & \phi(\mathbf{y}, \mathbf{x}_{sy}, \mathbf{g}(\mathbf{x}_{nn}, \mathbf{w}_{nn})) \\
        & \quad := (\max\{\mathbf{a}_{\phi, \mathbf{y}}^T \mathbf{y} + \mathbf{a}_{\phi, \mathbf{x}_{sy}}^T \mathbf{x}_{sy} + \mathbf{a}_{\phi, \mathbf{g}}^T \mathbf{g}(\mathbf{x}_{nn}, \mathbf{w}_{nn}) + b_{\phi}, 0\})^{p},
        % \label{eq:deep_hlmrf_potential}
    \end{align*}
    }%
    where $\mathbf{a}_{\phi, \mathbf{y}} \in \mathbb{R}^{n_{y}}$, $\mathbf{a}_{\phi, \mathbf{x}} \in \mathbb{R}^{n_{x}}$, and $\mathbf{a}_{\phi, \mathbf{g}} \in \mathbb{R}^{n_{g}}$ are variable coefficient vectors, $b_{\phi} \in \mathbb{R}$ is a vector of constants, and $p \in \{1, 2\}$.
    Let $\mathcal{T} = [\tau_i]_{i = 1}^{r}$ denote an ordered partition of a set of $m$ deep hinge-loss potentials. 
    Further, define $\mathbf{\Phi}(\mathbf{y} , \mathbf{x}_{sy}, \mathbf{g}(\mathbf{x}_{nn}, \mathbf{w}_{nn})) := [ \sum_{k \in \tau_i} \phi_{k}(\mathbf{y}, \mathbf{x}_{sy}, \mathbf{g}(\mathbf{x}_{nn}, \mathbf{w}_{nn})) ]_{i = 1}^{r}$.
    Let $\mathbf{w}_{sy}$ be a vector of $r$ non-negative symbolic weights corresponding to the partition $\mathcal{T}$.
    Then, a \textbf{deep hinge-loss energy function} is:
    {
    % \small
    \begin{align}
        \label{eq:deep_hlmrf_energy_function}
        & E(\mathbf{y}, \mathbf{x}_{sy}, \mathbf{x}_{nn}, \mathbf{w}_{sy}, \mathbf{w}_{nn}) \\
        & \quad := \mathbf{w}_{sy}^T \mathbf{\Phi}(\mathbf{y} , \mathbf{x}_{sy}, \mathbf{g}(\mathbf{x}_{nn}, \mathbf{w}_{nn})). \nonumber
    \end{align}
    }%
    Let $\mathbf{a}_{c_k, \mathbf{y}} \in \mathbb{R}^{n_{y}}$, $\mathbf{a}_{c_k, \mathbf{x}} \in \mathbb{R}^{n_{x}}$, $\mathbf{a}_{c_k, \mathbf{g}} \in \mathbb{R}^{n_{g}}$, and $b_{c_k} \in \mathbb{R}$ for each $k \in {1, \dotsc, q}$ and $q \geq 0$ be vectors defining linear inequality constraints and a feasible set:
    {
    % \small
    \begin{align*}
    & \mathbf{\Omega}(\mathbf{x}_{sy}, \mathbf{g}
    % (\mathbf{x}_{nn}, \mathbf{w}_{nn})
    )
    := 
    % \\ & \quad
    \Big \{
    \mathbf{y} \in [0, 1]^{n_y} \, \vert \, \\ 
            & \, 
            \mathbf{a}_{c_k, \mathbf{y}}^T \mathbf{y} + \mathbf{a}_{c_k, \mathbf{x}}^T \mathbf{x}_{sy} + \mathbf{a}_{c_k, \mathbf{g}}^T \mathbf{g}
            % (\mathbf{x}_{nn}, \mathbf{w}_{nn}) 
            + b_{c_k} \leq 0 
            \, , \forall \, k=1,\dotsc,q \,
    \Big \}.
    \end{align*}
    }%
    Then a \textbf{deep hinge-loss Markov random field} defines the conditional probability density:
    {\small
    % \begin{subequations}
        \begin{align}
            & P(\mathbf{y} \vert \mathbf{x}_{sy}, \mathbf{x}_{nn}) := \\ 
            & \begin{cases}
            \frac{\exp (-E(\mathbf{y}, \mathbf{x}_{sy}, \mathbf{x}_{nn}, \mathbf{w}_{sy}, \mathbf{w}_{nn}))}{\int_{\mathbf{y}} \exp(-E(\mathbf{y}, \mathbf{x}_{sy}, \mathbf{x}_{nn}, \mathbf{w}_{sy}, \mathbf{w}_{nn})) d\mathbf{y}} & \mathbf{y} \in \mathbf{\Omega}(\mathbf{x}_{sy}, \mathbf{g}(\mathbf{x}_{nn}, \mathbf{w}_{nn})) \\
            0 & o.w.
            \end{cases} \nonumber
            % \\
            % Z(\cdot) &:= \int_{\mathbf{y} \in \mathbf{\Omega}(\mathbf{x}_{sy}, \mathbf{g}(\mathbf{x}_{nn}, \mathbf{w}_{nn}))} \exp(-E(\mathbf{y}, \mathbf{x}_{sy}, \mathbf{x}_{nn}, \mathbf{w}_{sy}, \mathbf{w}_{nn})) d\mathbf{y}
        \end{align}
    % \end{subequations}
    \label{def:deep_hlmrf}
    }%
\end{definition}

Based on \defnref{def:deep_hlmrf}, NeuPSL is a NeSy-EBM with an extended-value deep HL-MRF energy function capturing the constraints defining the feasible set.
Further, NeuPSL inference is finding the MAP state of the conditional distribution defined by a deep HL-MRF, i.e., finding the minimizer of the energy function over the feasible set.
{
% \small
\begin{align}
    \label{eq:inference_primal}
    & \min_{\mathbf{y} \in \mathbb{R}^{n_{\mathbf{y}}}} \, \mathbf{w}_{sy}^T \mathbf{\Phi}(\mathbf{y} , \mathbf{x}_{sy}, \mathbf{g}(\mathbf{x}_{nn}, \mathbf{w}_{nn})) \\
        & \quad \textrm{s.t.}  \quad       
        \mathbf{y} \in \mathbf{\Omega}(\mathbf{x}_{sy}, \mathbf{g}(\mathbf{x}_{nn}, \mathbf{w}_{nn})). \nonumber
\end{align}
}%
As each of the potentials are convex, \eqref{eq:inference_primal} is a non-smooth convex linearly constrained program.

\subsection{A smooth formulation of inference}
\label{sec:lcqp_inference}

We introduce a primal and dual formulation of NeuPSL inference as a linearly constrained convex quadratic program (LCQP).
(See \appref{appendix:lcqp_inference} for details.)
In summary, $m$ slack variables with lower bounds and $2 \cdot n_{\mathbf{y}} + m$ linear constraints are defined to represent the target variable bounds and deep hinge-loss potentials.
All $2 \cdot n_{\mathbf{y}} + m$ variable bounds, $m$ potentials, and $q \geq 0$ constraints are collected into a $(2 \cdot n_{\mathbf{y}} + q + 2 \cdot m) \times (n_{\mathbf{y}} + m)$ dimensional matrix $\mathbf{A}$ and a vector of $(2 \cdot n_{\mathbf{y}} + q + 2 \cdot m)$ elements that is an affine function of the neural predictions and symbolic inputs $\mathbf{b}(\mathbf{x}_{sy}, \mathbf{g}(\mathbf{x}_{nn}, \mathbf{w}_{nn}))$.
Moreover, the slack variables and a $(n_{\mathbf{y}} + m) \times (n_{\mathbf{y}} + m)$ positive semi-definite diagonal matrix, $\mathbf{D}(\mathbf{w}_{sy})$, and a $(n_{\mathbf{y}} + m)$ dimensional vector, $\mathbf{c}(\mathbf{w}_{sy})$, are created using the symbolic weights to define a quadratic objective.
Further, we gather the original target variables and the slack variables into a vector $\mathbf{\nu} \in \mathbb{R}^{n_{\mathbf{y}} + m}$.
Altogether, the regularized convex LCQP reformulation of NeuPSL inference is:
{
% \small
\begin{align}
    \label{eq:regularized_lcqp_primal}
    & V(\mathbf{w}_{sy}, \mathbf{b}(\mathbf{x}_{sy}, \mathbf{g}(\mathbf{x}_{nn}, \mathbf{w}_{nn}))) := \\ 
    & \min_{\mathbf{\nu} \in \mathbb{R}^{n_{\mathbf{y}} + m}} \, 
        \mathbf{\nu}^T (\mathbf{D}(\mathbf{w}_{sy}) + \epsilon \mathbf{I}) \mathbf{\nu} + \mathbf{c}(\mathbf{w}_{sy})^T \mathbf{\nu}  \nonumber \\
    & \quad \textrm{s.t.} \quad       
     \mathbf{A} \mathbf{\nu} + \mathbf{b}(\mathbf{x}_{sy}, \mathbf{g}(\mathbf{x}_{nn}, \mathbf{w}_{nn})) \leq 0 \nonumber,
\end{align}
}%
where $\epsilon \geq 0$ is a scalar regularization parameter added to the diagonal of $\mathbf{D}$ to ensure strong convexity (needed in the next subsection).
The effect of the added regularization is empirically studied in \appref{appendix:empirical_lcqp_regularization}.
The function $V(\mathbf{w}_{sy}, \mathbf{b}(\mathbf{x}_{sy}, \mathbf{g}(\mathbf{x}_{nn}, \mathbf{w}_{nn})))$ in \eqref{eq:regularized_lcqp_primal} is the optimal value-function of the LCQP formulation of NeuPSL inference referred to in the previous section.

By Slater's constraint qualification, we have strong duality when there is a feasible solution to \eqref{eq:regularized_lcqp_primal}.
In this case, an optimal solution to the dual yields an optimal solution to the primal problem.
The Lagrange dual problem of \eqref{eq:regularized_lcqp_primal} is:
{
\small
\begin{align}
    \label{eq:dual_lcqp_inference}
    & \min_{\mathbf{\mu} \in \mathbb{R}_{\geq 0}^{2 \cdot n_{\mathbf{y}} + m + q}}
        \quad h(\mathbf{\mu}; \mathbf{w}_{sy}, \mathbf{b}(\mathbf{x}_{sy}, \mathbf{g}(\mathbf{x}_{nn}, \mathbf{w}_{nn}))) \\
        & \quad \quad := \frac{1}{4} \mathbf{\mu}^T \mathbf{A} (\mathbf{D}(\mathbf{w}_{sy}) + \epsilon \mathbf{I})^{-1} \mathbf{A}^T \mathbf{\mu} \nonumber \\ 
        & \quad \quad \quad + \frac{1}{2} (\mathbf{A} (\mathbf{D}(\mathbf{w}_{sy}) + \epsilon \mathbf{I})^{-1} \mathbf{c}(\mathbf{w}_{sy}) \nonumber \\
        & \quad \quad \quad - 2 \mathbf{b}(\mathbf{x}_{sy}, \mathbf{g}(\mathbf{x}_{nn}, \mathbf{w}_{nn})))^T \mathbf{\mu}, \nonumber
\end{align}
}%
where $\mathbf{\mu}$ is the vector of dual variables and $h(\mathbf{\mu}; \mathbf{w}_{sy}, \mathbf{b}(\mathbf{w}_{nn}))$ is the LCQP dual objective function.
% The dual formulation has more decision variables but is only over non-negativity constraints rather than a complex polyhedron.
As $(\mathbf{D}(\mathbf{w}_{sy}) + \epsilon \mathbf{I})$ is diagonal, it is easy to invert, and thus it is practical to work in the dual space and map dual to primal variables.
The dual-to-primal variable mapping is:
{
% \small
\begin{align}
    \mathbf{\nu} \gets - \frac{1}{2} (\mathbf{D}(\mathbf{w}_{sy}) + \epsilon \mathbf{I})^{-1} (\mathbf{A}^T \mathbf{\mu} + \mathbf{c}(\mathbf{w}_{sy})).
    % \label{eq:dual_primal_translation}
\end{align}
}%
On the other hand, the primal-to-dual mapping is more computationally expensive and requires calculating a pseudo-inverse of the constraint matrix $\mathbf{A}$.

\subsection{Continuity of inference}
\label{sec:continuity_of_inference}

We use the LCQP formulation in \eqref{eq:regularized_lcqp_primal} to establish continuity and curvature properties of the NeuPSL energy minimizer and the optimal value-function provided in the following theorem.
The proof is provided in \appref{appendix:continuity_of_inference}.

\begin{theorem}
    \label{thm:continuity_properties}
    Suppose for any setting of $\mathbf{w}_{nn} \in \mathbb{R}^{n_g}$ there is a feasible solution to NeuPSL inference \eqref{eq:regularized_lcqp_primal}.
    Further, suppose $\epsilon > 0$, $\mathbf{w}_{sy} \in \mathbb{R}_{+}^{r}$, and $\mathbf{w}_{nn} \in \mathbb{R}^{n_g}$.
    Then:
    \begin{itemize}[leftmargin=*,noitemsep,topsep=0pt]
        \item The minimizer of \eqref{eq:regularized_lcqp_primal}, $\mathbf{y}^*(\mathbf{w}_{sy}, \mathbf{w}_{nn})$, is a $O(1 / \epsilon)$ Lipschitz continuous function of $\mathbf{w}_{sy}$.
        \item $V(\mathbf{w}_{sy}, \mathbf{b}(\mathbf{x}_{sy}, \mathbf{g}(\mathbf{x}_{nn}, \mathbf{w}_{nn})))$, is concave over $\mathbf{w}_{sy}$ and convex over $\mathbf{b}(\mathbf{x}_{sy}, \mathbf{g}(\mathbf{x}_{nn}, \mathbf{w}_{nn}))$.
        \item $V(\mathbf{w}_{sy}, \mathbf{b}(\mathbf{x}_{sy}, \mathbf{g}(\mathbf{x}_{nn}, \mathbf{w}_{nn})))$ is differentiable with respect to $\mathbf{w}_{sy}$. Moreover,
        {
        % \small
        \begin{align*}
            & \nabla_{\mathbf{w}_{sy}} V(\mathbf{w}_{sy}, \mathbf{b}(\mathbf{x}_{sy}, \mathbf{g}(\mathbf{x}_{nn}, \mathbf{w}_{nn}))) \\
            & \quad = \mathbf{\Phi}(\mathbf{y}^{*}(\mathbf{w}_{sy}, \mathbf{w}_{nn}), \mathbf{x}_{sy}, \mathbf{g}(\mathbf{x}_{nn}, \mathbf{w}_{nn})).
        \end{align*}
        }%
        Furthermore, $\nabla_{\mathbf{w}_{sy}} V(\mathbf{w}_{sy}, \mathbf{b}(\mathbf{x}_{sy}, \mathbf{g}(\mathbf{x}_{nn}, \mathbf{w}_{nn})))$ is Lipschitz continuous over $\mathbf{w}_{sy}$.
        \item If there is a feasible point, $\nu$, strictly satisfying the $i'th$ constriaint of \eqref{eq:regularized_lcqp_primal}, then $V(\mathbf{w}_{sy}, \mathbf{b}(\mathbf{x}_{sy}, \mathbf{g}(\mathbf{x}_{nn}, \mathbf{w}_{nn})))$, is subdifferentiable with respect to the $i'th$ constraint constant, $\mathbf{b}(\mathbf{x}_{sy}, \mathbf{g}(\mathbf{x}_{nn}, \mathbf{w}_{nn}))[i]$. Moreover, 
        {
        % \small
        \begin{align*}
            & \partial_{\mathbf{b}[i]} V(\mathbf{w}_{sy}, \mathbf{b}(\mathbf{x}_{sy}, \mathbf{g}(\mathbf{x}_{nn}, \mathbf{w}_{nn}))) = \{\mathbf{\mu}^{*}[i] \, \vert \\ 
            & \quad \mathbf{\mu}^{*} \in \argmin_{\mathbf{\mu} \in \mathbb{R}_{\geq 0}^{2 \cdot n_{\mathbf{y}} + m + q}}
            h(\mathbf{\mu}; \mathbf{w}_{sy}, \mathbf{b}(\mathbf{x}_{sy}, \mathbf{g}(\mathbf{x}_{nn}, \mathbf{w}_{nn}))) \}.
        \end{align*}
        }%
        Furthermore, if $\mathbf{g}$ is a smooth function of $\mathbf{w}_{nn}$, then so is $\mathbf{b}$, and the set of regular subgradients of $V(\mathbf{w}_{sy}, \mathbf{b}(\mathbf{x}_{sy}, \mathbf{g}(\mathbf{x}_{nn}, \mathbf{w}_{nn})))$ is:
        {
        % \small
        \begin{align}
            \label{eq:value_function_subgradient}
            &\hat{\partial}_{\mathbf{w}_{nn}} V(\mathbf{w}_{sy}, \mathbf{b}(\mathbf{x}_{sy}, \mathbf{g}(\mathbf{x}_{nn}, \mathbf{w}_{nn}))) \supset \\ 
            & 
            \quad 
            \nabla_{\mathbf{w}_{nn}} \mathbf{b}(\mathbf{x}_{sy}, \mathbf{g}(\mathbf{x}_{nn}, \mathbf{w}_{nn}))^T 
            \nonumber \\ 
            & \quad \quad 
            \partial_{\mathbf{b}} V(\mathbf{w}_{sy}, \mathbf{b}(\mathbf{x}_{sy}, \mathbf{g}(\mathbf{x}_{nn}, \mathbf{w}_{nn}))) \nonumber.
        \end{align}
        }%
    \end{itemize}
\end{theorem}

\thmref{thm:continuity_properties} provides a simple explicit form of the value-function gradient with respect to the symbolic weights and regular subgradient with respect to the neural weights.
Moreover, this result is directly applicable to the Moreau envelope of the NeuPSL energy function used in \secref{sec:learning_framework} as it is a regularized value-function.
Thus, \thmref{thm:continuity_properties} supports the principled application of \algoref{alg:nesy_ebm_learning} for learning both the symbolic and neural weights of a NeuPSL model.
% Specifically, when there is a $\mathbf{\nu}$ that is both feasible and strictly satisfies every constraint $i \in \{1, \cdots, 2 \cdot n_{\mathbf{y}}, q + m\}$ in \eqref{eq:regularized_lcqp_primal} with a constant $\mathbf{b}(\mathbf{x}_{sy}, \mathbf{g}(\mathbf{x}_{nn}, \mathbf{w}_{nn}))[i]$ that is a non-trivial function of the neural weights, i.e., $\nabla_{\mathbf{w}_{nn}} \mathbf{b}(\mathbf{x}_{sy}, \mathbf{g}(\mathbf{x}_{nn}, \mathbf{w}_{nn}))[i] \neq \mathbf{0}$, then \eqref{eq:value_function_subgradient} produces a regular subgradient of the value-function that can be used for backpropagation.

\subsection{Dual block coordinate descent}
\label{sec:dual_bcd}
 
The regular subgradients in \thmref{thm:continuity_properties} are functions of the optimal dual variables of the LCQP inference problem in \eqref{eq:dual_lcqp_inference}.
Thus, one could compute value-function gradients for learning by solving the primal inference problem \eqref{eq:inference_primal} using an existing algorithm and then map the optimal primal variables to dual variables.
However, the primal-to-dual mapping requires computing a pseudo-inverse of the constraint matrix.
For this reason, we introduce a block coordinate descent (BCD)~\citep{wright:mp15} algorithm for working directly with the dual LCQP formulation of inference.
Details of the algorithm are provided in \appref{appendix:extended_dual_bcd}.
Our dual BCD algorithm is the first method specialized for dual LCQP inference. 
It is, therefore, also the first to produce optimal dual variables that directly yield both optimal primal variables and principled gradients for learning, all without the need to compute a pseudo-inverse of the constraint matrix.

The dual BCD algorithm proceeds by successively minimizing the objective along the subgradient of a block of dual variables.
For this reason, dual BCD guarantees descent at every iteration, partially explaining its effectiveness at leveraging warm-starts and improving learning runtimes.
The algorithm is stopped when the primal-dual gap drops below a threshold $\delta>0$.
We suggest a practical choice of variable blocks with efficient methods for computing the objective subgradients and solving the steplength subproblems.

Additionally, we develop an efficient method for identifying connected components of the deep HLM-MRF factor graph, yielding a variable partition that the dual objective is additively separable over to parallelize the BCD updates.
We call this parallelization approach connected component parallel dual BCD (CC D-BCD).

For sparse factor graphs with few connected components (e.g., a chain), the CC variant of D-BCD is ineffective as the updates cannot be distributed to maximize CPU utilization.
Thus, inspired by lock-free parallelization strategies~\citep{bertsekas:book95, recht:neurips11, liu:jmlr15}, we additionally propose lock-free parallel dual BCD (LF D-BCD), an alternative parallelization technique for the dual BCD inference algorithm that sacrifices the theoretical guaranteed descent property for significant runtime improvements.
In our empirical evaluation, we show that the lock-free dual BCD algorithm consistently finds a solution satisfying the stopping criterion and, surprisingly, is still highly effective at leveraging warm starts.

\begin{table*}[h]
    \centering
    \caption{Datasets used for empirical evaluations.}
    \vskip 0.1in
    \label{tab:datasets}
    \scalebox{1.0}{
    \begin{tabular}{l||c|c|c}
        \toprule
        \textbf{Dataset} & \textbf{Deep} & \textbf{Task} & \textbf{Perf. Metric} \\
        \midrule
        \midrule
        Debate~\cite{hasan:ijcnlp13} & & Stance Class. & AUROC \\
        4Forums~\cite{walker:lrec12} & & Stance Class. & AUROC \\
        Epinions~\citep{richardson:iswc03} & & Link Pred. & AUROC \\
        DDI~\citep{wishart:nar06} & & Link Pred. & AUROC \\
        Yelp~\citep{yelp:yelp23} & & Regression & MAE \\ 
        Citeseer~\citep{sen:aim08} & $\checkmark$ & Node Class. & Accuracy \\
        Cora~\citep{sen:aim08} & $\checkmark$ & Node Class. & Accuracy \\
        MNIST-Add.\citep{manhaeve:neurips18} & $\checkmark$ & Image Class. & Accuracy \\
        \bottomrule
    \end{tabular}
}
\end{table*}

\section{Empirical evaluation}
\label{sec:empirical_evaluation}

We evaluate the runtime and prediction performance of our proposed NeSy inference and parameter learning algorithms on the $8$ datasets in \tabref{tab:datasets}\footnote{All code and data is available at \url{https://github.com/linqs/dickens-icml24}.}.
The table includes the dataset's inference task, the associated prediction performance metric, and whether the corresponding NeuPSL model has deep neural network parameters.
Unless noted otherwise, all experiments are run on $5$ splits and the average and standard deviation of times and performance metric values are reported.
Details on the datasets, hardware specifications, hyperparameter searches, and model architectures are provided in \appref{appendix:extended_evaluation}.

For learning experiments in \secref{sec:learning_runtime} and \secref{sec:learning_performance}, NeuPSL models with weights trained using value-based learning losses, e.g., energy and structured perceptron (SP), use mirror descent~\citep{kivinen:ic97, shalevshwartz:ftml11} on the symbolic weights constrained to the unit simplex and Adam~\citep{kingma:iclr17} for the neural weights.
NeuPSL models with weights trained using minimizer-based losses, e.g., mean squared error (MSE) and binary cross entropy (BCE), use our proposed NeSy learning framework in \algoref{alg:nesy_ebm_learning} with a scaled energy loss term added to the objective as in \eqref{eq:bilevel_learning}.
Moreover, optimization of the augmented Lagrangian, line 4 of \algoref{alg:nesy_ebm_learning}, is performed using the bound-constrained augmented Lagrangian algorithm (\appref{appendix:extended_bilevel_nesy_learning}) with mirror descent on the symbolic weights and Adam for the neural weights. 

\subsection{Inference runtime}
\label{sec:inference_runtime}

We begin by examining the runtime of symbolic inference.  
We evaluate the alternating direction method of multipliers (ADMM)~\cite{boyd:ftml10}, the current state-of-the-art inference algorithm for NeuPSL, and our proposed inference algorithms: connected component parallel dual BCD (CC D-BCD) and lock-free parallel dual BCD (LF D-BCD).
We also evaluate the performance of Gurobi, a leading off-the-shelf optimizer, and subgradient descent (GD) in \appref{appendix:extended_inference_runtime} and show ADMM and dual BCD consistently match or outperform the proprietary solver.
All inference algorithms have access to the same computing resources.
We run a hyperparameter search, detailed in \appref{appendix:extended_inference_runtime}, for each algorithm, and the configuration yielding a prediction performance that is within a standard deviation of the best and completed with the lowest runtime is reported.
All algorithms are stopped when the $L_{\infty}$ norm of the primal variable change between iterates is less than $0.001$.

\begin{table}[H]
    \centering
    \caption{Time in seconds for inference using ADMM and our proposed CC D-BCD and LF D-BCD algorithms on each dataset.}
    \vskip 0.1in
    \label{tab:inference_time}
    \scalebox{0.85}{
    \begin{tabular}{l||c||c|c}
    \toprule  
    & \textbf{ADMM} & \textbf{CC D-BCD} & \textbf{LF D-BCD} \\
    \midrule
    \midrule
    \textbf{Debate} & $9.98 \pm 1.13$ & $\mathbf{0.05 \pm 0.02}$ & $\mathbf{0.05 \pm 0.03}$ \\
    \textbf{4Forums} &  $15.17 \pm 0.74$ & $0.11 \pm 0.02$ & $\mathbf{0.05 \pm 0.01}$ \\
    \textbf{Epinions} & $0.36 \pm 0.041$ & $1.84 \pm 0.4$ & $\mathbf{0.26 \pm 0.04}$ \\
    \hline
    \textbf{Citeseer} & $0.63 \pm 0.07$ & $1.36 \pm 0.24$ & $\mathbf{0.49 \pm 0.08}$ \\
    \textbf{Cora} & $\mathbf{0.71 \pm 0.07}$ & $6.46 \pm 3.5$ & $0.79 \pm 0.19$ \\
    \textbf{DDI} & $ 7.85 \pm 0.28 $ & $31.47 \pm 0.17$ & $\mathbf{1.76 \pm 0.17}$ \\
    \hline
    \textbf{Yelp} & $\mathbf{6.37 \pm 1.19}$ & $48.44 \pm 3.82$ & $7.58 \pm 0.48$ \\
    \textbf{MNIST-Add1} & $11.45 \pm 1.32 $ & $ \mathbf{10.23 \pm 1.04} $ & $ 115 \pm 45 $ \\
    \textbf{MNIST-Add2} & $285 \pm 66$ & $\mathbf{29.09 \pm 8.00}$ & $1,189 \pm 16$ \\
    \bottomrule
    \end{tabular}
    }
\end{table}

The total average inference runtime in seconds for each algorithm and model is provided in \tabref{tab:inference_time}.
Surprisingly, despite the potential for an inexact solution to the BCD steplength subproblem, LF D-BCD is faster than CC D-BCD in the first $7$ datasets and demonstrates up to $6 \times$ speedup over CC D-BCD in Yelp.
However, in MNIST-Add datasets, CC D-BCD is up to $10 \times$ faster than LF D-BCD as there is a high number of tightly connected components, one for each addition instance.
This behavior highlights the complementary strengths of the two parallelization strategies.
LF D-BCD should be applied to problems with larger factor graph representations that are connected, while CC D-BCD is effective when there are many similarly sized connected components. 

\begin{table*}
    \centering
    \caption{Cumulative time in seconds for ADMM and D-BCD inference during learning with SP and MSE losses.}
    \vskip 0.1in
    \label{tab:learning_runtime}
    \scalebox{0.9}{
    \begin{tabular}{l||c|c||c|c}
    \toprule  
    & \multicolumn{2}{c}{\textbf{SP}} & \multicolumn{2}{c}{\textbf{MSE}}\\
    & \textbf{ADMM} & \textbf{D-BCD} & \textbf{ADMM} & \textbf{D-BCD} \\
    \midrule
    \midrule
    \textbf{Debate} & $ 10.68 \pm 8.63 $  & $ \mathbf{0.34 \pm 0.36} $  & $ 49.00 \pm 31.23 $ & $ \mathbf{0.62 \pm 0.09} $ \\
    \textbf{4Forums} & $ 11.87 \pm 12.81 $ & $ \mathbf{0.65 \pm 0.05} $  & $ 67.09 \pm 13.79 $ & $ \mathbf{1.11 \pm 0.16} $  \\
    \textbf{Epinions} & $ 12.54 \pm 0.37 $  & $ \mathbf{1.33 \pm 0.06} $ & $ 17.48 \pm 0.62 $ & $ \mathbf{2.27 \pm 0.98} $ \\
    \hline
    \textbf{Citeseer} & $ 167 \pm 37 $ & $ \mathbf{41.57 \pm 6.39} $ & $ 225 \pm 32 $ & $ \mathbf{70.01 \pm 5.86} $ \\
    \textbf{Cora} & $ 183 \pm 26 $ & $ \mathbf{48.16 \pm 5.82} $ & $ 241 \pm 37 $ & $ \mathbf{79.62 \pm 13.77} $ \\
    \textbf{DDI} & $ 4,554 \pm 13 $ & $ \mathbf{19.65 \pm 0.30} $ & $ 7,652 \pm 218 $ & $ \mathbf{52.78 \pm 4.23} $  \\
    \hline
    \textbf{Yelp} & $1,835 \pm 47$ & $\mathbf{114 \pm 4}$ & $2,250 \pm 100$ & $\mathbf{170 \pm 12}$ \\
    \textbf{MNIST-Add1} & $1,624 \pm 34$ & $\mathbf{232 \pm 44}$ & $2,942 \pm 109$ & $\mathbf{2,738 \pm 93}$ \\
    \textbf{MNIST-Add2} & TIME-OUT & $\mathbf{804 \pm 106}$ & TIME-OUT & $\mathbf{4,291 \pm 114}$ \\
    \bottomrule
    \end{tabular}
    }
\end{table*}

\begin{table*}
    \centering
    \caption{Prediction performance of HL-MRF models trained on value and minimizer-based losses.}
    \vskip 0.1in
    \label{tab:hl_mrf_prediction_performance}
    \scalebox{0.9}{
    \begin{tabular}{l||c|c||c|c}
        \toprule  
        & \multicolumn{2}{c}{\textbf{Value-Based}} & \multicolumn{2}{c}{\textbf{Bilevel}} \\
        & \textbf{Energy} & \textbf{SP} & \textbf{MSE} & \textbf{BCE} \\
        \midrule
        \midrule
        \textbf{Debate} & $64.76 \pm 9.54$ & $64.68 \pm 11.05$ & $\mathbf{65.33 \pm 11.98}$ & $64.83 \pm 9.70$ \\
        \textbf{4Forums} & $62.96 \pm 6.11$ & $63.15 \pm 6.40$ & $64.22 \pm 6.41$ & $\mathbf{64.85 \pm 6.01}$ \\
        \hline
        \textbf{Epinions} & $78.96 \pm 2.29$ & $79.85 \pm 1.62$ & $\mathbf{81.18 \pm 2.21}$ & $80.89 \pm 2.32$ \\
        \textbf{Citeseer} & $70.29 \pm 1.54$ & $70.92 \pm 1.33$ & $71.22 \pm 1.56$ & $\mathbf{71.94 \pm 1.17}$ \\
        \hline
        \textbf{Cora} & $54.30 \pm 1.74$ & $74.16 \pm 2.32$ & $81.05 \pm 1.41$ & $\mathbf{81.07 \pm 1.31}$ \\
        \textbf{DDI} & $94.54 \pm 0.00$ & $94.61 \pm 0.00$ & $94.70 \pm 0.00$ & $\mathbf{95.08 \pm 0.00}$ \\
        \hline
        \textbf{Yelp} & $18.11 \pm 0.34$ & $18.57 \pm 0.66$ & $18.14 \pm 0.36$ & $\mathbf{17.93 \pm 0.50}$\\
        \bottomrule
    \end{tabular}
    }
\end{table*}

\subsection{Learning runtime}
\label{sec:learning_runtime}

Next, we study how the algorithms applied to solve inference affect the learning runtime with the SP and MSE losses.
Specifically, we examine the cumulative time required for ADMM and D-BCD inference to complete $500$ weight updates on the first $7$ datasets in \tabref{tab:datasets} and $100$ weight updates on MNIST-Add datasets.
Hyperparameters used for SP and MSE learning are reported in \appref{appendix:extended_learning_runtime}.
For inference, we apply the same hyperparameters used in the previous section and the fastest parallelization method for D-BCD. 

\tabref{tab:learning_runtime} shows that the D-BCD algorithm consistently results in the lowest total inference runtime, validating it's ability to leverage warm starts to improve learning runtimes.
Notably, on the DDI dataset, D-BCD achieves  
roughly a $100 \times$ speedup over ADMM.
Moreover, on MNIST-Add2, ADMM 
timed out with over $6$ hours of inference time for SP and MSE learning, while D-BCD accumulated less than $0.5$ and $1.2$ hours of inference runtime on average for SP and MSE, respectively.

\subsection{Learning prediction performance}
\label{sec:learning_performance}

In our final experiment, we analyze the prediction performance of NeuPSL models trained with our NeSy-EBM learning framework.
A hyperparameter search (detailed in \appref{appendix:extended_learning_performance}) is performed over learning steplengths, regularizations, and parameters for \algoref{alg:nesy_ebm_learning}.

\begin{table*}
    \caption{Accuracy of DeepStochlog, GCN, and NeuPSL on Citeseer and Cora.}
    \vskip 0.1in
    \label{tab:deep_hl_mrf_citation_prediction_performance}
    \centering
    \scalebox{0.9}{
    \begin{tabular}{l||c|c||c|c||c|c}
        \toprule 
        & \multicolumn{2}{c}{} & \multicolumn{4}{c}{\textbf{NeuPSL}} \\
        & \textbf{DeepStochlog} & \textbf{GCN} & \textbf{Energy} & \textbf{SP} & \textbf{MSE} & \textbf{BCE} \\
        \midrule
        \midrule
        \textbf{Citeseer} & $ 62.68 \pm 3.84 $ & $ 67.42 \pm 0.66 $ & $ 69.63 \pm 1.33 $ & $ \mathbf{69.78 \pm 1.42} $  & $ 69.62 \pm 1.27 $ & $ 69.64 \pm 1.33 $ \\
        \textbf{Cora} & $ 71.28 \pm 1.98 $ & $ 80.32 \pm 1.11 $ & $ 80.41 \pm 1.81 $ & $ 78.59 \pm 4.93 $ & $ \mathbf{81.48 \pm 1.45} $ & $ 81.28 \pm 1.45 $ \\
        \bottomrule
    \end{tabular}
    }
\end{table*}

\begin{table*}
    \caption{Accuracy of CNN, LTN, DeepProblog and NeuPSL on MNIST-Addition.}
    \vskip 0.1in
    \label{tab:deep_hl_mrf_mnist_add_prediction_performance}
    \centering
    \scalebox{0.9}{
    \begin{tabular}{lc||c|c|c||c||c}
        \toprule 
        & & \multicolumn{3}{c}{} & \multicolumn{2}{c}{\textbf{NeuPSL}} \\
        & \textbf{Additions} & \textbf{CNN} & \textbf{LTN} & \textbf{DeepProblog} & \textbf{Energy} & \textbf{BCE} \\
        \midrule
        \midrule
        \multirow{3}{*}{\textbf{MNIST-Add1}} & $ 300 $ & $17.16 \pm 00.62$ & $69.23 \pm 15.68$ & $85.61 \pm 01.28$ & $87.96 \pm 01.58$ & $\mathbf{88.84 \pm 02.07}$ \\
        & $ 3,000 $ & $ 78.99 \pm 01.14 $ & $93.90 \pm 00.51$ & $92.59 \pm 01.40$ & $95.60 \pm 0.91$ & $\mathbf{95.70 \pm 0.84}$ \\
        \hline
        \multirow{3}{*}{\textbf{MNIST-Add2}} & $ 150 $ & $01.31 \pm 00.23$ & $02.02 \pm 00.97$ & $71.37 \pm 03.90$ & $59.20 \pm 32.79$ & $\mathbf{76.00 \pm 2.61}$ \\
        & $ 1,500 $ & $01.69 \pm 00.27$ & $71.79 \pm 27.76$ & $87.44 \pm 02.15$ & $90.56 \pm 0.61$ & $\mathbf{93.04 \pm 2.26}$ \\
        \bottomrule
    \end{tabular}
    }
\end{table*}

\textbf{HL-MRF learning}
We first evaluate the prediction performance on non-deep variants of NeuPSL models for the first $7$ datasets, i.e., only symbolic weights are learned.
\tabref{tab:hl_mrf_prediction_performance} shows that across all $7$ datasets, NeuPSL models trained with \algoref{alg:nesy_ebm_learning} obtain a better average prediction performance than those trained using a valued-based loss. 
On the Cora dataset, the NeuPSL model fit with the BCE loss achieves over a $6\%$ point improvement over SP, the higher-performing value-based loss.
% This results shows \algoref{alg:nesy_ebm_learning} is applicable for improving traditional (Neu)PSL models without neural parameters.

\textbf{Deep HL-MRF learning}
Next, we evaluate the prediction performance of deep NeuPSL models.
Here, we study the standard low-data setting for Citeseer and Cora.
Specifically, results are averaged over $10$ randomly sampled splits using $5\%$ of the nodes for training, $5\%$ of the nodes for validation, and $1,000$ nodes for testing.
We also report the prediction performance of the same strong baseline models used in \citenoun{pryor:ijcai23} for this task: DeepStochLog \citep{winters:aaai22}, and a Graph Convolutional Network (GCN) \citep{kipf:iclr17}.
Additionally, we investigate performance on MNIST-Addition, a widely used NeSy evaluation task first introduced by \citenoun{manhaeve:neurips18}.
In MNIST-Addition, models must determine the sum of two lists of MNIST images, for example, $(\big[\inlinegraphics{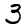}\big] + \big[\inlinegraphics{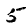}\big] = \mathbf{8})$.
The challenge stems from the lack of labels for the MNIST images; only the final sum of the equation is provided during training, $8$ in this example.
Implementation details for the neural and symbolic components of the NeuPSL models for both citation network and MNIST-Add experiments are provided in \appref{appendix:extended_learning_performance}.

\tabref{tab:deep_hl_mrf_citation_prediction_performance} shows that fitting the neural network weights of a NeuPSL model with our NeSy-EBM learning framework is effective.
NeuPSL models fit with the MSE and BCE losses consistently outperform both DeepStochlog and the GCN baseline.
Moreover, \tabref{tab:deep_hl_mrf_mnist_add_prediction_performance} demonstrates NeuPSL models trained with \algoref{alg:nesy_ebm_learning} and a BCE loss can achieve up to a $16\%$ point performance improvement over those trained with a value-based loss.

\section{Limitations}
\label{sec:limitations}

Our learning framework is limited to NeSy-EBMs satisfying the two assumptions made in \secref{sec:learning_framework}.
We do not explore methods for supporting NeSy-EBMs with non-differentiable value-functions.
One approach is to substitute the inference program with a principled approximation.
Lastly, although the idea to leverage inference algorithms such as BCD that effectively use warm-starts and improve learning runtimes is general, the inference algorithms were implemented for a NeSy system with an LCQP structure.

\section{Conclusions and future work}
\label{sec:conclusion}

We introduced a general learning framework for NeSy-EBMs and demonstrated its applicability with NeuPSL.
Additionally, we proposed a novel NeuPSL inference formulation and algorithm with practical and theoretical advantages.
A promising direction for future work is to extend the learning framework to support approximate inference solutions for estimating the objective gradient to further improve learning runtimes.
In addition, the empirical results presented in this work motivate generalizing and applying our learning framework to more NeSy systems and tasks.

\section*{Acknowledgements}
This work was partially supported by the National Science Foundation grants CCF-2023495 and CCF-2224213 and a Google Faculty Research Award.

\section*{Impact Statement}
% This paper presents work that advances the general field of machine learning. 
% Therefore, there are many established societal consequences of our work.
This paper presents work that advances the field of NeSy AI.
Therefore, our contributions further enable the integration of neural and symbolic systems and will be used to empower models with domain knowledge and symbolic reasoning. 
Potential positive societal consequences of our work include more accurate and reliable models that obey symbolic constraints.
More generally, our research advances the broader field of machine learning.
Thus, there are many established societal consequences of our work. 
However, we feel they do not need to be highlighted here.

\bibliography{dickens-icml24}
\bibliographystyle{icml2024}

%%%%%%%%%%%%%%%%%%%%%%%%%%%%%%%%%%%%%%%%%%%%%%%%%%%%%%%%%%%%%%%%%%%%%%%%%%%%%%%
%%%%%%%%%%%%%%%%%%%%%%%%%%%%%%%%%%%%%%%%%%%%%%%%%%%%%%%%%%%%%%%%%%%%%%%%%%%%%%%
% APPENDIX
%%%%%%%%%%%%%%%%%%%%%%%%%%%%%%%%%%%%%%%%%%%%%%%%%%%%%%%%%%%%%%%%%%%%%%%%%%%%%%%
%%%%%%%%%%%%%%%%%%%%%%%%%%%%%%%%%%%%%%%%%%%%%%%%%%%%%%%%%%%%%%%%%%%%%%%%%%%%%%%
\newpage
\onecolumn
\appendix

\section{Appendix}
\label{appendix:appendix}

This appendix includes the following sections: extended bilevel NeSy Learning framework, extended NeuPSL and deep hinge-loss Markov random fields, extended dual block coordinate descent, and extended empirical evaluation.

Code for running the experiments along with all data, models, and hyperparameters are available at: \url{https://github.com/linqs/dickens-icml24}.
Code for the NeuPSL implementation of our proposed learning framework and inference algorithms is available at: \url{https://github.com/linqs/psl}.

\section{Notation}
\label{appendix:notation}

\begin{table}[h!]
\centering
\caption{Summary of notation in main paper}
\scalebox{0.75}{
    \begin{tabular}{|c|l|}
        \hline
        \textbf{Symbol} & \textbf{Description} \\ 
        \hline
        \hline
        $\mathcal{W}_{nn}, \mathcal{W}_{sy}$ & NeSy-EBM neural and symbolic weight domain \\ \hline
        $\mathcal{Y}$ & NeSy-EBM target variable domain \\ \hline
        $\mathcal{X}_{nn}, \mathcal{X}_{sy}$ & NeSy-EBM neural and symbolic input domains\\ \hline
        $E: \mathcal{Y} \times \mathcal{X}_{sy} \times \mathcal{X}_{nn} \times \mathcal{W}_{sy} \times \mathcal{W}_{nn} \to \mathbb{R}$ & NeSy-EBM energy function \\ \hline
        $\mathcal{S} := \{ S_{1} := (\mathbf{y}_{1}, \mathbf{x}_{1, sy}, \mathbf{x}_{1, nn}), \cdots, S_{P} := (\mathbf{y}_{P}, \mathbf{x}_{P, sy}, \mathbf{x}_{P, nn})\} $ & NeSy-EBM training data \\ \hline
        $\mathcal{T}_{\mathcal{Y}}, \mathcal{Z}$ & Labeled and latent variable target partition domains \\ \hline
        $\mathbf{z}_{i}^{*} \in \argmin_{\mathbf{z} \in \mathcal{Z}}E((\mathbf{t}_{i}, \mathbf{z}), \mathbf{x}_{i, sy}, \mathbf{x}_{i, nn}, \mathbf{w}_{sy}, \mathbf{w}_{nn})$ & Latent variable minimizer \\ \hline
        $\mathbf{y}_{i}^{*} \in \argmin_{\mathbf{y} \in \mathcal{Y}} E(\mathbf{y}, \mathbf{x}_{i, sy}, \mathbf{x}_{i, nn}, \mathbf{w}_{sy}, \mathbf{w}_{nn})$ & Full minimizer \\ \hline
        $V_{\mathbf{z}^{*}_{i}}(\mathbf{w}_{sy}, \mathbf{w}_{nn}) := E((\mathbf{t}_{i}, \mathbf{z}^{*}_{i}), \mathbf{x}_{i, sy}, \mathbf{x}_{i, nn}, \mathbf{w}_{sy}, \mathbf{w}_{nn})$ & Latent optimal value function \\ \hline
        $V_{\mathbf{y}^{*}_{i}}(\mathbf{w}_{sy}, \mathbf{w}_{nn}) := E(\mathbf{y}^{*}_{i}, \mathbf{x}_{i, sy}, \mathbf{x}_{i, nn}, \mathbf{w}_{sy}, \mathbf{w}_{nn})$ & Full optimal value function \\ \hline
        $L_{Energy}(E(\cdot, \cdot, \cdot, \mathbf{w}_{sy}, \mathbf{w}_{nn}), S_{i}) :=  V_{\mathbf{z}^{*}_{i}}(\mathbf{w}_{sy}, \mathbf{w}_{nn})$ & Energy loss function \\ \hline
        $L_{SP}(E(\cdot, \cdot, \cdot, \mathbf{w}_{sy}, \mathbf{w}_{nn}), S_{i}) := V_{\mathbf{z}^{*}_{i}}(\mathbf{w}_{sy}, \mathbf{w}_{nn}) - V_{\mathbf{y}^{*}_{i}}(\mathbf{w}_{sy}, \mathbf{w}_{nn})$ & Structured perceptron loss function \\ \hline
        $L_{d}(E(\cdot, \cdot, \cdot, \mathbf{w}_{sy}, \mathbf{w}_{nn}), S_{i}) := d(\mathbf{y}_{i}^{*}(\mathbf{w}_{sy}, \mathbf{w}_{nn}), \mathbf{t}_{i})$ & Minimizer-based loss function \\ \hline
        $L_{Val}$ & A value-based loss function \\ \hline
        $\mathcal{R}: \mathcal{W}_{sy} \times \mathcal{W}_{nn} \to \mathbb{R}$ & Learning regularization function \\ \hline
        $\rho$ & A positive, scalar-valued, Moreau envelope parameter \\ \hline
        $M_{i}(\mathbf{y}; \mathbf{w}_{sy}, \mathbf{w}_{nn}, \rho)$ & Moreau envelope of the energy function with $\mathbf{x}_{i, sy}, \mathbf{x}_{i, nn}$ \\ \hline
        $\iota$ & Value-function formulation relaxation constant. \\ \hline
        $c_{i}(\mathbf{y}_{i}, \mathbf{w}_{sy}, \mathbf{w}_{nn}; \iota)$ & Equality constraint $i$ in \eqref{eq:relaxed_smoothed_value_function_bound_constrained_approach} \\ \hline
        $\mu, \mathbf{\lambda} := [ \lambda_{i} ]_{i = 1}^{P}$ & Quadratic and linear penalty parameters \\ \hline
        $\mathcal{L}_{A}(\mathbf{w}_{sy}, \mathbf{w}_{nn}, \mathbf{y}_{1}, \cdots, \mathbf{y}_{p}, \mathcal{S}; \mathbf{\lambda}, \mu, \iota)$ & The augmented Lagrangian function corresponding to \eqref{eq:relaxed_smoothed_value_function_bound_constrained_approach} \\ \hline
        $\mathbf{g} = [g_{i}]_{i = 1}^{n_{g}}$ & Deep HL-MRF Neural components \\ \hline
        $n_{y}, n_{x}, n_{g}$ & Dimensions of target, input, and neural outputs \\ \hline
        $\mathbf{a}_{\phi, \mathbf{y}} \in \mathbb{R}^{n_{y}}, \mathbf{a}_{\phi, \mathbf{x}} \in \mathbb{R}^{n_{x}}, \mathbf{a}_{\phi, \mathbf{g}} \in \mathbb{R}^{n_{g}}$ & Deep HL-MRF potential variable coefficient vectors \\ \hline
        $b_{\phi} \in \mathbb{R}$ & Deep HL-MRF potential constant vector \\ \hline
        $\phi(\mathbf{y}, \mathbf{x}_{sy}, \mathbf{g}(\mathbf{x}_{nn}, \mathbf{w}_{nn}))$ & A deep HL-MRF potential function\\ \hline
        $p \in \{1, 2\}$ & Deep HL-MRF potential exponent parameter \\ \hline
        $m$ & The number of deep HL-MRF potentials \\ \hline
        $\mathcal{T} = [\tau_i]_{i = 1}^{r}$ & Ordered partition of a set of $m$ deep HL-MRF potentials \\ \hline 
        $\mathbf{\Phi}(\mathbf{y} , \mathbf{x}_{sy}, \mathbf{g}(\mathbf{x}_{nn}, \mathbf{w}_{nn})) := [ \sum_{k \in \tau_i} \phi_{k}(\mathbf{y}, \mathbf{x}_{sy}, \mathbf{g}(\mathbf{x}_{nn}, \mathbf{w}_{nn})) ]_{i = 1}^{r}$ & Vector of $m$ deep HL-MRF potentials \\ \hline
        $E(\mathbf{y}, \mathbf{x}_{sy}, \mathbf{x}_{nn}, \mathbf{w}_{sy}, \mathbf{w}_{nn}) := \mathbf{w}_{sy}^T \mathbf{\Phi}(\mathbf{y} , \mathbf{x}_{sy}, \mathbf{g}(\mathbf{x}_{nn}, \mathbf{w}_{nn}))$ & A deep HL-MRF energy function \\ \hline
        $\mathbf{a}_{c_k, \mathbf{y}} \in \mathbb{R}^{n_{y}}, \mathbf{a}_{c_k, \mathbf{x}} \in \mathbb{R}^{n_{x}}, \mathbf{a}_{c_k, \mathbf{g}} \in \mathbb{R}^{n_{g}}$ & Deep HL-MRF constraint variable coefficient vectors \\ \hline
        $b_{c_k} \in \mathbb{R}$ & Deep HL-MRF constraint constant vector \\ \hline
        $\phi(\mathbf{y}, \mathbf{x}_{sy}, \mathbf{g}(\mathbf{x}_{nn}, \mathbf{w}_{nn}))$ & A deep HL-MRF potential function\\ \hline
        $\mathbf{\Omega}(\mathbf{x}_{sy}, \mathbf{g})$ & Deep HL-MRF feasible set \\ \hline
        $V(\mathbf{w}_{sy}, \mathbf{b}(\mathbf{x}_{sy}, \mathbf{g}(\mathbf{x}_{nn}, \mathbf{w}_{nn})))$ & Optimal value function of deep HL-MRF energy \\ \hline
        $\mathbf{\nu}$ & Primal variables of LCQP deep HL-MRF inference \\ \hline
        $A$ & Constraint matrix of LCQP deep HL-MRF inference \\ \hline
        $\mathbf{b}(\mathbf{x}_{sy}, \mathbf{g}(\mathbf{x}_{nn}, \mathbf{w}_{nn}))$ & Constraint constant vector of LCQP deep HL-MRF inference \\ \hline
        $\mathbf{D}(\mathbf{w}_{sy})$ & Objective matrix of LCQP deep HL-MRF inference \\ \hline
        $\mathbf{c}(\mathbf{w}_{sy})$ & Objective constant vector of LCQP deep HL-MRF inference \\ \hline
        $\epsilon$ & Scalar regularization parameter \\ \hline
        $\mathbf{\mu}$ & Dual variables of LCQP deep HL-MRF inference \\ \hline
        $h(\mathbf{\mu}; \mathbf{w}_{sy}, \mathbf{b}(\mathbf{w}_{nn}))$ & Dual objective of LCQP deep HL-MRF inference \\ \hline
    \end{tabular}
}
\label{table:notation}
\end{table}

\section{Extended Bilevel NeSy Learning Framework}
\label{appendix:extended_bilevel_nesy_learning}

In this section we provide the further details on our proposed NeSy learning framework.
A complete version of \algoref{alg:nesy_ebm_learning} is provided in \algoref{alg:full_nesy_ebm_learning}.

\begin{algorithm}[ht]
\caption{Full NeSy-EBM Learning Framework}
\label{alg:full_nesy_ebm_learning}
\begin{algorithmic}[1]
    \REQUIRE{Constraint Tolerance: $\sigma^*$, Movement Tolerance: $\omega^*$, Moreau Param.: $\rho$ \\  
    Starting points: $\mu^{(0)} > 1$, $\lambda_{1}^{(0)}, \cdots, \lambda_{P}^{(0)}$, $(\mathbf{w}^{(0)}_{sy}, \mathbf{w}^{(0)}_{nn}) \in \mathcal{W}_{sy} \times \mathcal{W}_{nn}$}
    \STATE{$\mathbf{y}_{i}^{(0)} \gets (\mathbf{t}_{i}, \mathbf{z}_{i}^{*}), \, \forall{i = 1, \cdots, P};$}
    \STATE{$\iota^{(0)} \gets \max_{i \in \{1, \cdots, P\}} M_{i}(\mathbf{y}_{i}^{(0)}; \mathbf{w}_{sy}^{(0)}, \mathbf{w}_{nn}^{(0)}, \rho) - V_{\mathbf{y}_{i}^{*}}(\mathbf{w}_{sy}^{(0)}, \mathbf{w}_{nn}^{(0)})$;}
    \FOR{$t = 0, 1, 2, \cdots$}
        \STATE{Set $\omega^{(0)} = \frac{1}{\mu^{(0)}}$, and $\sigma^{(0)} = \frac{1}{(\mu^{(0)})^{0.1}}$}
        \FOR{$k = 0, 1, 2, \cdots$}
            \STATE{Find $(\mathbf{w}^{(k)}_{sy}, \mathbf{w}^{(k)}_{nn}) \in \mathcal{W}_{sy} \times \mathcal{W}_{nn}$, $(\mathbf{y}_{1}^{(k)}, \cdots, \mathbf{y}_{P}^{(k)}) \in \mathcal{Y} \times \cdots \times \mathcal{Y}$, and $\mathbf{s}^{(k)} \in \mathbb{R}_{\geq 0}^{P}$ s.t.
            \begin{align*}
               \delta^{(k)} \gets \delta(\mathbf{w}^{(k)}_{sy}, \mathbf{w}^{(k)}_{nn}, \mathbf{y}^{(k)}_{1}, \cdots, \mathbf{y}^{(k)}_{p}, \mathbf{s}^{(k)}; \mathbf{\lambda}^{(k)}, \mu^{(k)}, \iota^{(k)}) \leq \omega^{(k)};
            \end{align*}}
            \IF{$\left(\sum_{i = 1}^{P} c_{i}(\mathbf{y}^{(k)}_{i}, \mathbf{w}^{(k)}_{sy}, \mathbf{w}^{(k)}_{nn}, \iota^{(k)}) + s_{i} \right) < \sigma^{(k)}$}
                \IF{$\left(\sum_{i = 1}^{P} c_{i}(\mathbf{y}^{(k)}_{i}, \mathbf{w}^{(k)}_{sy}, \mathbf{w}^{(k)}_{nn}, \iota^{(k)}) + s_{i} \right) < \sigma^{*}$ and $\delta^{(k)} \leq \omega^{*}$}
                    \STATE{Break with the approximate solution: $\mathbf{w}_{sy}^{(k)}, \mathbf{w}_{nn}^{(k)}, \mathbf{y}_{1}^{(k)}, \cdots, \mathbf{y}_{P}^{(k)}, \mathbf{s}^{(k)}$;}
                \ENDIF
                \STATE{$\lambda_{i}^{(k + 1)} \gets \lambda_{i}^{(k)} + \mu^{(k)} \left( c_{i}(\mathbf{y}^{(k)}_{i}, \mathbf{w}^{(k)}_{sy}, \mathbf{w}^{(k)}_{nn}, \iota^{(k)}) + s_{i} \right), \quad \forall{i = 1, \cdots, P}$; \\
                $\mu^{(k + 1)} \gets \mu^{(k)}$; \, 
                $\sigma^{(k + 1)} \gets \frac{\sigma^{(k)}}{(\mu^{(k + 1)})^{0.9}}$; \,
                $\omega^{(k + 1)} \gets \frac{\omega^{(k)}}{\mu^{(k + 1)}}$;}
            \ELSE
                \STATE{$\mu^{(k + 1)} \gets 2 \cdot \mu^{(k)}$; \, $\lambda_{i}^{(k + 1)} \gets \lambda_{i}^{(k)}, \, \forall{i = 1, \cdots, P}$;} \\
                \STATE{$\sigma^{(k + 1)} \gets \frac{1}{(\mu^{(k + 1)})^{0.1}}$; \,
                $\omega^{(k + 1)} \gets \frac{1}{\mu^{(k + 1)}}$;}
            \ENDIF
        \ENDFOR
        \IF{Stopping criterion satisified}
            \STATE{Stop with the approximate solution: $\mathbf{w}_{sy}^{(k)}, \mathbf{w}_{nn}^{(k)}, \mathbf{y}_{1}^{(k)}, \cdots, \mathbf{y}_{P}^{(k)}, \mathbf{s}^{(k)}$;}
        \ENDIF
        \STATE{$\mu^{(0)} \gets \mu^{(k)}; \, \lambda_{i}^{(0)} \gets \lambda_{i}^{(k)}, \, \forall{i = 1, \cdots, P};$}
        \STATE{$\iota^{(t + 1)} \gets \frac{1}{2} \cdot \iota^{(t)};$}
    \ENDFOR
    \end{algorithmic}
\end{algorithm} 

As stated in the main paper, each instance of \eqref{eq:relaxed_smoothed_value_function_bound_constrained_approach} is optimized using the bound constrained augmented Lagrangian algorithm, Algorithm 17.4 from \citenoun{nocedal:book06}.
This algorithm is applied in lines 4 through 16 in \algoref{alg:full_nesy_ebm_learning}.
The algorithm iteratively finds approximate minimizers of the problem's augmented Lagrangian, \eqref{eq:augmented_lagrangian}, for a fixed setting of the penalty parameters using randomized incremental gradient descent, line 6 in \algoref{alg:full_nesy_ebm_learning}.
Specifically, gradient descent is applied to find an approximate minimizer of \eqref{eq:augmented_lagrangian} satisfying the following stopping criterion:
\begin{align}
    & \delta(\mathbf{w}_{sy}, \mathbf{w}_{nn}, \mathbf{y}_{1}, \cdots, \mathbf{y}_{p}, \mathbf{s}; \mathbf{\lambda}, \mu, \iota) := \nonumber \\
    & \quad \left \Vert \mathbf{w}_{sy} - \Pi \left( \mathbf{w}_{sy} - \nabla_{\mathbf{w}_{sy}} \mathcal{L}_{A} \right) \right \Vert + \left \Vert \mathbf{w}_{nn} - \Pi \left( \mathbf{w}_{nn} - \nabla_{\mathbf{w}_{nn}}  \mathcal{L}_{A} \right) \right \Vert \nonumber \\
    & \quad + \sum_{i=1}^{P} \left \Vert \mathbf{y}_{i} - \Pi \left( \mathbf{y}_{i} - \nabla_{\mathbf{y}_{i}} \mathcal{L}_{A} \right) \right \Vert + \left \Vert \mathbf{s} - \Pi \left( \mathbf{s} - \nabla_{\mathbf{s}} \mathcal{L}_{A} \right) \right \Vert \leq \omega,
\end{align}
where $\omega > 0$ is a positive tolerance that is updated with the Lagrange variables.
Further, note the Lagrangian gradients are evaluated at the iterate specified as arguments of $\delta$. 
Practically, the parameter movement between an epoch of incremental gradient descent is used to approximate $\delta$.

As stated in the main paper, employing the bound constrained augmented Lagrangian algorithm to solve the instances of \eqref{eq:relaxed_smoothed_value_function_bound_constrained_approach} ensures fundamental convergence properties of our learning framework.
Specifically, theorem 17.2 in \cite{nocedal:book06} is applicable to \algoref{alg:full_nesy_ebm_learning}.
This theorem states that limit points of the iterate sequence are stationary points of $\Vert c(\mathbf{y}_{1}, \cdots, \mathbf{y}_{P}, \mathbf{w}_{sy}, \mathbf{w}_{nn})) + \mathbf{s} \Vert^{2}$ when they are infeasible or, when the LICQ holds and the iterates are feasible, are KKT points of \eqref{eq:relaxed_smoothed_value_function_bound_constrained_approach}.

% For completeness we provide the definition of the LICQ, adapted from \citenoun{nocedal:book06}, here.
% \begin{definition}[Active Set and Linear Independence Constraint Qualification (LICQ)]
%     Consider the general constrained optimization problem:
%     \begin{align*}
%         & \min_{\mathbf{x} \in \mathbb{R}^{n}} f(\mathbf{x}) \\
%         & \textrm{s.t.} \quad c_{i} (\mathbf{x}) = 0, \quad \forall i \in \mathcal{E}, \\
%         & \quad \quad c_{i} (\mathbf{x}) \geq 0, \quad \forall i \in \mathcal{I},
%     \end{align*}
%     where $f$ and all $c_{i}$ are real-valued functions on $\mathbb{R}^{n}$ and $\mathcal{E}$ and $\mathcal{I}$ are two finite sets of indices.

%     The \textbf{active set} $\mathcal{A}(\mathbf{x})$ at any feasible point $\mathbf{x}$ consists of the equality constraint indices from $\mathcal{E}$ together with the indices of the inequality constraints $i$ for which $c_{i}(\mathbf{x}) = 0$; that is,
%     \begin{align}
%         \mathcal{A}(\mathbf{x}) := \mathcal{E} \cup \{i \in \mathcal{I} \, \vert \, c_{i}(\mathbf{x}) = 0 \}.
%     \end{align}

%     Given the point $\mathbf{x}$ and the active set $\mathcal{A}(\mathbf{w})$, the \textbf{linear independence constraint qualification (LICQ)} holds if the set of active constraint gradients $\{ \nabla c_{i}(\mathbf{x}), \, i \in \mathcal{A}(\mathbf{x}) \}$ is linearly independent.
% \end{definition}

\section{Extended NeuPSL and deep hinge-loss Markov random fields}
\label{appendix:extended_neupsl_deep_hlmrfs}

In this section, we expand on the smooth formulation of NeuPSL inference and provide proofs for the continuity results presented in \secref{sec:continuity_of_inference}.

\subsection{Extended smooth formulation of inference}
\label{appendix:lcqp_inference}
Recall the primal formulation of NeuPSL inference restated below:
\begin{align}
    \argmin_{\mathbf{y} \in \mathbb{R}^{n_{\mathbf{y}}}} \, \mathbf{w}_{sy}^T \mathbf{\Phi}(\mathbf{y} , \mathbf{x}_{sy}, \mathbf{g}(\mathbf{x}_{nn}, \mathbf{w}_{nn})) \quad
        \textrm{s.t. }        
        \mathbf{y} \in \mathbf{\Omega}(\mathbf{x}_{sy}, \mathbf{g}(\mathbf{x}_{nn}, \mathbf{w}_{nn}))
    \label{eq:inference_primal_app}.
\end{align}
Importantly, note the structure of the deep hinge-loss potentials defining $\mathbf{\Phi}$:
\begin{equation}
        \phi_k(\mathbf{y}, \mathbf{x}_{sy}, \mathbf{g}(\mathbf{x}_{nn}, \mathbf{w}_{nn})) := (\max\{\mathbf{a}_{\phi_k, y}^T \mathbf{y} + \mathbf{a}_{\phi_k, \mathbf{x}_{sy}}^T \mathbf{x}_{sy} + \mathbf{a}_{\phi_k, \mathbf{g}}^T \mathbf{g}(\mathbf{x}_{nn}, \mathbf{w}_{nn}) + b_{\phi_k}, 0\})^{p_{k}}.
        \label{eq:deep_hlmrf_potential_app}
\end{equation}
The LCQP NeuPSL inference formulation is defined using ordered index sets: $\mathbf{I}_{S}$  for the partitions of squared hinge potentials (indices $k$ which for all $j \in t_{k}$ the exponent term $p_j=2$) and $\mathbf{I}_{L}$ for the partitions of linear hinge potentials (indices $k$ which for all $j \in t_{k}$ the exponent term $p_j=1$).
With the index sets, we define 
\begin{align}
    \mathbf{W}_{S} :=
        \begin{bmatrix} 
            w_{\mathbf{I}_{S}[1]} \mathbf{I} & 0 & \cdots & 0 \\
            0 & w_{\mathbf{I}_{S}[2]} \mathbf{I} & &  \\
            \vdots & & \ddots & 
        \end{bmatrix} 
   \quad \quad \textrm{and} \quad \quad
    \mathbf{w}_{L} := 
        \begin{bmatrix} 
            w_{\mathbf{I}_{L}[1]} \mathbf{1} \\
           w_{\mathbf{I}_{L}[2]} \mathbf{1} \\
            \vdots 
        \end{bmatrix} 
    \label{eq:symbolic_weight_matrix_and_vector}
\end{align}
Let $m_{S} := \vert \cup_{\mathbf{I}_{S}} t_{k} \vert$ and $m_{L} := \vert \cup_{\mathbf{I}_{L}} t_{k} \vert$, be the total number of squared and linear hinge potentials, respectively, and define slack variables $\mathbf{s}_{S} := [s_{j}]_{j = 1}^{m_{S}}$ and $\mathbf{s}_{L} := [s_{j}]_{j = 1}^{m_{L}}$ for each of the squared and linear hinge potentials, respectively.
NeuPSL inference is equivalent to the following LCQP:
{
\begin{subequations}  
\label{eq:lcqp}
\begin{align}
\label{eq:lcqp.1}
        &\min_{\mathbf{y} \in [0, 1]^{n_{y}}, \, \mathbf{s}_{S} \in \mathbb{R}^{m_{S}}, \, \mathbf{s_H} \in \mathbb{R}^{m_{L}}_{+}} \,  
        \mathbf{s}_{S}^T \mathbf{W}_{S} \mathbf{s}_{S} + \mathbf{w}_{L}^T \mathbf{s}_{L} \\
        \label{eq:lcqp.2}
        & \quad \quad\textrm{s.t.} \,         
        \quad  \mathbf{a}_{c_i, \mathbf{y}}^T \mathbf{y} + \mathbf{a}_{c_i, \mathbf{x}_{sy}}^T \mathbf{x}_{sy} + \mathbf{a}_{c_i, \mathbf{g}}^T \mathbf{g}(\mathbf{x}_{nn}, \mathbf{w}_{nn}) + b_{c_i} \leq 0 \quad \forall \, i =1,\dotsc,q, \\
        \label{eq:lcqp.3}
        & \quad \quad \quad \mathbf{a}_{\phi_j, \mathbf{y}}^T \mathbf{y} + \mathbf{a}_{\phi_j, \mathbf{x}_{sy}}^T \mathbf{x}_{sy} + \mathbf{a}_{\phi_j, \mathbf{g}}^T \mathbf{g}(\mathbf{x}_{nn}, \mathbf{w}_{nn}) + b_{\phi_j} - s_j \leq 0 \quad \forall j \in I_{S} \cup I_{L}.
\end{align}       
\end{subequations}
}%
We ensure strong convexity by adding a square regularization with parameter $\epsilon$ to the objective.
Let the bound constraints on $\mathbf{y}$ and $\mathbf{s}_{L}$ and linear inequalities in the LCQP be captured by the $(2 \cdot n_{y} + q + m_{S} + 2 \cdot m_{L}) \times (n_{y} + m_{S} + m_{L})$ matrix $\mathbf{A}$ and $(2 \cdot n_{y} + q + m_{S} + m_{L})$ dimensional vector $\mathbf{b}(\mathbf{x}_{sy}, \mathbf{g}(\mathbf{x}_{nn}, \mathbf{w}_{nn}))$.
More formally, $\mathbf{A} := [a_{ij}]$ where $a_{ij}$ is the coefficient of a decision variable in the implicit and explicit constraints in the formulation above:
\begin{align}
    a_{i, j} := 
    \begin{cases}
        0 & (i \leq q) \, \land \, (j \leq m_{S} + m_{L}) \\
        \mathbf{a}_{c_{i}, \mathbf{y}}[j - (m_{S} + m_{L})] & (i \leq q) \, \land \, (j > m_{S} + m_{L}) \\
        0 & (q < i \leq q + m_{S} + m_{L}) \, \land \, (j \leq m_{S} + m_{L}) \land (j \neq i - q) \\
        -1 & (q < i \leq q + m_{S} + m_{L}) \, \land \, (j \leq m_{S} + m_{L}) \land (j = i - q) \\
        \mathbf{a}_{\phi_{i - q}, \mathbf{y}}[j - (m_{S} + m_{L})] & (q < i \leq q + m_{S} + m_{L}) \, \land \, (j > m_{S} + m_{L}) \\
        0 & (q + m_{S} + m_{L} < i \leq q + m_{S} + 2 \cdot m_{L} + n_{y}) \, \\ & \land \, (j \neq i - (q + m_{L})) \\
        -1 & (q + m_{S} + m_{L} < i \leq q + m_{S} + 2 \cdot m_{L} + n_{y}) \, \\ & \land \, (j = i - (q + m_{L})) \\
        0 & (q + m_{S} + 2 \cdot m_{L} + n_{y} < i \leq q + m_{S} + 2 \cdot m_{L} + 2 \cdot n_{y}) \, \\ & \land \, (j \neq i - (q + m_{S} + m_{L})) \\
        1 & (q + m_{S} + 2 \cdot m_{L} + n_{y} < i \leq q + m_{S} + 2 \cdot m_{L} + 2 \cdot n_{y}) \, \\ & \land \, (j = i - (q + m_{S} + m_{L}))
    \end{cases}.
\end{align}
Furthermore, $\mathbf{b}(\mathbf{x}_{sy}, \mathbf{g}(\mathbf{x}_{nn}, \mathbf{w}_{nn})) = [b_{i}(\mathbf{x}_{sy}, \mathbf{g}(\mathbf{x}_{nn}, \mathbf{w}_{nn}))]$ is the vector of constants corresponding to each constraint in the formulation above:
\begin{align}
    & b_{i}(\mathbf{x}_{sy}, \mathbf{g}(\mathbf{x}_{nn}, \mathbf{w}_{nn})) \\
    & \quad \quad := 
    \begin{cases}
        \mathbf{a}_{c_{i}, \mathbf{x}_{sy}}^T \mathbf{x}_{sy} + \mathbf{a}_{c_{i}, \mathbf{g}}^T \mathbf{g}(\mathbf{x}_{nn}, \mathbf{w}_{nn}) + b_{c_i} & i \leq q \\
        \mathbf{a}_{\phi_{i - q}, \mathbf{x}_{sy}}^T \mathbf{x}_{sy} + \mathbf{a}_{\phi_{i - q}, \mathbf{g}}^T \mathbf{g}(\mathbf{x}_{nn}, \mathbf{w}_{nn}) + b_{\phi_{i - q}} & q < i \leq q + m_{S} + m_{L} \\
        0 & q + m_{S} + m_{L} < i \\ & \quad \leq q + m_{S} + 2 \cdot m_{L} + n_{y} \\
        -1 & q + m_{S} + 2 \cdot m_{L} + n_{y} < i \\ & \quad \leq q + m_{S} + 2 \cdot m_{L} + 2 \cdot n_{y}
    \end{cases}.
\end{align}
Note that $\mathbf{b}(\mathbf{x}_{sy}, \mathbf{g}(\mathbf{x}_{nn}, \mathbf{w}_{nn}))$ is a linear function of the neural network outputs, hence, if $\mathbf{g}(\mathbf{x}_{nn}, \mathbf{w}_{nn})$ is a smooth function of the neural parameters, then $\mathbf{b}(\mathbf{x}_{sy}, \mathbf{g}(\mathbf{x}_{nn}, \mathbf{w}_{nn}))$ is also smooth.

With this notation, the regularized inference problem is:
{
\begin{align}
      V(\mathbf{w}_{sy}, \mathbf{b}(\mathbf{x}_{sy}, \mathbf{g}(\mathbf{x}_{nn}, \mathbf{w}_{nn}))) :=  \min_{\mathbf{y}, \mathbf{s_{S}}, \mathbf{s_H}} \, &
        \begin{bmatrix}
            \mathbf{s}_{S} \\ \mathbf{s}_{L} \\ \mathbf{y}
        \end{bmatrix}^T
        \begin{bmatrix}
            \mathbf{W}_{S} + \epsilon I & 0 & 0\\
            0 & \epsilon I & 0 \\
            0 & 0 & \epsilon I \\
        \end{bmatrix}
        \begin{bmatrix}
            \mathbf{s}_{S} \\
            \mathbf{s}_{L} \\
            \mathbf{y}  
        \end{bmatrix} 
        + 
        \begin{bmatrix}
           0 \\ \mathbf{w}_{L} \\ 0
        \end{bmatrix}^T
        \begin{bmatrix}
            \mathbf{s}_{S} \\
            \mathbf{s}_{L} \\
            \mathbf{y}
        \end{bmatrix} \nonumber \\
        \textrm{s.t.} \quad         
        & \mathbf{A}         
        \begin{bmatrix}
            \mathbf{s}_{S} \\
            \mathbf{s}_{L} \\
            \mathbf{y}
        \end{bmatrix} + \mathbf{b}(\mathbf{x}_{sy}, \mathbf{g}(\mathbf{x}_{nn}, \mathbf{w}_{nn})) \leq 0 \label{eq:lcqp_primal_appendix}.
\end{align}
}%
For ease of notation, let 
{
\begin{align}
    D(\mathbf{w}_{sy}) :=         
    \begin{bmatrix}
        \mathbf{W}_{S} & 0 & 0\\
        0 & 0 & 0 \\
        0 & 0 & 0
    \end{bmatrix},
    \,
    \mathbf{c}(\mathbf{w}_{sy}) :=         
    \begin{bmatrix}
        0 \\
        \mathbf{w}_{L} \\
        0 
    \end{bmatrix}, 
    \,
    \mathbf{\nu} :=         
    \begin{bmatrix}
        \mathbf{s}_{S} \\
        \mathbf{s}_{L} \\
        \mathbf{y}
    \end{bmatrix}.
\end{align} 
}%
Then the regularized primal LCQP MAP inference problem is concisely expressed as
{
\begin{align}
    \min_{\mathbf{\nu} \in \mathbb{R}^{n_{\mathbf{y}} + m_{S} + m_{L}}} & \, 
        \mathbf{\nu}^T (\mathbf{D}(\mathbf{w}_{sy}) + \epsilon \mathbf{I}) \mathbf{\nu} + \mathbf{c}(\mathbf{w}_{sy})^T \mathbf{\nu} \label{eq:regularized_lcqp_primal_appendix}\\
        \textrm{s.t.} \quad         
        & \mathbf{A} \mathbf{\nu} + \mathbf{b}(\mathbf{x}_{sy}, \mathbf{g}(\mathbf{x}_{nn}, \mathbf{w}_{nn})) \leq 0 \nonumber.
\end{align} 
}%

By Slater's constraint qualification, we have strong-duality when there is a feasible solution.
In this case, an optimal solution to the dual problem yields an optimal solution to the primal problem.
The Lagrange dual problem of \eqref{eq:regularized_lcqp_primal_appendix} is
{
\begin{align}
    & \argmax_{\mathbf{\mu}\ge 0} \min_{\mathbf{\nu} \in \mathbb{R}^{n_{\mathbf{y}} + m_{S} + m_{L}}} \, 
        \mathbf{\nu}^T
        (\mathbf{D}(\mathbf{w}_{sy}) + \epsilon \mathbf{I})
        \mathbf{\nu}
        + 
        \mathbf{c}(\mathbf{w}_{sy})^T \mathbf{\nu}
        +
        \mathbf{\mu}^T (\mathbf{A} \mathbf{\nu}
        + \mathbf{b}(\mathbf{x}_{sy}, \mathbf{g}(\mathbf{x}_{nn}, \mathbf{w}_{nn}))) \nonumber \\
    & \quad = \argmax_{\mathbf{\mu}\ge 0}
        \,
        - \frac{1}{4} \mathbf{\mu}^T \mathbf{A} (\mathbf{D}(\mathbf{w}_{sy}) + \epsilon \mathbf{I})^{-1} \mathbf{A}^T \mathbf{\mu}  \label{eq:dual_lcqp_appendix} \\ 
        & \quad \quad \quad \quad \quad \quad \quad - \frac{1}{2} (\mathbf{A} (\mathbf{D}(\mathbf{w}_{sy}) + \epsilon \mathbf{I})^{-1} \mathbf{c}(\mathbf{w}_{sy}) - 2 \mathbf{b}(\mathbf{x}_{sy}, \mathbf{g}(\mathbf{x}_{nn}, \mathbf{w}_{nn})))^T \mathbf{\mu} \nonumber
\end{align}
}%
where $\mathbf{\mu} = [\mu_{i}]_{i = 1}^{n_{\mu}}$ are the Lagrange dual variables.
For later reference, denote the negative of the Lagrange dual function of MAP inference as: 
\begin{align}
    & h(\mathbf{\mu}; \mathbf{w}_{sy}, \mathbf{b}(\mathbf{x}_{sy}, \mathbf{g}(\mathbf{x}_{nn}, \mathbf{w}_{nn}))) \label{eq:lcqp_lagrangian_appendix} \\
        & := \frac{1}{4} \mathbf{\mu}^T \mathbf{A} (\mathbf{D}(\mathbf{w}_{sy}) + \epsilon \mathbf{I})^{-1} \mathbf{A}^T \mathbf{\mu} + \frac{1}{2} (\mathbf{A} (\mathbf{D}(\mathbf{w}_{sy}) + \epsilon \mathbf{I})^{-1} \mathbf{c}(\mathbf{w}_{sy}) - 2 \mathbf{b}(\mathbf{x}_{sy}, \mathbf{g}(\mathbf{x}_{nn}, \mathbf{w}_{nn})))^T \mathbf{\mu}. \nonumber
\end{align}
The dual LCQP has more decision variables but is only over non-negativity constraints rather than the complex polyhedron feasible set.
The dual-to-primal variable translation is:
\begin{align}
    \mathbf{\nu} = - \frac{1}{2} (\mathbf{D}(\mathbf{w}_{sy}) + \epsilon \mathbf{I})^{-1} (\mathbf{A}^T \mathbf{\mu} + \mathbf{c}(\mathbf{w}_{sy}))
    \label{eq:dual_primal_translation_appendix}
\end{align}
As $(\mathbf{D}(\mathbf{w}_{sy}) + \epsilon \mathbf{I})$ is diagonal, it is easy to invert and hence it is practical to work in the dual space to obtain a solution to the primal problem.

\subsection{Extended continuity of inference}
\label{appendix:continuity_of_inference}

We now provide background on sensitivity analysis that we then apply in our proofs on the continuity properties of NeuPSL inference.

\subsubsection{Background}
\label{appendix:continuity_of_inference_preliminaries}

\begin{theorem}[\cite{boyd:book04} p. 81]
    \label{thm:pointwise_supremum_over_convex}
    If for each $\mathbf{y} \in \mathcal{A}$, $f(\mathbf{x}, \mathbf{y})$ is convex in $\mathbf{x}$ then the function
    \begin{align}
        g(\mathbf{x}) := \sup_{\mathbf{y} \in \mathcal{A}} f(\mathbf{x}, \mathbf{y})
    \end{align}
    is convex in $\mathbf{x}$.
\end{theorem}

\begin{theorem}[\cite{boyd:book04} p. 81]
    \label{thm:pointwise_infimum_over_concave}
    If for each $\mathbf{y} \in \mathcal{A}$, $f(\mathbf{x}, \mathbf{y})$ is concave in $\mathbf{x}$ then the function
    \begin{align}
        g(\mathbf{x}) := \inf_{\mathbf{y} \in \mathcal{A}} f(\mathbf{x}, \mathbf{y})
    \end{align}
    is concave in $\mathbf{x}$.
\end{theorem}

\begin{definition}[Convex Subgradient: \cite{boyd:book04} and \cite{shalevshwartz:ftml11}]
    Consider a convex function $f: \mathbb{R}^{n} \to [-\infty, \infty]$ and a point $\overline{\mathbf{x}}$ with $f(\overline{\mathbf{x}})$ finite.
    For a vector $\mathbf{v} \in \mathbf{R}^{n}$, one says that $\mathbf{v}$ is a (convex) subgradient of $f$ at $\overline{\mathbf{x}}$, written $\mathbf{v} \in \partial f(\overline{\mathbf{x}})$, iff
    \begin{align}
        f(\mathbf{x}) \geq f(\overline{\mathbf{x}}) + <\mathbf{v}, \mathbf{x} - \overline{\mathbf{x}}>, \quad \forall \mathbf{x} \in \mathbf{R}^{n}.
    \end{align}
\end{definition}

\begin{definition}[Closedness: \citenoun{bertsekas:book09}]
    If the epigraph of a function $f : \mathbb{R}^{n} \to [-\infty, \infty]$ is a closed set, we say that $f$ is a closed function.
\end{definition}

\begin{definition}[Lower Semicontinuity: \cite{bertsekas:book09}]
    The function $f : \mathbb{R}^{n} \to [-\infty, \infty]$ is \textit{lower semicontinuous} (lsc) at a point $\overline{\mathbf{x}} \in \mathbb{R}^{n}$ if
    \begin{align}
        f(\overline{\mathbf{x}}) \leq \liminf_{k \to \infty} f(\mathbf{x}_{k}),
    \end{align}
    for every sequence $\{\mathbf{x}_{k}\} \subset \mathbb{R}^{n}$ with $\mathbf{x}_{k} \to \overline{\mathbf{x}}$.
    We say $f$ is \textit{lsc} if it is lsc at each $\overline{\mathbf{x}}$ in its domain.
\end{definition}

\begin{theorem}[Closedness and Semicontinuity: \cite{bertsekas:book09} Proposition 1.1.2.]
    For a function $f: \mathbb{R}^{n} \to [-\infty, \infty]$, the following are equivalent:
    \begin{enumerate}
        \item The level set $V_{\gamma} = \{\mathbf{x} \, \vert \, f(\mathbf{x}) \leq \gamma \}$ is closed for every scalar $\gamma$.
        \item $f$ is lsc.
        \item $f$ is closed.
    \end{enumerate}
\end{theorem}

The following definition and theorem are from \cite{rockafellar:book97} and they generalize the notion of subgradients to non-convex functions and the chain rule of differentiation, respectively.
For complete statements see \cite{rockafellar:book97} \cite{rockafellar:book97}.

\begin{definition}[Regular Subgradient: \cite{rockafellar:book97} Definition 8.3]
    Consider a function $ f: \mathbb{R}^{n} \to [-\infty, \infty] $ and a point $\overline{\mathbf{x}}$ with $f(\overline{\mathbf{x}})$ finite.
    For a vector $\mathbf{v} \in \mathbf{R}^{n}$, one says that $\mathbf{v}$ is a regular subgradient of $f$ at $\overline{\mathbf{x}}$, written $\mathbf{v} \in \hat{\partial} f(\overline{\mathbf{x}})$, iff
    \begin{align}
        f(\mathbf{x}) \geq f(\overline{\mathbf{x}}) + \langle \mathbf{v}, \mathbf{x} - \overline{\mathbf{x}} \rangle + \textrm{o}(\mathbf{x} - \overline{\mathbf{x}}), \quad \forall \mathbf{x} \in \mathbf{R}^{n},
    \end{align}
    where the $\textrm{o}(t)$ notation indicates a term with the property that 
    \begin{align}
        \lim_{t \to 0} \frac{\textrm{o}(t)}{t} = 0.
    \end{align}
\end{definition}

The relation of the regular subgradient defined above and the more familiar convex subgradient is the addition of the $o(\mathbf{x} - \mathbf{\overline{x}})$ term.
Evidently, a convex subgradient is a regular subgradient.

\begin{theorem}[Chain Rule for Regular Subgradients: \cite{rockafellar:book97} Theorem 10.6]
\label{theorem:subgradient_chain_rule}
Suppose $f(\mathbf{x}) = g(F(\mathbf{x}))$ for a proper, lsc function $g : \mathbb{R}^{m} \to [-\infty, \infty]$ and a smooth mapping $F : \mathbb{R}^{n} \to \mathbb{R}^{m}$.
Then at any point $\overline{\mathbf{x}} \in \textrm{dom} \, f = F^{-1} (\textrm{dom} \, g)$ one has
\begin{align}
    \hat{\partial} f(\overline{\mathbf{x}}) & \supset \nabla F(\overline{\mathbf{x}}) ^T \hat{\partial} g(F(\overline{\mathbf{x}})),
\end{align}
where $\nabla F(\overline{\mathbf{x}}) ^T$ is the Jacobian of $F$ at $\overline{\mathbf{x}}$.
\end{theorem}

\begin{theorem}[Danskin's Theorem: \cite{danskin:siam66} and \cite{bertsekas:phd71} Proposition A.22]
    \label{thm:danskin}
    Suppose $\mathcal{Z} \subseteq \mathbb{R}^{m}$ is a compact set and $g(\mathbf{x}, \mathbf{z}): \mathbb{R}^{n} \times \mathcal{Z} \to (-\infty, \infty]$ is a function.
    Suppose $g(\cdot, \mathbf{z}): \mathbb{R}^{n} \to \mathbb{R}$ is closed proper convex function for every $\mathbf{z} \in \mathcal{Z}$.
    Further, define the function $f: \mathbb{R}^{n} \to \mathbb{R}$ such that
    \begin{align}
        f(\mathbf{x}) := \max_{\mathbf{z} \in \mathcal{Z}} g(\mathbf{x}, \mathbf{z}). \nonumber
    \end{align}
    Suppose $f$ is finite somewhere.
    Moreover, let $\mathcal{X} := \textrm{int}(\textrm{dom}f)$, i.e., the interior of the set of points in $\mathbb{R}^{n}$ such that $f$ is finite.
    Suppose $g$ is continuous on $\mathcal{X} \times \mathcal{Z}$. 
    Further, define the set of maximizing points of $g(\mathbf{x}, \cdot)$ for each $\mathbf{x}$
    \begin{align}
        Z(\mathbf{x}) = \argmax_{\mathbf{z} \in \mathcal{Z}} g(\mathbf{x}, \mathbf{z}). \nonumber
    \end{align}
    Then the following properties of $f$ hold.
    \begin{enumerate}
        \item The function $f(\mathbf{x})$ is a closed proper convex function.
        \item For every $\mathbf{x} \in \mathcal{X}$,
        \begin{align}
            \partial f(\mathbf{x}) = \textrm{conv}\left \{ 
                \partial_{\mathbf{x}} g(\mathbf{x}, \mathbf{z}) \, \vert \, \mathbf{z} \in Z(\mathbf{x})
            \right \}.
        \end{align}
    \end{enumerate}
\end{theorem}

\begin{corollary}
    \label{corollary:danskins_unique_solution}
    Assume the conditions for Danskin's Theorem above hold.
    For every $\mathbf{x} \in \mathcal{X}$, if $Z(\mathbf{x})$ consists of a unique point, call it $\mathbf{z}^{*}$, and $g(\cdot, \mathbf{z}^{*})$ is differentiable at $\mathbf{x}$, then $f(\cdot)$ is differentiable at $\mathbf{x}$, and 
    \begin{align}
        \nabla f(\mathbf{x}) := \nabla_{\mathbf{x}} g(\mathbf{x}, \mathbf{z}^{*}).
    \end{align}
\end{corollary}

\begin{theorem}[\cite{bonnans:siam98} Theorem 4.2, \cite{rockafellar:rcsam74} p. 41]
    \label{thm:value_function_subdifferentiable_wrt_constraint_constants}
    Let $\mathbf{X}$ and $\mathbf{U}$ be Banach spaces.
    Let $\mathbf{K}$ be a closed convex cone in the Banach space $\mathbf{U}$.
    Let $G: \mathbf{X} \to \mathbf{U}$ be a convex mapping with respect to the cone $\mathbf{C} := -\mathbf{K}$ and $f: \mathbf{X} \to (-\infty, \infty]$ be a (possibly infinite-valued) convex function.
    Consider the following convex program and its optimal value function:
    \begin{align*}
        v_{P}(\mathbf{u}) := \min_{\mathbf{x} \in \mathbf{X}} & \, \, f(\mathbf{x}) & \textrm{(P)} \\
        \textit{s.t.} & \quad G(\mathbf{x}) + \mathbf{u} \in \mathbf{K}. \nonumber
    \end{align*}
    Moreover, consider the (Lagrangian) dual of the program:
    \begin{align*}
        v_{D}(\mathbf{u}) := \max_{\mathbf{\lambda} \in \mathbf{K}^{-}} & \, \, \min_{\mathbf{x} \in \mathbf{X}} f(\mathbf{x}) + \mathbf{\lambda}^T (G(\mathbf{x}) + \mathbf{u}) & \textrm{(D)} 
    \end{align*}
    Suppose $v_{P}(\mathbf{0})$ is finite.
    Further, suppose the feasible set of the program is nonempty for all $\mathbf{u}$ in a neighborhood of $\mathbf{0}$, i.e.,
    \begin{align}
        \label{eq:cone_constraint_qualification}
        \mathbf{0} \in \textrm{int} \{ G(\mathbf{X}) - \mathbf{K} \}.
    \end{align}
    Then, 
    \begin{enumerate}
        \item There is no primal dual gap at $u = 0$, i.e., $v_{P}(0) = v_{D}(0)$.
        \item The set, $\Lambda_{0}$, of optimal solutions to the dual problem with $\mathbf{u} = 0$ is non-empty and bounded.
        \item The optimal value function $v_{P}(\mathbf{u})$ is continuous at $\mathbf{u} = 0$ and $\partial v_{P}(\mathbf{0}) = \Lambda_{0}$.
    \end{enumerate}
\end{theorem}

\begin{theorem}[\cite{bonnans:book00} Proposition 4.3.2]
    \label{thm:bonnans-shapiro-00}
    Consider two optimization problems over a non-empty feasible set $\mathbf{\Omega}$:
    \begin{align}
        \min_{\mathbf{x} \in \mathbf{\Omega}} f_{1}(\mathbf{x})
        \quad \quad \textrm{and} \quad \quad
        \min_{\mathbf{x} \in \mathbf{\Omega}} f_{2}(\mathbf{x})
    \end{align}
    where $f_{1}, f_{2}: \mathcal{X} \to \mathbb{R}$.
    Suppose $f_{1}$ has a non-empty set $\mathbf{S}$ of optimal solutions over $\mathbf{\Omega}$.
    Suppose the second order growth condition holds for $\mathbf{S}$, i.e., there exists a neighborhood $\mathcal{N}$ of $\mathbf{S}$ and a constant $\alpha > 0$ such that
    \begin{align}
        f_{1}(\mathbf{x}) \geq f_{1}(\mathbf{S}) + \alpha (dist(\mathbf{x}, \mathbf{S}))^2,
        \quad \quad \forall \mathbf{x} \in \mathbf{\Omega} \cap \mathcal{N},
    \end{align}
    where $f_{1}(\mathbf{S}) := \textrm{inf}_{\mathbf{x} \in \mathbf{\Omega}} f_{1}(\mathbf{x})$.
    Define the difference function:
    \begin{align}
        \Delta(\mathbf{x}) := f_{2}(\mathbf{x}) - f_{1}(\mathbf{x}).
    \end{align}
    Suppose $\Delta(\mathbf{x})$ is $L$-Lipschitz continuous on $\mathbf{\Omega} \cap \mathcal{N}$.
    Let $\mathbf{x}^{*} \in \mathcal{N}$ be an $\delta$-solution to the problem of minimizing $f_{2}(\mathbf{x})$ over $\mathbf{\Omega}$. Then
    \begin{align}
        dist(\mathbf{x}^{*}, \mathbf{S}) \leq \frac{L}{\alpha} + \sqrt{\frac{\delta}{\alpha}}.
    \end{align}
\end{theorem}

\subsubsection{Proofs}
We provide proofs of theorems presented in the main paper and restated here for completeness.

\newtheorem*{theorem5-2}{\textbf{Theorem 5.2}}
\begin{theorem5-2}
    Suppose for any setting of $\mathbf{w}_{nn} \in \mathbb{R}^{n_g}$ there is a feasible solution to NeuPSL inference \eqref{eq:regularized_lcqp_primal}.
    Further, suppose $\epsilon > 0$, $\mathbf{w}_{sy} \in \mathbb{R}_{+}^{r}$, and $\mathbf{w}_{nn} \in \mathbb{R}^{n_g}$.
    Then:
    \begin{itemize}[leftmargin=*,noitemsep,topsep=0pt]
        \item The minimizer of \eqref{eq:regularized_lcqp_primal}, $\mathbf{y}^*(\mathbf{w}_{sy}, \mathbf{w}_{nn})$, is a $O(1 / \epsilon)$ Lipschitz continuous function of $\mathbf{w}_{sy}$.
        \item $V(\mathbf{w}_{sy}, \mathbf{b}(\mathbf{x}_{sy}, \mathbf{g}(\mathbf{x}_{nn}, \mathbf{w}_{nn})))$, is concave over $\mathbf{w}_{sy}$ and convex over $\mathbf{b}(\mathbf{x}_{sy}, \mathbf{g}(\mathbf{x}_{nn}, \mathbf{w}_{nn}))$.
        \item $V(\mathbf{w}_{sy}, \mathbf{b}(\mathbf{x}_{sy}, \mathbf{g}(\mathbf{x}_{nn}, \mathbf{w}_{nn})))$ is differentiable with respect to $\mathbf{w}_{sy}$. Moreover,
        {\small
        \begin{align*}
            \nabla_{\mathbf{w}_{sy}} V(\mathbf{w}_{sy}, & \mathbf{b}(\mathbf{x}_{sy}, \mathbf{g}(\mathbf{x}_{nn}, \mathbf{w}_{nn}))) = \mathbf{\Phi}(\mathbf{y}^{*}(\mathbf{w}_{sy}, \mathbf{w}_{nn}), \mathbf{x}_{sy}, \mathbf{g}(\mathbf{x}_{nn}, \mathbf{w}_{nn})).
        \end{align*}
        }%
        Furthermore, $\nabla_{\mathbf{w}_{sy}} V(\mathbf{w}_{sy}, \mathbf{b}(\mathbf{x}_{sy}, \mathbf{g}(\mathbf{x}_{nn}, \mathbf{w}_{nn})))$ is Lipschitz continuous over $\mathbf{w}_{sy}$.
        \item If there is a feasible point $\nu$ strictly satisfying the $i'th$ inequality constraint of \eqref{eq:regularized_lcqp_primal}, i.e., $\mathbf{A}[i] \mathbf{\nu} + \mathbf{b}(\mathbf{x}_{sy}, \mathbf{g}(\mathbf{x}_{nn}, \mathbf{w}_{nn}))[i] < 0$, then $V(\mathbf{w}_{sy}, \mathbf{b}(\mathbf{x}_{sy}, \mathbf{g}(\mathbf{x}_{nn}, \mathbf{w}_{nn})))$ is subdifferentiable with respect to the $i'th$ constraint constant $\mathbf{b}(\mathbf{x}_{sy}, \mathbf{g}(\mathbf{x}_{nn}, \mathbf{w}_{nn}))[i]$. Moreover, 
        {\small
        \begin{align*}
            \partial_{\mathbf{b}[i]} V(\mathbf{w}_{sy}, & \mathbf{b}(\mathbf{x}_{sy}, \mathbf{g}(\mathbf{x}_{nn}, \mathbf{w}_{nn}))) 
            % \\ &
            = \{\mathbf{\mu}^{*}[i] \, \vert \, \mathbf{\mu}^{*} \in \argmin_{\mathbf{\mu} \in \mathbb{R}_{\geq 0}^{2 \cdot n_{\mathbf{y}} + m + q}}
            h(\mathbf{\mu}; \mathbf{w}_{sy}, \mathbf{b}(\mathbf{x}_{sy}, \mathbf{g}(\mathbf{x}_{nn}, \mathbf{w}_{nn}))) \}.
        \end{align*}
        }%
        Furthermore, if $\mathbf{g}(\mathbf{x}_{nn}, \mathbf{w}_{nn})$ is a smooth function of $\mathbf{w}_{nn}$, then so is $\mathbf{b}(\mathbf{x}_{sy}, \mathbf{g}(\mathbf{x}_{nn}, \mathbf{w}_{nn}))$, and the set of regular subgradients of $V(\mathbf{w}_{sy}, \mathbf{b}(\mathbf{x}_{sy}, \mathbf{g}(\mathbf{x}_{nn}, \mathbf{w}_{nn})))$ is:
        {\small
        \begin{align}
            \hat{\partial}_{\mathbf{w}_{nn}} & V(\mathbf{w}_{sy}, \mathbf{b}(\mathbf{x}_{sy}, \mathbf{g}(\mathbf{x}_{nn}, \mathbf{w}_{nn}))) 
            \\ & 
            \supset \nabla_{\mathbf{w}_{nn}} \mathbf{b}(\mathbf{x}_{sy}, \mathbf{g}(\mathbf{x}_{nn}, \mathbf{w}_{nn}))^T \partial_{\mathbf{b}} V(\mathbf{w}_{sy}, \mathbf{b}(\mathbf{x}_{sy}, \mathbf{g}(\mathbf{x}_{nn}, \mathbf{w}_{nn}))) 
            \nonumber
            .
        \end{align}
        }%
    \end{itemize}
\end{theorem5-2}

\begin{proof}[\textbf{Proof of \thmref{thm:continuity_properties}}]
    \label{proof:continuity_properties}
    We first show the minimizer of the LCQP formulation of NeuPSL inference, $\mathbf{\nu}^*$, with $\epsilon > 0$, $\mathbf{w}_{sy} \in \mathbb{R}_{+}^{r}$, and $\mathbf{w}_{nn} \in \mathbb{R}^{n_g}$ is a Lipschitz continuous function of $\mathbf{w}_{sy}$.
    Suppose $\epsilon > 0$.
    To show continuity over  $\mathbf{w}_{sy} \in \mathbb{R}^{r}_{+}$, first note the matrix $(\mathbf{D} + \epsilon \mathbf{I})$ is positive definite and the primal inference problem \eqref{eq:dual_lcqp_inference} is an $\epsilon$-strongly convex LCQP with a unique minimizer denoted by $\nu^{*}(\mathbf{w}_{sy}, \mathbf{w}_{nn})$.
    We leverage the Lipschitz stability result for optimal values of constrained problems from \cite{bonnans:book00} and presented here in \thmref{thm:bonnans-shapiro-00}.
    Define the primal objective as an explicit function of the weights:
    
    \begin{align}
         f(\mathbf{\nu}, \mathbf{w}_{sy}, \mathbf{w}_{nn}) := \mathbf{\nu}^T (\mathbf{D}(\mathbf{w}_{sy}) + \epsilon \mathbf{I}) \mathbf{\nu} + \mathbf{c}^T(\mathbf{w}_{sy}) \mathbf{\nu}
    \end{align}
    Note that the solution \(\mathbf{\nu}^* =
    \begin{bmatrix}
        \mathbf{s}_{S}^* \\
        \mathbf{s}_{L}^* \\
        \mathbf{y}^*
    \end{bmatrix}\) will always be bounded, since from \eqref{eq:lcqp.3} in LCQP we always have for all \(j \in I_{S} \cup I_{L}\),
    \begin{align}
        0 \le s_j^* &= \max ( \mathbf{a}_{\phi_k, y}^T \mathbf{y}^* + \mathbf{a}_{\phi_k, \mathbf{x}_{sy}}^T \mathbf{x}_{sy} + \mathbf{a}_{\phi_k, \mathbf{g}}^T \mathbf{g}(\mathbf{x}_{nn}, \mathbf{w}_{nn}) +  b_{\phi_k}, 0) \\
        &\le \|\mathbf{a}_{\phi_k, y}\| + |\mathbf{a}_{\phi_k, \mathbf{x}_{sy}}^T \mathbf{x}_{sy} + \mathbf{a}_{\phi_k, \mathbf{g}}^T \mathbf{g}(\mathbf{x}_{nn}, \mathbf{w}_{nn}) +  b_{\phi_k}|. 
    \end{align}
    Thus, setting these trivial upper bounds for \(s_j\) will not change the solution of the problem. 
    We can henceforth consider the problem in a bounded domain \(\| \mathbf{\nu} \| \le C\) where \(C\) does not depend on \(\mathbf{w}\)'s.
    
    Let $\mathbf{w}_{1, sy}, \mathbf{w}_{2, sy} \in \mathbb{R}_{+}^{r}$ and $\mathbf{w}_{nn} \in \mathcal{W}_{nn}$ be arbitrary.
    As $\epsilon > 0$, $f(\mathbf{\nu}, \mathbf{w}_{1, sy}, \mathbf{w}_{nn})$ is strongly convex in $\mathbf{\nu}$ and it therefore satisfies the second-order growth condition in $\mathbf{\nu}$. 
    Define the difference function:
    \begin{align}
            \Delta_{\mathbf{w}_{sy}}(\mathbf{\nu}) &:= f(\mathbf{\nu}, \mathbf{w}_{2, sy}, \mathbf{w}_{nn}) - f(\mathbf{\nu}, \mathbf{w}_{1, sy}, \mathbf{w}_{nn})\\
            &= \mathbf{\nu}^T (\mathbf{D}(\mathbf{w}_{2, sy}) + \epsilon \mathbf{I}) \mathbf{\nu} + \mathbf{c}^T(\mathbf{w}_{2, sy}) \mathbf{\nu} - \big( \mathbf{\nu}^T (\mathbf{D}(\mathbf{w}_{1, sy}) + \epsilon \mathbf{I}) \mathbf{\nu} + \mathbf{c}^T(\mathbf{w}_{1, sy}) \mathbf{\nu} \big) \\
            &= \mathbf{\nu}^T (\mathbf{D}(\mathbf{w}_{2, sy}) - \mathbf{D}(\mathbf{w}_{1, sy})) \mathbf{\nu} + (\mathbf{c}(\mathbf{w}_{2, sy}) - \mathbf{c}(\mathbf{w}_{1, sy}))^T \mathbf{\nu}. 
    \end{align}
    The difference function $\Delta_{\mathbf{w}_{sy}}(\mathbf{\nu})$ over $\mathcal{N}$ has a finitely bounded gradient:
        \begin{align} 
        \Vert \nabla \Delta_{\mathbf{w}_{sy}}(\mathbf{\nu}) \Vert_{2}
        &= \Big \Vert 
        2 (\mathbf{D}(\mathbf{w}_{2, sy}) - \mathbf{D}(\mathbf{w}_{1, sy})) \mathbf{\nu} + \mathbf{c}(\mathbf{w}_{2, sy}) - \mathbf{c}(\mathbf{w}_{1, sy})
        \Big \Vert_{2} \\
        &\le \Vert \mathbf{c}(\mathbf{w}_{2, sy}) - \mathbf{c}(\mathbf{w}_{1, sy}) \Vert_{2} + 2 \Vert (\mathbf{D}(\mathbf{w}_{2, sy}) - \mathbf{D}(\mathbf{w}_{1, sy})) \mathbf{\nu} \Vert_{2} \\
        &\le \Vert \mathbf{w}_{2, sy} - \mathbf{w}_{1, sy} \Vert_{2} +
        2\Vert \mathbf{w}_{2, sy} - \mathbf{w}_{1, sy} \Vert_{2} \; \| \mathbf{\nu} \|_{2} \\
        &\le \Vert \mathbf{w}_{2, sy} - \mathbf{w}_{1, sy} \Vert_{2} (1 + 2C) =: 
        L_{\mathcal{N}}(\mathbf{w}_{1, sy}, \mathbf{w}_{2, sy})
        .
    \end{align}
    Thus, the distance function, $\Delta_{\mathbf{w}_{sy}}(\mathbf{\nu})$ is \(L_{\mathcal{N}}(\mathbf{w}_{1, sy}, \mathbf{w}_{2, sy})\)-Lipschitz continuous over $\mathcal{N}$.
    Therefore, by \cite{bonnans:book00} (\thmref{thm:bonnans-shapiro-00}), the distance between $\mathbf{\nu}^{*}(\mathbf{w}_{1, sy}, \mathbf{w}_{nn})$ and $\mathbf{\nu}^{*}(\mathbf{w}_{2, sy}, \mathbf{w}_{nn})$ is bounded above:
    \begin{align}
        \Vert \mathbf{\nu}^{*}(\mathbf{w}_{2, sy}, \mathbf{w}_{nn}) - \mathbf{\nu}^{*}(\mathbf{w}_{1, sy}, \mathbf{w}_{nn}) \Vert_{2} \leq \frac{L_{\mathcal{N}}(\mathbf{w}_{1, sy}, \mathbf{w}_{2, sy})}{\epsilon} = \frac{(1 + 2C)}{\epsilon} \Vert \mathbf{w}_{2, sy} - \mathbf{w}_{1, sy} \Vert_{2}.
    \end{align}
    Therefore, the function $\mathbf{\nu}^{*}(\mathbf{w}_{sy}, \mathbf{w}_{nn})$ is 
\(O(1/\epsilon)\)-Lipschitz continuous in $\mathbf{w}_{sy}$ for any $\mathbf{w}_{nn}$.

    Next, we prove curvature properties of the value-function with respect to the weights.
    Observe NeuPSL inference is an infimum over a set of functions that are concave (affine) in $\mathbf{w}_{sy}$.
    Therefore, by \thmref{thm:pointwise_infimum_over_concave}, we have that $V(\mathbf{w}_{sy}, \mathbf{b}(\mathbf{x}_{sy}, \mathbf{g}(\mathbf{x}_{nn}, \mathbf{w}_{nn})))$ is concave in $\mathbf{w}_{sy}$.

    We use a similar argument to show $V(\mathbf{w}_{sy}, \mathbf{b}(\mathbf{x}_{sy}, \mathbf{g}(\mathbf{x}_{nn}, \mathbf{w}_{nn})))$ is convex in the constraint constants, $\mathbf{b}(\mathbf{x}_{sy}, \mathbf{g}(\mathbf{x}_{nn}, \mathbf{w}_{nn}))$.
    Assuming for any setting of the neural weights, $\mathbf{w}_{nn} \in \mathbb{R}^{n_g}$, there is a feasible solution to the NeuPSL inference problem, then \eqref{eq:regularized_lcqp_primal} satisfies the conditions for Slater's constraint qualification.
    Therefore, strong duality holds, i.e., $V(\mathbf{w}_{sy}, \mathbf{b}(\mathbf{x}_{sy}, \mathbf{g}(\mathbf{x}_{nn}, \mathbf{w}_{nn})))$ is equal to the optimal value of the dual inference problem \eqref{eq:dual_lcqp_appendix}.
    Observe that the dual NeuPSL inference problem is a supremum over a set of functions convex (affine) in $\mathbf{b}(\mathbf{x}_{sy}, \mathbf{g}(\mathbf{x}_{nn}, \mathbf{w}_{nn}))$.
    Therefore, by \thmref{thm:pointwise_supremum_over_convex}, we have that $V(\mathbf{w}_{sy}, \mathbf{b}(\mathbf{x}_{sy}, \mathbf{g}(\mathbf{x}_{nn}, \mathbf{w}_{nn})))$ is convex in $\mathbf{b}(\mathbf{x}_{sy}, \mathbf{g}(\mathbf{x}_{nn}, \mathbf{w}_{nn}))$.

    We can additionally prove convexity in \(\mathbf{b}\) from first principles. 
    For simplicity we fix other parameters, and write the objective and the value function as \(Q(\nu)\) and \(V(\mathbf{b})\). 
    Given two values $\mathbf{b}_1$ and $\mathbf{b}_2$, let corresponding optimal values of (33) be $\nu_1$ and $\nu_2$. Take any $\alpha \in [0,1]$, note that when $\mathbf{b} = \alpha \mathbf{b}_1 + (1-\alpha) \mathbf{b}_2$, then $\alpha \nu_1  + (1-\alpha) \nu_2$ is feasible for this $\mathbf{b}$. Because we take the inf over all $\nu$s, the optimal $\nu$ for this $\mathbf{b}$ might be even smaller. Thus, we have  (for convex quadratic objective $Q$) that
    \begin{equation}
    \begin{aligned}
    V(\alpha b_1 + (1-\alpha) b_2) & \le  Q(\alpha \nu_1 + (1-\alpha) \nu_2)  \\
    & \le \alpha Q(\nu_1) + (1-\alpha) Q(\nu_2) \\
    & = \alpha V(b_1) + (1-\alpha) V(b_2),
    \end{aligned}
    \end{equation}
    which shows that \(V\) is convex in \(\mathbf{b}\).

    Next, we prove (sub)differentiability properties of the value-function.
    Suppose $\epsilon > 0$.
    First, we show the optimal value function, $V(\mathbf{w}_{sy}, \mathbf{b}(\mathbf{x}_{sy}, \mathbf{g}(\mathbf{x}_{nn}, \mathbf{w}_{nn})))$, is differentiable with respect to the symbolic weights.
    Then we show subdifferentiability properties of the optimal value function with respect to the constraint constants.
    Finally, we apply the Lipschitz continuity of the minimzer result to show the gradient of the optimal value function is Lipschitz continuous with respect to $\mathbf{w}_{sy}$.

    Starting with differentiability with respect to the symbolic weights, $\mathbf{w}_{sy}$, note, the optimal value function of the regularized LCQP formulation of NeuPSL inference, \eqref{eq:regularized_lcqp_primal}, is equivalently expressed as the following maximization over a continuous function in the primal target variables, $\mathbf{y}$, the slack variables, $\mathbf{s}_{S}$ and $\mathbf{s}_{L}$, and the symbolic weights, $\mathbf{w}_{sy}$:
    \begin{align}
        & V(\mathbf{w}_{sy}, \mathbf{b}(\mathbf{x}_{sy}, \mathbf{g}(\mathbf{x}_{nn}, \mathbf{w}_{nn}))) \\
        & \quad = - \Bigg ( \max_{\mathbf{y}, \mathbf{s_{H}}, \mathbf{s_L}} \,
        - \Big ( \begin{bmatrix}
            \mathbf{s}_{S} \\ \mathbf{s}_{L} \\ \mathbf{y}
        \end{bmatrix}^T
        \begin{bmatrix}
            \mathbf{W}_{S} + \epsilon I & 0 & 0\\
            0 & \epsilon I & 0 \\
            0 & 0 & \epsilon I \\
        \end{bmatrix}
        \begin{bmatrix}
            \mathbf{s}_{S} \\
            \mathbf{s}_{L} \\
            \mathbf{y}  
        \end{bmatrix}
        + 
        \begin{bmatrix}
           0 \\ \mathbf{w}_{L} \\ 0
        \end{bmatrix}^T
        \begin{bmatrix}
            \mathbf{s}_{S} \\
            \mathbf{s}_{L} \\
            \mathbf{y}
        \end{bmatrix}
        \Big ) \Bigg ) \nonumber \\
        & \quad \quad \quad \quad \textrm{s.t.} \quad         
        \mathbf{A}         
        \begin{bmatrix}
            \mathbf{s}_{S} \\
            \mathbf{s}_{L} \\
            \mathbf{y}
        \end{bmatrix} + \mathbf{b}(\mathbf{x}_{sy}, \mathbf{g}(\mathbf{x}_{nn}, \mathbf{w}_{nn}) \leq 0 \nonumber,
    \end{align}
    where the matrix $\mathbf{W}_{s}$ and vector $\mathbf{w}_{L}$ are functions of the symbolic parameters $\mathbf{w}_{sy}$ as defined in \eqref{eq:symbolic_weight_matrix_and_vector}.
    Moreover, the objective above is and convex (affine) in $\mathbf{w}_{sy}$.
    Additionally, note that the decision variables can be constrained to a compact domain without breaking the equivalence of the formulation.
    Specifically, the target variables are constrained to the box $[0, 1]^{\mathbf{n}_{y}}$, while the slack variables are nonnegative and have a trivial upper bound derived from \eqref{eq:lcqp.3}:,
    \begin{align}
        0 \le s_j^* &= \max ( \mathbf{a}_{\phi_k, y}^T \mathbf{y}^* + \mathbf{a}_{\phi_k, \mathbf{x}_{sy}}^T \mathbf{x}_{sy} + \mathbf{a}_{\phi_k, \mathbf{g}}^T \mathbf{g}(\mathbf{x}_{nn}, \mathbf{w}_{nn}) +  b_{\phi_k}, 0) \nonumber \\
        &\le \|\mathbf{a}_{\phi_k, y}\| + |\mathbf{a}_{\phi_k, \mathbf{x}_{sy}}^T \mathbf{x}_{sy} + \mathbf{a}_{\phi_k, \mathbf{g}}^T \mathbf{g}(\mathbf{x}_{nn}, \mathbf{w}_{nn}) +  b_{\phi_k}|, 
    \end{align}
    for all \(j \in I_{S} \cup I_{L}\).
    Therefore, the negative optimal value function satisfies the conditions for Danskin's theorem \cite{danskin:siam66} (stated in \appref{appendix:continuity_of_inference_preliminaries}).
    Moreover, as there is a single unique solution to the inference problem when $\epsilon > 0$, and the quadratic objective in \eqref{eq:regularized_lcqp_primal} is differentiable for all $\mathbf{w}_{sy} \in \mathbb{R}^{r}_{+}$, we can apply \corollaryref{corollary:danskins_unique_solution}.
    The optimal value function is therefore concave and differentiable with respect to the symbolic weights with
    \begin{align}
        \nabla_{\mathbf{w}_{sy}} V(\mathbf{w}_{sy}, \mathbf{b}(\mathbf{x}_{sy}, \mathbf{g}(\mathbf{x}_{nn}, \mathbf{w}_{nn})) = \mathbf{\Phi(\mathbf{y}^{*}, \mathbf{x}_{sy}, \mathbf{g}(\mathbf{x}_{nn}, \mathbf{w}_{nn}))}.
    \end{align}

    Next, we show subdifferentiability of the optimal value-function with respect to the constraint constants, $\mathbf{b}(\mathbf{x}_{sy}, \mathbf{g}(\mathbf{x}_{nn}, \mathbf{w}_{nn}))$.
    Suppose at a setting of the neural weights $\mathbf{w}_{nn} \in \mathbb{R}^{n_g}$ there is a feasible point $\nu$ for the NeuPSL inference problem. 
    Moreover, suppose $\nu$ strictly satisfies the $i'th$ inequality constraint of \eqref{eq:regularized_lcqp_primal}, i.e., $\mathbf{A}[i] \mathbf{\nu} + \mathbf{b}(\mathbf{x}_{sy}, \mathbf{g}(\mathbf{x}_{nn}, \mathbf{w}_{nn}))[i] < 0$.
    Observe that the following strongly convex conic program is equivalent to the LCQP formulation of NeuPSL inference, \eqref{eq:regularized_lcqp_primal}:
    \begin{align}
        \min_{\mathbf{\nu} \in \mathbb{R}^{n_{\mathbf{y}} + m_{S} + m_{L}}} & \, 
            \mathbf{\nu}^T (\mathbf{D}(\mathbf{w}_{sy}) + \epsilon \mathbf{I}) \mathbf{\nu} + \mathbf{c}(\mathbf{w}_{sy})^T \mathbf{\nu} + P_{\Omega \setminus i}(\mathbf{\nu}) \label{eq:regularized_lcqp_cone_primal}\\
            \textrm{s.t.} \quad         
            & \mathbf{A}[i] \mathbf{\nu} + \mathbf{b}(\mathbf{x}_{sy}, \mathbf{g}(\mathbf{x}_{nn}, \mathbf{w}_{nn}))[i] \in \mathbb{R}_{\leq 0} \nonumber,
    \end{align}
    where $P_{\Omega \setminus i}(\mathbf{\nu}): \mathbb{R}^{n_{\mathbf{y}} + m_{S} + m_{L}} \to \{0, \infty\}$ is the indicator function identifying feasibility w.r.t. all the constraints of the LCQP formulation of NeuPSL inference in \eqref{eq:regularized_lcqp_primal} except the $i'th$ constraint: $\mathbf{A}[i] \mathbf{\nu} + \mathbf{b}(\mathbf{x}_{sy}, \mathbf{g}(\mathbf{x}_{nn}, \mathbf{w}_{nn}))[i] \leq 0$.
    In other words, in the conic formulation above only the $i'th$ constraint is explicit.
    Note that $\mathbb{R}_{\leq 0}$ is a closed convex cone in $\mathbb{R}$.
    Moreover, both the objective in the program and the mapping $G(\mathbf{\nu}) := \mathbf{A}[i] \mathbf{\nu} + \mathbf{b}(\mathbf{x}_{sy}, \mathbf{g}(\mathbf{x}_{nn}, \mathbf{w}_{nn}))[i]$ are convex.
    Lastly, note the constraint qualification \eqref{eq:cone_constraint_qualification} is similar to Slater's condition in the case of \eqref{eq:regularized_lcqp_cone_primal} which is satisfied by the supposition there exists a feasible $\nu$ that strictly satisfies the $i'th$ inequality constraint of \eqref{eq:regularized_lcqp_primal}.
    Therefore, \eqref{eq:regularized_lcqp_cone_primal} satisfies the conditions of \thmref{thm:value_function_subdifferentiable_wrt_constraint_constants}.
    Thus, the value function is continuous in the constraint constant $\mathbf{b}(\mathbf{x}_{sy}, \mathbf{g}(\mathbf{x}_{nn}, \mathbf{w}_{nn}))[i]$ at $\mathbf{w}_{nn}$ and 
    \begin{align}
        \partial_{\mathbf{b}[i]} V(\mathbf{w}_{sy}, \mathbf{b}(\mathbf{x}_{sy}, \mathbf{g}(\mathbf{x}_{nn}, \mathbf{w}_{nn}))) &= \{\mathbf{\mu}^{*}[i] \, \vert \, \mathbf{\mu}^{*} \in \argmin_{\mathbf{\mu} \in \mathbb{R}_{\geq 0}^{2 \cdot n_{\mathbf{y}} + m + q}}
        h(\mathbf{\mu}; \mathbf{w}_{sy}, \mathbf{b}(\mathbf{x}_{sy}, \mathbf{g}(\mathbf{x}_{nn}, \mathbf{w}_{nn}))) \}.
    \end{align}
    Moreover, when $\mathbf{b}(\mathbf{x}_{sy}, \mathbf{g}(\mathbf{x}_{nn}, \mathbf{w}_{nn}))$ is a smooth function of the neural weights $\mathbf{w}_{nn}$, then we can apply the chain rule for regular subgradients, \thmref{theorem:subgradient_chain_rule}, to get 
    \begin{align}
        \hat{\partial}_{\mathbf{w}_{nn}} V(\mathbf{w}_{sy}, \mathbf{b}(\mathbf{x}_{sy}, \mathbf{g}(\mathbf{x}_{nn}, \mathbf{w}_{nn})) & \supset \nabla \mathbf{b}(\mathbf{x}_{sy}, \mathbf{g}(\mathbf{x}_{nn}, \mathbf{w}_{nn}) ^T \partial_{\mathbf{b}} V(\mathbf{w}_{sy}, \mathbf{b}(\mathbf{x}_{sy}, \mathbf{g}(\mathbf{x}_{nn}, \mathbf{w}_{nn})).
    \end{align}

    To prove the optimal value function is Lipschitz smooth over $\mathbf{w}_{sy}$, it is equivalent to show it is continuously differentiable and that all gradients have bounded magnitude.
    To show the value function is continuously differentiable, we first apply the result asserting the minimizer is unique and a continuous function of the symbolic parameters $\mathbf{w}_{sy}$.
    Therefore, the optimal value function gradient is a composition of continuous functions, hence continuous in $\mathbf{w}_{sy}$.
    The fact that the value function has a bounded gradient magnitude follows from the fact that the decision variables $\mathbf{y}$ have a compact domain over which the gradient is finite; hence a trivial and finite upper bound exists on the gradient magnitude.
\end{proof}

\section{Extended dual block coordinate descent}
\label{appendix:extended_dual_bcd}

We introduce a novel block coordinate descent (BCD) algorithm for the dual LCQP formulation of NeuPSL inference in \eqref{eq:dual_lcqp_appendix}, a bound-constrained, strongly convex quadratic program.
In this section, we omit the symbolic and neural weights from the function arguments to simplify notation.
We define $U_i$, $i = 1,2,\dotsc, p$ to be a cover of the dual variable components $\{1,2,\dotsc,n_{\mathbf{y}} + m + q\}$.
In practice, blocks are defined as a single dual variable corresponding to a constraint from the feasible set or a deep hinge-loss function, along with the dual variables corresponding to the bounds of the primal variables in the constraint or hinge-loss.

We will deal with a slightly more general objective,
\begin{align}
    h(\mathbf{\mu}) := \frac{1}{2} \mathbf{\mu}^T \mathbf{A} \tilde{\mathbf{D}} \mathbf{A}^T \mathbf{\mu} + \tilde{\mathbf{c}}^T \mathbf{\mu},
\end{align}
from which we can recover \eqref{eq:lcqp_lagrangian_appendix} by replacing \(\tilde{\mathbf{D}} \gets (\mathbf{D} + \epsilon \mathbf{I})^{-1}\) and \(\tilde{\mathbf{c}} \gets \mathbf{A} (\mathbf{D} + \epsilon \mathbf{I})^{-1} \mathbf{c} - 2 \mathbf{b}\).

We will use the superscript \(\cdot^{(l)}\) to denote values in the \(l\)-th iteration and subscript \(\cdot_{[i]}\) for the values corresponding to the block \(U_i\). 
The row submatrix of $\mathbf{A}$ that corresponds to block $i$ is denote by $\mathbf{A}_{[i]}$.

At each iteration $l$, we choose one block $i \in \{1,2,\dotsc,p\}$ at random and compute the subvector of $\nabla h(\mu^{[l]})$ that corresponds to this block,
\begin{align}
    \mathbf{d}^{(l)}_{[i]} := \nabla_{[i]} h(\mathbf{\mu}^{(l)}) = (\mathbf{A} \tilde{\mathbf{D}} \mathbf{A}^T \mathbf{\mu}^{(l)} + \tilde{\mathbf{c}})_{[i]}.
\end{align}
Defining $\mathbf{d}^{(l)}$ to be the vector in $\mathbb{R}^N$ whose $i$th block is $\mathbf{d}^{(l)}_{[i]}$ with zeros elsewhere, we  perform a line search along the negative of this direction. 
Note that 
\begin{align}
        h(\mathbf{\mu}^{(l)} - \alpha \mathbf{d}^{(l)}) &= \frac{1}{2} \alpha^2 \mathbf{d}^{(l)T} \mathbf{A} \tilde{\mathbf{D}} \mathbf{A}^T \mathbf{d}^{(l)} - \alpha \mathbf{d}^{(l)T}(\mathbf{A} \tilde{\mathbf{D}} \mathbf{A}^T \mathbf{\mu}^{(l)} + \tilde{\mathbf{c}}) + \textbf{constant} \\
        % &= \frac{1}{2} \alpha^2 \mathbf{d}^{(l)T} \mathbf{A} \mathbf{D} \mathbf{A}^T \mathbf{d}^{(l)} - \alpha \mathbf{d}^{(l)T} \mathbf{d}^{(l)} + \textbf{constant} \\
        &= \frac{1}{2} \alpha^2 \mathbf{d}_{[i]}^{(l)T} \mathbf{A}_{[i]} \tilde{\mathbf{D}} \mathbf{A}_{[i]}^T \mathbf{d}_{[i]} ^{(l)} - \alpha \mathbf{d}_{[i]}^{(l)T} \mathbf{d}_{[i]}^{(l)} + \textbf{constant}.
\end{align}
The unconstrained minimizer of this expression is
\begin{equation}
    \alpha_l^* = \frac{\mathbf{d}_{[i]}^{(l)T}\mathbf{d}_{[i]}^{(l)}}{\mathbf{d}_{[i]}^{(l)T} \mathbf{A}_{[i]} \tilde{\mathbf{D}} \mathbf{A}_{[i]}^T \mathbf{d}_{[i]}^{(l)}}.
\end{equation}
% Since we are doing block updates, we can replace every occurrence of \(\mathbf{d}^{(l)}\) by \(\mathbf{d}^{(l)}_{[i]}\) and the block \(U_i\) part of the \(ADA^T\).
Given the nonnegativity  constraints, we also need to ensure that \(\mathbf{\mu}^{(l)}_{[i]} - \alpha \mathbf{d}^{(l)}_{[i]} \ge 0\). 
Therefore, our choice of steplength is 
\begin{equation}
    \label{eq:step_length}
    \alpha_l = \min \left \{ \alpha_l^*, \min_{j \in U_i \, : \, \mathbf{d}^{(l)}_j>0} \frac{\mathbf{\mu}^{(l)}_j}{\mathbf{d}^{(l)}_j} \right \}.
\end{equation}
To save some computation, we introduce intermediate variables  \(\mathbf{f}^{(l)} := \mathbf{A}^T \mathbf{d}^{(l)} = \mathbf{A}^T_{[i]} \mathbf{d}^{(l)}_{[i]}\), and \(\mathbf{m}^{(l)} := \mathbf{A}^T \mu^{(l)}\). 
With the intermediate variables, the updates of the BCD algorithm are:
\begin{align}
    \mathbf{d}_{[i]}^{(l)} &\gets \mathbf{A}_{[i]} \tilde{\mathbf{D}} \mathbf{m}^{(l)} + \tilde{\mathbf{c}}_{[i]}, \, \mathbf{f}^{(l)} \gets \mathbf{A}^T_{[i]} \mathbf{d}^{(l)}_{[i]} \\
    \mathbf{m}^{(l+1)} &\gets \mathbf{A}^T (\mu^{(l)} - \alpha_l \mathbf{d}^{(l)}) = \mathbf{m}^{(l)} - \alpha_l \mathbf{f}^{(l)}.
\end{align}
With the steplength suggested by \eqref{eq:step_length}, descent is guaranteed at each iteration.
This property is partially why the dual BCD algorithm is effective at leveraging warmstarts which is valuable for improving the runtime of learning algorithms, as is demonstrated in \secref{sec:learning_runtime}.

\begin{algorithm}[ht]
\caption{Dual LCQP Block Coordinate Descent}
\label{algo:dual_lcqp_bcd}
\begin{algorithmic}[1]
    \STATE Set $l = 0$ and compute an initial feasible point $\mathbf{\mu}^{(0)}$;
    \STATE Compute \(\mathbf{m}^{(0)} = \mathbf{A}^T \mu^{(0)}\);
    \WHILE{Stopping Criterion Not Satisfied}
    \STATE \(S_k \gets\) \textrm{Permutation}(\([1, 2, \dotsc,p]\));
        \FORALL{$i \in S_k$ (in order)}
            \STATE Compute \(\mathbf{d}^{(l)}_{[i]} \gets \mathbf{A}_{[i]} \tilde{\mathbf{D}} \mathbf{m}^{(l)} + \tilde{\mathbf{c}}_{[i]}\); \quad \(\mathbf{f}^{(l)} \gets \mathbf{A}^T_{[i]} \mathbf{d}^{(l)}_{[i]}\);
            \STATE Compute \(\alpha_l \gets \min \left \{\frac{\mathbf{d}^{(l)T}_{[i]} \mathbf{d}^{(l)}_{[i]}}{\mathbf{f}^{(l)T} \tilde{\mathbf{D}} \mathbf{f}^{(l)}},
            \min_{j \in U_i :\,  \mathbf{d}^{(l)}_j>0} \frac{\mathbf{\mu}^{(l)}_j}{\mathbf{d}^{(l)}_j} \right \}\) 
            \STATE \(
                \mathbf{\mu}^{(l + 1)}_{[i]} \gets \mathbf{\mu}^{(l)}_{[i]} - \alpha_l\mathbf{d}^{(l)}_{[i]}\); \quad $\mu^{(l+1)}_{[j]} \gets \mu^{(l)}_{[j]}$ for all $j \ne i$;
            \STATE \(\mathbf{m}^{(l+1)} \gets \mathbf{m}^{(l)} - \alpha_l \mathbf{f}^{(l)}\);
            \STATE{$l \leftarrow l + 1$};
        \ENDFOR
    \STATE \(k \gets k + 1\);
    \ENDWHILE
    \end{algorithmic}
\end{algorithm}

As strong duality holds for the LCQP formulation of deep HL-MRF inference, stopping when the primal-dual gap is below a given threshold $\delta>0$, is a principled stopping criterion.
Formally, at any iteration  \algoref{algo:dual_lcqp_bcd} applied to \eqref{eq:dual_lcqp_appendix}, we recover an estimate of the primal variable $\mathbf{v}$ from \eqref{eq:dual_primal_translation_appendix} and terminate when the gap between the primal and the dual objective falls below $\delta$.
The stopping criterion is checked after every permutation block has been completely iterated over. 

\paragraph{Connected Component Parallel D-BCD}
Oftentimes, the NeuPSL dual inference objective is additively separable over partitions of the variables.
In this case, the dual BCD algorithm is parallelizable over the partitions.
We propose identifying the separable components via the primal objective and constraints.
More formally, prior to the primal problem instantiation, a disjoint-set data structure~\citep{cormen:book09} is initialized such that every primal variable belongs to a single unique disjoint set. 
Then, during instantiaion, the disjoint-set data structure is maintained to preserve the property that two primal variables exist in the same set if and only if they occur together with a non-zero coefficient in a constraint or a potential.
This is achieved by merging the sets of variables in every generated constraint or potential.
This process is made extremely efficient with a path compression strategy implemented to optimize finding set representatives.
This parallelization strategy is empirically studied in \secref{sec:empirical_evaluation} where we refer to it as CC D-BCD.

\paragraph{Lock Free Parallel D-BCD}
In general, there may only be a few connected components in the factor graph of the inference problem.
In this case, D-BCD cannot fully leverage computational resources using the CC D-BCD parallelization strategy.
One solution to overcome this issue and preserve the guaranteed descent property is to lock access and updates to dual variables.
In other words, processes checkout locks on the dual variables to access and update its value and corresponding statistics.
Unfortunately, in practice there is too much overlap in the blocks for this form of synchronization to see runtime improvements.
For this reason, we additionally propose a method that sacrifices the theoretical guaranteed descent property of the dual BCD algorithm for significant runtime improvements.
Our approach is inspired by lock free parallelization strategies in optimization literature \citep{bertsekas:book95, recht:neurips11, liu:jmlr15}.
Specifically, rather than having processes checkout locks on dual variables for the entire iteration, we only assume dual and intermediate variable updates are atomic.
This assumption ensures the dual variables and intermediate variables are synchronized across processes.
However, the steplength subproblem solution and the gradient may be incorrect.
Despite this, in \secref{sec:inference_runtime} we show this distributed variant of the dual BCD algorithm consistently finds a solution satisfying the stopping criterion and realizes significant runtime improvements over the CC D-BCD algorithm in some datasets.

\section{Extended Empirical Evaluation}
\label{appendix:extended_evaluation}

In this section, we provide additional details on the datasets and NeuPSL models used in experiments, hardware used to run experiments, an additional evaluation on the effect of the LCQP regularization on the prediction performance of NeuPSL, more inference runtime experiments, and the hyperparameter details for all the experiments in the main paper.

\subsection{Datasets and NeuPSL Models}
\label{appendix:datasets}

In this section, we provide additional information on all five evaluation datasets and corresponding NeuPSL models.

\subsubsection{4Forums and CreateDebate}
Stance-4Forums and Stance-CreateDebate are two datasets containing dialogues from online debate websites: \url{4forums.com} and \url{createdebate.com}, respectively.
In this paper, we study stance classification, i.e., the task of identifying the stance of a speaker in a debate as being for or against.

The $5$ data splits and the NeuPSL model we evaluate in this paper originated from \citenoun{sridhar:acl15}. 
The data and NeuPSL models are available at: \url{https://github.com/linqs/psl-examples/tree/main/stance-4forums} and \url{https://github.com/linqs/psl-examples/tree/main/stance-createdebate}.

\subsubsection{Epinions} 
Epinions is a trust network with $2,000$ individuals connected by $8,675$ directed edges representing whether they know each other and whether they trust each other \cite{richardson:iswc03}. 
We study link prediction, i.e., we predict if two individuals trust each other.

In each of the $5$ data splits, the entire network is available, and the prediction performance is measured on $\frac{1}{8}$ of the trust labels. 
The remaining set of labels are available for training.
We use The NeuPSL model from \citenoun{bach:jmlr17}.
The data and NeuPSL model are available at \url{https://github.com/linqs/psl-examples/tree/main/epinions}.

\subsubsection{Citeseer and Cora} 
Citeseer and Cora are citation networks introduced by \citenoun{sen:aim08}.
For Citeseer, $3,312$ documents are connected by $4,732$ edges representing citation links. 
For Cora, $2,708$ documents are connected by $5,429$ edges representing citation links.
We study node classification, i.e., we classify the documents into one of $6$ topics for Citeseer and $7$ topics for Cora.

We study two different data settings for evaluations.
For the inference and learning runtime experiments and the HL-MRF learning prediction performance experiments, \secref{sec:inference_runtime}, \secref{sec:learning_runtime}, and \secref{sec:learning_performance}, respectively, the data is split following \citenoun{bach:jmlr17}.
Specifically, for each of the $5$ folds, $1/2$ of the nodes are sampled and specify a graph for training, and the remaining $1/2$ of the nodes define the graph for testing.
$1/2$ of the node labels are observed for both the training and test graphs.
For the deep HL-MRF learning prediction performance setting, \secref{sec:learning_performance}, for each of the $10$ folds, we randomly sample $5\%$ of the node labels for training $5\%$ of the node labels for validation and $1,000$ for testing.

Moreover, we use three different NeuPSL models for this dataset.
The inference and learning runtime experiment models are from \citenoun{bach:jmlr17} \cite{bach:jmlr17}.
The data and NeuPSL models for these experiments are available at: \url{https://github.com/linqs/psl-examples/tree/main/citeseer} and \url{https://github.com/linqs/psl-examples/tree/main/cora} for Citeseer and Cora, respectively.
The models for HL-MRF learning prediction performance experiments are extended versions of those in the inference and learning runtime experiments.
Specifically, a copy of each rule is made that is specialized for the topic.
Moreover, topic propagation across citation links is considered for papers with differing topics. 
For instance, the possibility of a citation from a paper with topic $'\pslarg{A}'$ could imply a paper is more or less likely to be topic $'\pslarg{B}'$.
The extended models are available at \url{https://github.com/convexbilevelnesylearning/experimentscripts/hlmrf_learning/psl-extended-examples}.
The models for deep HL-MRF learning prediction performance experiments are from \citenoun{pryor:ijcai23}.
The data and models are available at: \url{https://github.com/linqs/neupsl-ijcai23}.

\subsubsection{DDI}
Drug-drug interaction (DDI) is a network of $315$ drugs and $4,293$ interactions derived from the DrugBank database \citep{wishart:nar06}.
The edges in the drug network represent interactions and seven different similarity metrics.
In this paper, we perform link prediction, i.e., we infer unknown drug-drug interactions.

The $5$ data splits and the NeuPSL model we evaluate in this paper originated from \citenoun{sridhar:bio16}. 
The data and NeuPSL models are available at: \url{https://github.com/linqs/psl-examples/tree/main/drug-drug-interaction}.

\subsubsection{Yelp}
Yelp is a network of $34,454$ users and $3,605$ items connected by $99,049$ edges representing ratings.
The task is to predict missing ratings, i.e., regression, which could be used in a recommendation system.

In each of the $5$ folds, $80\%$ of the ratings are randomly sampled and available for training, and the remaining $20\%$ is held out for testing.
We use The NeuPSL model from \citenoun{kouki:recsys15}.
The data and NeuPSL model are available at: \url{https://github.com/linqs/psl-examples/tree/main/yelp}.

\subsubsection{MNIST-Addition}
% \begin{figure*}[t]
%     \centering
%     \begin{subfigure}[t]{0.4\linewidth}
%         \centering
%         \includegraphics[width=\linewidth]{images/mnist-addition-1.png}
%         \caption{MNIST-Add1}
%     \end{subfigure}
%     % \hfill
%     \begin{subfigure}[t]{0.4\linewidth}
%         \centering
%         \includegraphics[width=\linewidth]{images/mnist-addition-2.png}
%         \caption{MNIST-Add2}
%     \end{subfigure}
%     \caption{
%         Example of MNIST-Add1 and MNIST-Add2.
%     }
%     \label{fig:mnist-addition}
% \end{figure*}

\begin{figure*}[htp]
  \centering
  \subfigure[Example of MNIST-Add1.]{\includegraphics[scale=0.75]{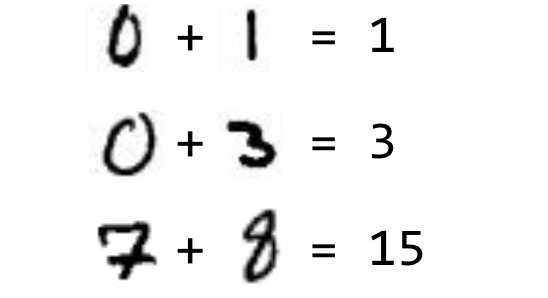}}\quad
  \subfigure[Example of MNIST-Add2.]{\includegraphics[scale=0.75]{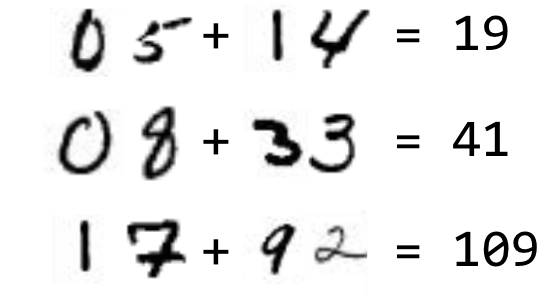}}
  \caption{
    Example of MNIST-Add1 and MNIST-Add2.
  }
  \label{fig:mnist-addition}
\end{figure*}

MNIST Addition is a canonical NeSy image classification dataset first introduced by \cite{manhaeve:neurips18}.
In MNIST-Addition, models must determine the sum of two lists of MNIST images, for example, $\big[\inlinegraphics{images/MNIST-3.png}\big] + \big[\inlinegraphics{images/MNIST-5.png}\big] = \mathbf{8}$.
The challenge stems from the lack of labels for the MNIST images; only the final sum of the equation is provided during training, $8$ in this example.
$5$ MNIST-Addition train splits are generated by randomly sampling, without replacement, $n\in\{600, 6,000, 50,000\}$ unique MNIST images from the original MNIST dataset and converted to MNIST additions.
Specifically, additions are created by creating $n / 2$ non-overlapping pairs of digits from the sample for MNIST-Add1 and $n / 4$ non-overlapping sets of digits from the sample for MNIST-Add2.
Then the MNIST image labels are then added together, as shown in \figref{fig:mnist-addition}, to define the addition label used in the task.
This process is repeated to create five corresponding validation and test splits, with $1,0000$ MNIST examples being sampled per test split from the original MNIST dataset.

\begin{table}[t]
    \centering
    \scriptsize
    \begin{tabular}{cccc}
        \toprule
            Order & Layer       & Parameter       & Value   \\
        \toprule
            1 & ResNet18~\citep{he:cvpr15}\\
        \cmidrule{2-4}
            \multirow{3}{*}{2} & \multirow{3}{*}{Fully Connected} & Input Shape     & 128      \\
              &                 & Output Shape    & 64      \\
              &                 & Activation      & ReLU \\
        \cmidrule{2-4}
            \multirow{3}{*}{3} & \multirow{3}{*}{Fully Connected} & Input Shape     & 64      \\
              &                 & Output Shape    & 10      \\
              &                 & Activation      & Gumbel Softmax~\citep{jang:iclr17} \\
        \bottomrule
    \end{tabular}

    \caption{
        Neural architecture used in NeuPSL MNIST-Add models.
    }

    \label{tab:mnist-nn-arch}
\end{table}

\textbf{MNIST-Add1} The MNIST-Add1 NeuPSL model integrates the neural component summarized in \tabref{tab:mnist-nn-arch} with the symbolic model summarized in \figref{fig:mnist-addition1-symbolic-model}.
The symbolic model contains the following predicates:

\begin{itemize}
    \item \textbf{$\pslpred{Neural}(\pslarg{Img}, \pslarg{X})$}
    The $ \pslpred{Neural} $ predicate is the class probability for each image as inferred by the neural network.
    $ \pslarg{Img} $ is MNIST image identifier and $\pslarg{X}$ is a digit class that the image may represent.

    \item \textbf{$ \pslpred{DigitSumOnesPlace}(\pslarg{W}, \pslarg{X}, \pslarg{Y}, \pslarg{Z})$}
    The $\pslpred{DigitSumOnesPlace}$ predicate represents whether the ones place of the sum of the digits $W$, $X$, and $Y$ is $Z$. 
    For example, substituting $0$, $1$, $2$, and $3$ for $W$, $X$, $Y$ and $Z$, the predicate value would be $1$, but  substituting $1$, $1$, $2$, and $3$ for $W$, $X$, $Y$ and $Z$ would be $0$ since $1 + 1 + 2 + 3 \neq 3$.

    \item \textbf{$ \pslpred{DigitSumTensPlace}(\pslarg{W}, \pslarg{X}, \pslarg{Y}, \pslarg{Z})$}
    The $\pslpred{DigitSumTensPlace}$ predicate represents whether the tens place of the sum of the digits $W$, $X$, and $Y$ is $Z$. 
    For example, substituting $0$, $1$, $2$, and $0$ for $W$, $X$, $Y$ and $Z$, the predicate value would be $1$, but  substituting $0$, $1$, $9$, and $0$ for $W$, $X$, $Y$ and $Z$ would be $0$ since $1 + 1 + 2 + 3 = 10$, i.e., the tens place digit of the sum is $1$ not $0$.

    \item \textbf{$\pslpred{SumPlace}(\pslarg{Img1}, \pslarg{Img2}, \pslarg{place}, \pslarg{Z})$}
    The $ \pslpred{SumPlace} $ predicate is the probability that the digits represented in the images identified by arguments $\pslarg{Img1}$ and $\pslarg{Img2}$ add up to a number with a $\pslarg{place}$'s place of $\pslarg{Z}$.

    \item \textbf{$\pslpred{Carry}(\pslarg{Img1}, \pslarg{Img2}, \pslarg{W})$}
    The $ \pslpred{Carry} $ predicate represents is the probability that the digits represented in the images identified by arguments $\pslarg{Img1}$ and $\pslarg{Img2}$ add up to a number with a carry value of $\pslarg{W}$.
    For example, images representing digits $1$ and $2$ do not have a carry.
    These variables are considered latent in the NeuPSL model as there are no truth labels for carries.

    \item \textbf{$\pslpred{PossibleDigit}(\pslarg{X}, \pslarg{Z})$}
    The $ \pslpred{PossibleDigits} $ predicate represents whether a digit ($\pslarg{X}$) can be included in a sum that equals a number ($\pslarg{Z}$).
    For example, $\pslpred{PossibleDigits}(9, 0)$ would return $0$ as no positive digit when added to $9$ will equal $0$. Conversely, $\pslpred{PossibleDigits}(9, 17)$ would return $1$ as $8$ added to $9$ equals $17$.

    \item \textbf{$\pslpred{ImageSum}(\pslarg{Img1}, \pslarg{Img2}, \pslarg{Z})$}
    The $ \pslpred{ImageSum} $ predicate is the probability that the digits represented by the images specified by $\pslarg{Img1}$ and $\pslarg{Img2}$ sum up to the number indicated by the argument $\pslarg{Z}$. 
    This predicate instantiates decision variables, i.e., variables from this predicate are not fixed during inference and learning as described in the NeSy EBM, NeuPSL, and Inference and Learning sections.

    \item \textbf{$\pslpred{PlacedRepresentation}(\pslarg{Z}_{10}, \pslarg{Z}_1, \pslarg{Z})$}
    The $\pslpred{PlacedRepresentation}$ predicate represents whether the number $Z$ has tens place digit $\pslarg{Z}_{10}$ and ones place digit $\pslarg{Z}_1$. 
\end{itemize}

\begin{figure*}[t] 
    \centering
    \noindent\fbox{%
        \begin{minipage}{0.99\hsize}
            \begin{scriptsize}
            \begin{flalign*}
                & w_{1}: \pslpred{DigitSumOnesPlace}('0', \pslarg{X}, \pslarg{Y}, \pslarg{Z}) \psland \pslpred{Neural}(\pslarg{Img1}, \pslarg{X}) \psland \pslpred{Neural}(\pslarg{Img2}, \pslarg{Y}) \pslthen \pslpred{SumPlace}(\pslarg{Img1}, \pslarg{Img2}, '1', \pslarg{Z}) \\
                & w_{2}: \pslpred{DigitSumTensPlace}('0', \pslarg{X}, \pslarg{Y}, \pslarg{Z}) \psland \pslpred{Neural}(\pslarg{Img1}, \pslarg{X}) \psland \pslpred{Neural}(\pslarg{Img2}, \pslarg{Y}) \pslthen \pslpred{SumPlace}(\pslarg{Img1}, \pslarg{Img2}, '10', \pslarg{Z}) \\[0.3cm]
                & w_{3}: \pslpred{DigitSumOnesPlace}(\pslarg{W}, '0', '0', \pslarg{Z}) \psland \pslpred{Carry}(\pslarg{Img1}, \pslarg{Img2}, \pslarg{W}) \pslthen \pslpred{SumPlace}(\pslarg{Img1}, \pslarg{Img2}, '10', \pslarg{Z}) \\[0.3cm]
                & w_{4}: \pslpred{SumPlace}(\pslarg{ImageId1}, \pslarg{ImageId2}, '1', \pslarg{Z}_{1}) \psland \pslpred{SumPlace}(\pslarg{ImageId1}, \pslarg{ImageId2}, '10', \pslarg{Z}_{10})\\
                & \quad \quad \quad \psland \pslpred{PlacedRepresentation}(\pslarg{Z}_{10}, \pslarg{Z}_{1}, \pslarg{Z}) \pslthen \pslpred{ImageSum}(\pslarg{ImageId1}, \pslarg{ImageId2}, \pslarg{Z})
            \end{flalign*}
            \end{scriptsize}
        \end{minipage}
    }
    \caption{Summarized NeuPSL MNIST-Add1 Symbolic Model. The full model is available at: \url{https://github.com/convexbilevelnesylearning/experimentscripts/mnist_addition/neupsl_models}.}
    \label{fig:mnist-addition1-symbolic-model}
\end{figure*}

\textbf{MNIST-Add2} The MNIST-Add2 NeuPSL model integrates the neural component summarized in \tabref{tab:mnist-nn-arch} with the symbolic model summarized in \figref{fig:mnist-addition2-symbolic-model}.
The symbolic model contains the following predicates:

\begin{itemize}
    \item \textbf{$\pslpred{SumPlace}(\pslarg{Img1}, \pslarg{Img2}, \pslarg{Img3}, \pslarg{Img4}, \pslarg{place}, \pslarg{Z})$}
    The $ \pslpred{SumPlace} $ predicate is the probability that the digits represented in the images identified by arguments $\pslarg{Img1}$, $\pslarg{Img2}$, $\pslarg{Img3}$, and $\pslarg{Img4}$ add up to a number with a $\pslarg{place}$'s place of $\pslarg{Z}$.

    \item \textbf{$\pslpred{PossibleOnesDigit}(\pslarg{X}, \pslarg{Z})$}
    The $ \pslpred{PossibleOnesDigit} $ predicate represents whether a digit ($\pslarg{X}$) can be included in a sum as a ones place digit that equals a number ($\pslarg{Z}$).
    For example, $\pslpred{PossibleDigits}(9, 0)$ would return $0$ as no positive digit when added to $9$ will equal $0$. Conversely, $\pslpred{PossibleDigits}(9, 17)$ would return $1$ as $8$ added to $9$ equals $17$.

    \item \textbf{$\pslpred{PossibleTensDigit}(\pslarg{X}, \pslarg{Z})$}
    The $ \pslpred{PossibleTensDigits} $ predicate represents whether a digit ($\pslarg{X}$) can be included in a sum as a tens place digit that equals a number ($\pslarg{Z}$).
    For example, $\pslpred{PossibleDigits}(9, 0)$ would return $0$ as no positive digit when added to $90$, $91$, $\cdots$, $99$ will equal $0$. Conversely, $\pslpred{PossibleDigits}(9, 97)$ would return $1$ as $7$ added to $90$ equals $97$, for instance.

    \item \textbf{$\pslpred{ImageSum}(\pslarg{Img1}, \pslarg{Img2}, \pslarg{Img3}, \pslarg{Img4}, \pslarg{Z})$}
    The $ \pslpred{ImageSum} $ predicate is the probability that the digits represented by the images specified by $\pslarg{Img1}$, $\pslarg{Img2}$, $\pslarg{Img3}$, and $\pslarg{Img4}$ sum up to the number indicated by the argument $\pslarg{Z}$. 
    This predicate instantiates decision variables, i.e., variables from this predicate are not fixed during inference and learning as described in the NeSy EBM, NeuPSL, and Inference and Learning sections.

    \item \textbf{$\pslpred{PlacedRepresentation}(\pslarg{Z}_{100}, \pslarg{Z}_{10}, \pslarg{Z}_{1}, \pslarg{Z})$}
    The $\pslpred{PlacedRepresentation}$ predicate represents whether the number $Z$ has hundereds place digit $\pslarg{Z}_{100}$ , tens place digit $\pslarg{Z}_{10}$ and ones place digit $\pslarg{Z}_1$. 
\end{itemize}

\begin{figure*}[t] 
    \centering
    \noindent\fbox{%
        \begin{minipage}{0.99\hsize}
            \begin{scriptsize}
            \begin{flalign*}
                & w_{1}: \pslpred{DigitSumOnesPlace}('0', \pslarg{X}, \pslarg{Y}, \pslarg{Z}) \psland \pslpred{Neural}(\pslarg{Img2}, \pslarg{X}) \psland \pslpred{Neural}(\pslarg{Img4}, \pslarg{Y}) \pslthen \pslpred{SumPlace}(\pslarg{Img1}, \pslarg{Img2}, \pslarg{Img3}, \pslarg{Img4} '1', \pslarg{Z}) \\
                & w_{2}: \pslpred{DigitSumTensPlace}('0', \pslarg{X}, \pslarg{Y}, \pslarg{Z}) \psland \pslpred{Neural}(\pslarg{Img2}, \pslarg{X}) \psland \pslpred{Neural}(\pslarg{Img4}, \pslarg{Y}) \pslthen \pslpred{Carry}(\pslarg{Img2}, \pslarg{Img4}, \pslarg{Z}) \\[0.3cm]
                & w_{3}: \pslpred{DigitSumOnesPlace}(\pslarg{W}, \pslarg{X}, \pslarg{Y}, \pslarg{Z}) \psland \pslpred{Neural}(\pslarg{Img1}, \pslarg{X}) \psland \pslpred{Neural}(\pslarg{Img3}, \pslarg{Y}) \psland \pslpred{Carry}(\pslarg{Img2}, \pslarg{Img4}, \pslarg{W}) \\
                & \quad \quad \quad \pslthen \pslpred{SumPlace}(\pslarg{Img1}, \pslarg{Img2}, \pslarg{Img3}, \pslarg{Img4} '10', \pslarg{Z}) \\
                & w_{4}: \pslpred{DigitSumTensPlace}(\pslarg{W}, \pslarg{X}, \pslarg{Y}, \pslarg{Z}) \psland \pslpred{Neural}(\pslarg{Img1}, \pslarg{X}) \psland \pslpred{Neural}(\pslarg{Img3}, \pslarg{Y}) \psland \pslpred{Carry}(\pslarg{Img2}, \pslarg{Img4}, \pslarg{W}) \\
                & \quad \quad \quad \pslthen \pslpred{SumPlace}(\pslarg{Img1}, \pslarg{Img2}, \pslarg{Img3}, \pslarg{Img4} '100', \pslarg{Z})\\
                & w_{5}: \pslpred{DigitSumOnesPlace}(\pslarg{W}, '0', '0', \pslarg{Z}) \psland \pslpred{Carry}(\pslarg{Img1}, \pslarg{Img3}, \pslarg{W}) \pslthen \pslpred{SumPlace}(\pslarg{Img1}, \pslarg{Img2}, \pslarg{Img3}, \pslarg{Img4} '100', \pslarg{Z}) \\[0.3cm]
                & w_{6}: \pslpred{SumPlace}(\pslarg{ImageId1}, \pslarg{ImageId2}, \pslarg{ImageId3}, \pslarg{ImageId4}, '1', \pslarg{Z}_{1}) \\
                & \quad \quad \quad \psland \pslpred{SumPlace}(\pslarg{ImageId1}, \pslarg{ImageId2}, \pslarg{ImageId3}, \pslarg{ImageId4}, '10', \pslarg{Z}_{10}) \\
                & \quad \quad \quad \psland \pslpred{SumPlace}(\pslarg{ImageId1}, \pslarg{ImageId2}, \pslarg{ImageId3}, \pslarg{ImageId4}, '100', \pslarg{Z}_{100}) \psland \pslpred{PlacedRepresentation}(\pslarg{Z}_{100}, \pslarg{Z}_{10}, \pslarg{Z}_{1}, \pslarg{Z})\\
                & \quad \quad \quad \pslthen \pslpred{ImageSum}(\pslarg{ImageId1}, \pslarg{ImageId2}, \pslarg{ImageId3}, \pslarg{ImageId4}, \pslarg{Z})
            \end{flalign*}
            \end{scriptsize}
        \end{minipage}
    }
    \caption{Summarized NeuPSL MNIST-Add2 Symbolic Model. The full model is available at: \url{https://github.com/convexbilevelnesylearning/experimentscripts/mnist_addition/neupsl_models}.}
    \label{fig:mnist-addition2-symbolic-model}
\end{figure*}

\subsection{Hardware}
\label{appendix:hardware}

All timing experiments were performed on an Ubuntu 22.04.1 Linux machine with Intel Xeon Processor E5-2630 v4 at 3.10GHz and 128 GB of RAM.

\subsection{Dual BCD and Regularization}
\label{appendix:empirical_lcqp_regularization}

The regularization parameter added to the LCQP formulation of NeuPSL inference in \eqref{eq:regularized_lcqp_primal} ensures strong convexity of the optimal value of the energy function.
However, adding regularization makes the new formulation an approximation.
In this section, the runtime and prediction performance of the D-BCD inference algorithm is evaluated at varying levels of regularization to understand its effect on NeuPSL inference.
The regularization parameter varies in the range $\epsilon \in \{100, 10, 1, 0.1, 0.01\}$.
The D-BCD algorithm is stopped when the primal-dual gap drops below $\delta = 0.1$
Inference time is provided in seconds, and the performance metric is consistent with \tabref{tab:datasets}.
Results are provided in \tabref{tab:dual-bcd-regularization}.

\begin{table}[ht]
    \centering
    \caption{D-BCD Inference time in seconds and prediction performance with varying values for the LCQP regularization parameter $\epsilon$.}
    \label{tab:dual-bcd-regularization}
    \scalebox{0.7}{
    \begin{tabular}{cc||c|c}
    \toprule 
    \textbf{Dataset} & $\epsilon$ & \textbf{Time (sec)} & \textbf{Perf.} \\
    \midrule
    \midrule
    \multirow{5}{*}{\textbf{CreateDebate (AUROC)}} 
    & $100$ & $0.02 \pm 0.01$ & $ 64.77 \pm 10.61 $ \\
    & $10$ & $0.02 \pm 0.01$ & $ 64.83 \pm 10.53 $ \\
    & $1$ & $0.02 \pm 0.01$ & $ 64.74 \pm 10.67 $ \\
    & $0.1$ & $0.05 \pm 0.02$ & $ 65.39 \pm 9.07 $ \\
    & $0.01$ & $0.42 \pm 0.51$ & $ 66.01 \pm 9.35 $ \\
    \hline    
    \multirow{5}{*}{\textbf{4Forums (AUROC)}} 
    & $100$ & $0.11 \pm 0.02$ & $ 61.31 \pm 6.17 $ \\
    & $10$ & $0.10 \pm 0.03$ & $ 61.26 \pm 6.16 $ \\
    & $1$ & $0.09 \pm 0.01$ & $ 61.12 \pm 6.18 $ \\
    & $0.1$ & $0.43 \pm 0.11$ & $ 62.73 \pm 5.46 $ \\
    & $0.01$ & $7.11 \pm 3.05$ & $ 62.31 \pm 5.47 $ \\
    \hline    
    \multirow{5}{*}{\textbf{Epinions (AUROC)}} 
    & $100$ & $0.33 \pm 0.05$ & $ 72.59 \pm 2.27 $ \\
    & $10$ & $0.28 \pm 0.04$ & $ 72.69 \pm 2.21 $ \\
    & $1$ & $0.33 \pm 0.05$ & $ 74.24 \pm 1.95 $ \\
    & $0.1$ & $1.08 \pm 0.16$ & $ 77.05 \pm 1.06 $ \\
    & $0.01$ & $5.21 \pm 0.37$ & $ 77.45 \pm 0.70 $ \\
    \hline    
    \multirow{5}{*}{\textbf{Citeseer (Accuracy)}} 
    & $100$ & $0.95 \pm 0.14$ & $71.28 \pm 1.31$ \\
    & $10$ & $1.00 \pm 0.12$ & $71.28 \pm 1.30$ \\
    & $1$ & $1.48 \pm 0.29$ & $71.59 \pm 1.01$ \\
    & $0.1$ & $7.01 \pm 1.57$ & $71.75 \pm 1.10$ \\
    & $0.01$ & $62.41 \pm 14.67$ & $71.92 \pm 1.09$ \\
    \hline    
    \multirow{5}{*}{\textbf{Cora (Accuracy)}} 
    & $100$ & $4.53 \pm 2.20$ & $81.31 \pm 1.73$ \\
    & $10$ & $4.56 \pm 2.39$ & $81.57 \pm 1.83$ \\
    & $1$ & $7.36 \pm 4.19$ & $81.48 \pm 1.70$ \\
    & $0.1$ & $42.24 \pm 25.06$ & $81.88 \pm 1.82$ \\
    & $0.01$ & $269.45 \pm 49.50$ & $81.79 \pm 1.72$ \\
    \hline    
    \multirow{5}{*}{\textbf{DDI (AUROC)}} 
    & $100$ & $24.56 \pm 0.25$ & $94.85 \pm 0.00$ \\
    & $10$ & $29.23 \pm 0.59$ & $94.85 \pm 0.00$ \\
    & $1$ & $47.15 \pm 0.95$ & $94.82 \pm 0.00$ \\
    & $0.1$ & $280.62 \pm 5.19$ & $94.80 \pm 0.00$ \\
    & $0.01$ & $266.07 \pm 42.68$ & $94.81 \pm 0.00$ \\
    \hline    
    \multirow{5}{*}{\textbf{Yelp (MAE)}} 
    & $100$ & $105.60 \pm 5.03$ & $ 0.23 \pm 0.01 $ \\
    & $10$ & $3,239 \pm 81$ & $ 0.22 \pm 0.01 $ \\
    & $1$ & $3,227 \pm 54$ & $ 0.19 \pm 0.01 $ \\
    & $0.1$ & $421 \pm 202$ & $ 0.18 \pm 0.00 $ \\
    & $0.01$ & $2,472 \pm 297$ & $ 0.18 \pm 0.00 $ \\
    \bottomrule
    \end{tabular}
    }
\end{table}

\tabref{tab:dual-bcd-regularization} shows there is a consistent correlation between the LCQP regularization parameter and the runtime and performance of inference.
As $\epsilon$ increases, there is a significant decrease in the runtime performance as the D-BCD algorithm can find a solution with a gradient meeting the stopping criterion in fewer iterations.
Notably, for the Citeseer inference problem, the D-BCD algorithm realizes a roughly $45 \times$ speedup. 
On the other hand, while the runtime performance improves with increasing $\epsilon$, the prediction performance can sometimes decay.
There is a tradeoff between runtime and prediction performance when setting the $\epsilon$ regularization parameter.

\subsection{Extended Inference Runtime}
\label{appendix:extended_inference_runtime}

\begin{table}[h!]
    \centering
    \caption{Inference time in seconds for each inference optimization technique.}
    \label{tab:appendix_inference_time}
    \scalebox{0.7}{
    \begin{tabular}{l||c|c|c||c|c}
    \toprule  
    & \textbf{Gurobi} & \textbf{GD} & \textbf{ADMM} & \textbf{CC D-BCD} & \textbf{LF D-BCD} \\
    \midrule
    \midrule
    \textbf{Epinions} & $0.46 \pm 0.01$ & $ 34.63 \pm 0.33 $ & $0.36 \pm 0.041$ & $1.84 \pm 0.4$ & $\mathbf{0.26 \pm 0.04}$ \\
    \textbf{Citeseer} & $0.66 \pm 0.08$ & $ 47.17 \pm 0.61 $ & $0.63 \pm 0.07$ & $1.36 \pm 0.24$ & $\mathbf{0.49 \pm 0.08}$ \\
    \textbf{Cora} & $\mathbf{0.71 \pm 0.08}$ & $ 48.66 \pm 1.24 $ & $\mathbf{0.71 \pm 0.07}$ & $6.46 \pm 3.5$ & $0.79 \pm 0.19$ \\
    \textbf{Yelp} & $7.38 \pm 0.20$ & $6,961 \pm 46$ & $\mathbf{6.37 \pm 1.19}$ & $48.44 \pm 3.82$ & $7.58 \pm 0.48$ \\
    \bottomrule
    \end{tabular}
    }
\end{table}

\begin{table}[ht]
    \centering
    \caption{Hyperparameter ranges and final values for the inference runtime experiments.}
    \label{tab:inference_runtime_hyperparameters}
    \scalebox{0.7}{
    \begin{tabular}{c|c||c|c}
        \toprule
            \textbf{Dataset} & \textbf{Parameter} & \textbf{Range} & \textbf{Final Value} \\
         \midrule
         \midrule
            \multirow{2}{*}{\textbf{CreateDebate}} 
            & \textbf{ADMM Step Length} & $\{10.0, 1.0, 0.1, 0.01\}$ & $1.0$ \\
            & \textbf{LCQP Regularization} & $\{100, 10, 1, 0.1, 0.01\}$ & $0.1$ \\
        \hline
        \multirow{2}{*}{\textbf{4Forums}} 
            & \textbf{ADMM Step Length} & $\{10.0, 1.0, 0.1, 0.01\}$ & $1.0$ \\
            & \textbf{LCQP Regularization} & $\{100, 10, 1, 0.1, 0.01\}$ & $0.1$ \\
        \hline
            \multirow{3}{*}{\textbf{Epinions}} 
            & \textbf{GD Step Length} & $\{10.0, 1.0, 0.1, 0.01, 0.001\}$ & $0.01$ \\
            & \textbf{ADMM Step Length} & $\{10.0, 1.0, 0.1, 0.01\}$ & $0.1$ \\
            & \textbf{LCQP Regularization} & $\{100, 10, 1, 0.1, 0.01\}$ & $0.1$ \\
        \hline
            \multirow{3}{*}{\textbf{Citeseer}} 
            & \textbf{GD Step Length} & $\{10.0, 1.0, 0.1, 0.01, 0.001\}$ & $0.1$ \\
            & \textbf{ADMM Step Length} & $\{10.0, 1.0, 0.1, 0.01\}$ & $10.0$ \\
            & \textbf{LCQP Regularization} & $\{100, 10, 1, 0.1, 0.01\}$ & $10.0$ \\
        \hline
            \multirow{3}{*}{\textbf{Cora}} 
            & \textbf{GD Step Length} & $\{10.0, 1.0, 0.1, 0.01, 0.001\}$ & $0.1$ \\
            & \textbf{ADMM Step Length} & $\{10.0, 1.0, 0.1, 0.01\}$ & $10.0$ \\
            & \textbf{LCQP Regularization} & $\{100, 10, 1, 0.1, 0.01\}$ & $10.0$ \\
        \hline
            \multirow{2}{*}{\textbf{DDI}} 
            & \textbf{ADMM Step Length} & $\{10.0, 1.0, 0.1, 0.01\}$ & $1.0$ \\
            & \textbf{LCQP Regularization} & $\{100, 10, 1, 0.1, 0.01\}$ & $10.0$ \\
        \hline
            \multirow{3}{*}{\textbf{Yelp}} 
            & \textbf{GD Step Length} & $\{10.0, 1.0, 0.1, 0.01, 0.001\}$ & $0.001$ \\
            & \textbf{ADMM Step Length} & $\{10.0, 1.0, 0.1, 0.01\}$ & $1.0$ \\
            & \textbf{LCQP Regularization} & $\{100, 10, 1, 0.1, 0.01\}$ & $0.1$ \\
        \hline
            \multirow{2}{*}{\textbf{MNIST-Add1}} 
            & \textbf{ADMM Step Length} & $\{10.0, 1.0, 0.1, 0.01\}$ & $1.0$ \\
            & \textbf{LCQP Regularization} & $\{100, 10, 1, 0.1, 0.01, 0.001\}$ & $0.001$ \\
        \hline
            \multirow{2}{*}{\textbf{MNIST-Add2}} 
            & \textbf{ADMM Step Length} & $\{10.0, 1.0, 0.1, 0.01\}$ & $1.0$ \\
            & \textbf{LCQP Regularization} & $\{100, 10, 1, 0.1, 0.01, 0.001\}$ & $0.001$ \\
        \bottomrule
    \end{tabular}
    }
\end{table}

This section details the hyperparameter settings and search process for the inference runtime experiments in \secref{sec:inference_runtime}.
The GD, ADMM, and D-BCD algorithms are stopped when the $L_{\infty}$ norm of the primal variable change between iterates is less than $0.001$.
For the D-BCD algorithms, the regularization parameter from \appref{appendix:empirical_lcqp_regularization} resulting in the fastest runtime and yielding a prediction performance within a standard error of the best is used.
The default Gurobi optimizer hyperparameters are used.
\tabref{tab:inference_runtime_hyperparameters} reports the range of hyperparameters searched over and the final values.
Furthermore, for the MNIST-Add1 and MNIST-Add2 models, the highest performing trained neural models for each split from the performance experiments in \secref{sec:learning_performance} are used.

\tabref{tab:appendix_inference_time} reports the average and standard deviation of the inference runtime for Gurobi, GD, ADMM, and D-BCD algorithms on $4$ of the datasets from \tabref{tab:datasets}.
As in the main paper, we see the D-BCD algorithms are competitive with ADMM, the current state of the art optimizer for NeuPSL inference.
Moreover, here we see the LF D-BCD algorithm is also competitive with Gurobi for a single round of inference.

\subsection{Extended Learning Runtime}
\label{appendix:extended_learning_runtime}

This section provides details of the hyperparameter settings for the learning runtime experiments in \secref{sec:learning_runtime}.
For both learning losses, a negative log regularization with coefficient $1.0e-3$ on the symbolic weights is added to the learning loss as suggested by \citenoun{pryor:ijcai23}.
For ADMM inference on both learning losses, the same steplength from the inference runtime experiment is used for the first $7$ datasets in \tabref{tab:datasets}.
Similarly, for D-BCD inference on both learning losses, the same regularization parameter from the inference runtime experiment is used for the first $7$ datasets in \tabref{tab:datasets}.
For the MNIST-Add experiments, we use the regularization parameter $\epsilon = 1.0e-3$ and ADMM steplength $1.0$ as the values were found to achieve the highest final validation prediction performance.

Mirror descent is applied to learn the symbolic weights for both SP and MSE losses. 
The mirror descent steplength is set to a default value of $1.0e-3$ for the first $7$ datasets in \tabref{tab:datasets}.
For the MNIST-Add datasets the mirror descent steplength is set to $1.0e-14$ as in this problem we only need to learn the neural weights.
The Adam steplength for the neural component of the MNIST-Add models is set to a default value of $1.0e-3$.

Our learning framework, \algoref{alg:nesy_ebm_learning}, is used to fit the MSE learning loss.
We set the initial squared penalty parameter to a default value of $2.0$ for all datasets.
Moreover, for the first $7$ datasets in \tabref{tab:datasets} we set the Moreau parameter to $0.01$, the energy loss coefficient to $0.1$, and the steplength on the target variables $\mathbf{y}$ to $0.01$.
For the MNIST-Add datasets we set the Moreau parameter to $1.0e-3$, the energy loss coefficient to $10.0$, and the steplength on the target variables $\mathbf{y}$ to $1.0e-3$.

\subsection{Extended Learning Prediction Performance}
\label{appendix:extended_learning_performance}

This section details the hyperparameter settings and search process for the prediction performance experiments in \secref{sec:learning_performance}.
For all learning losses, a negative log regularization with coefficient $1.0e-3$ on the symbolic weights is added to the learning loss as suggested by \citenoun{pryor:ijcai23}.
The remaining hyperparameter search and setting details are described separately for the HL-MRF learning and deep HL-MRF learning experiments.

\begin{table}[ht]
    \centering
    \caption{Hyperparameter ranges and final values for the HL-MRF learning prediction performance experiments in \tabref{tab:hl_mrf_prediction_performance}.}
    \label{tab:hlmrf_prediction_performance_hyperparameters}
    \scalebox{0.5}{
    \begin{tabular}{c|c|c||c|c}
        \toprule
            \textbf{Dataset} & \textbf{Learning Loss} & \textbf{Parameter} & \textbf{Range} & \textbf{Final Value} \\
         \midrule
         \midrule
            \multirow{14}{*}{\textbf{CreateDebate}} & \multirow{2}{*}{\textbf{Energy}}
            & \textbf{Mirror Descent Step Length} & $\{1.0e-3, 1.0e-2\}$ & $1.0e-3$ \\
            & & \textbf{LCQP Regularization} & $\{1.0e-3, 1.0e-2\}$ & $1.0e-2$ \\
            \cline{2-5}
            & \multirow{2}{*}{\textbf{SP}}
            & \textbf{Mirror Descent Step Length} & $\{1.0e-3, 1.0e-2\}$ & $1.0e-2$ \\
            & & \textbf{LCQP Regularization} & $\{1.0e-3, 1.0e-2\}$ & $1.0e-3$ \\
            \cline{2-5}
            & \multirow{5}{*}{\textbf{MSE}}
            & \textbf{Mirror Descent Step Length} & $\{1.0e-3, 1.0e-2\}$ & $1.0e-2$ \\
            & & \textbf{$\mathbf{y}$ Step Length} & $\{1.0e-2, 1.0e-1\}$ & $1.0e-1$ \\
            & & \textbf{Moreau Parameter} & $\{1.0e-3, 1.0e-2, 1.0e-1\}$ & $1.0e-2$ \\
            & & \textbf{Energy Loss Coefficient} & $\{0, 1.0e-1, 1, 10 \}$ & $0.1$ \\
            & & \textbf{LCQP Regularization} & $\{1.0e-3, 1.0e-2\}$ & $1.0e-3$ \\
            \cline{2-5}
            & \multirow{5}{*}{\textbf{BCE}}
            & \textbf{Mirror Descent Step Length} & $\{1.0e-3, 1.0e-2\}$ & $1.0e-3$ \\
            & & \textbf{$\mathbf{y}$ Step Length} & $\{1.0e-2, 1.0e-1\}$ & $1.0e-1$ \\
            & & \textbf{Moreau Parameter} & $\{1.0e-3, 1.0e-2, 1.0e-1\}$ & $1.0e-2$ \\
            & & \textbf{Energy Loss Coefficient} & $\{0, 1.0e-1, 1, 10 \}$ & $10$ \\
            & & \textbf{LCQP Regularization} & $\{1.0e-3, 1.0e-2\}$ & $1.0e-2$ \\
        \hline
        \multirow{14}{*}{\textbf{4Forums}} & \multirow{2}{*}{\textbf{Energy}}
            & \textbf{Mirror Descent Step Length} & $\{1.0e-3, 1.0e-2\}$ & $1.0e-3$ \\
            & & \textbf{LCQP Regularization} & $\{1.0e-3, 1.0e-2\}$ & $1.0e-3$ \\
            \cline{2-5}
            & \multirow{2}{*}{\textbf{SP}}
            & \textbf{Mirror Descent Step Length} & $\{1.0e-3, 1.0e-2\}$ & $1.0e-3$ \\
            & & \textbf{LCQP Regularization} & $\{1.0e-3, 1.0e-2\}$ & $1.0e-3$ \\
            \cline{2-5} 
            & \multirow{5}{*}{\textbf{MSE}}
            & \textbf{Mirror Descent Step Length} & $\{1.0e-3, 1.0e-2\}$ & $1.0e-3$ \\
            & & \textbf{$\mathbf{y}$ Step Length} & $\{1.0e-2, 1.0e-1\}$ & $1.0e-2$ \\
            & & \textbf{Moreau Parameter} & $\{1.0e-3, 1.0e-2, 1.0e-1\}$ & $1.0e-3$ \\
            & & \textbf{Energy Loss Coefficient} & $\{0, 1.0e-1, 1, 10 \}$ & $0$ \\
            & & \textbf{LCQP Regularization} & $\{1.0e-3, 1.0e-2\}$ & $1.0e-3$ \\
            \cline{2-5}
            & \multirow{5}{*}{\textbf{BCE}}
            & \textbf{Mirror Descent Step Length} & $\{1.0e-3, 1.0e-2\}$ & $1.0e-3$ \\
            & & \textbf{$\mathbf{y}$ Step Length} & $\{1.0e-2, 1.0e-1\}$ & $1.0e-2$ \\
            & & \textbf{Moreau Parameter} & $\{1.0e-3, 1.0e-2, 1.0e-1\}$ & $1.0e-3$ \\
            & & \textbf{Energy Loss Coefficient} & $\{0, 1.0e-1, 1, 10 \}$ & $0$ \\
            & & \textbf{LCQP Regularization} & $\{1.0e-3, 1.0e-2\}$ & $1.0e-3$ \\
        \hline
            \multirow{14}{*}{\textbf{Epinions}} & \multirow{2}{*}{\textbf{Energy}}
            & \textbf{Mirror Descent Step Length} & $\{1.0e-3, 1.0e-2\}$ & $1.0e-3$ \\
            & & \textbf{LCQP Regularization} & $\{1.0e-3, 1.0e-2\}$ & $1.0e-3$ \\
            \cline{2-5}
            & \multirow{2}{*}{\textbf{SP}}
            & \textbf{Mirror Descent Step Length} & $\{1.0e-3, 1.0e-2\}$ & $1.0e-3$ \\
            & & \textbf{LCQP Regularization} & $\{1.0e-3, 1.0e-2\}$ & $1.0e-3$ \\
            \cline{2-5} 
            & \multirow{5}{*}{\textbf{MSE}}
            & \textbf{Mirror Descent Step Length} & $\{1.0e-3, 1.0e-2\}$ & $1.0e-2$ \\
            & & \textbf{$\mathbf{y}$ Step Length} & $\{1.0e-2, 1.0e-1\}$ & $1.0e-1$ \\
            & & \textbf{Moreau Parameter} & $\{1.0e-3, 1.0e-2, 1.0e-1\}$ & $1.0e-2$ \\
            & & \textbf{Energy Loss Coefficient} & $\{0, 1.0e-1, 1, 10 \}$ & $0.1$ \\
            & & \textbf{LCQP Regularization} & $\{1.0e-3, 1.0e-2\}$ & $1.0e-2$ \\
            \cline{2-5}
            & \multirow{5}{*}{\textbf{BCE}}
            & \textbf{Mirror Descent Step Length} & $\{1.0e-3, 1.0e-2\}$ & $1.0e-2$ \\
            & & \textbf{$\mathbf{y}$ Step Length} & $\{1.0e-2, 1.0e-1\}$ & $1.0e-1$ \\
            & & \textbf{Moreau Parameter} & $\{1.0e-3, 1.0e-2, 1.0e-1\}$ & $1.0e-2$ \\
            & & \textbf{Energy Loss Coefficient} & $\{0, 1.0e-1, 1, 10 \}$ & $1$ \\
            & & \textbf{LCQP Regularization} & $\{1.0e-3, 1.0e-2\}$ & $1.0e-2$ \\
        \hline
            \multirow{14}{*}{\textbf{Citeseer}} & \multirow{2}{*}{\textbf{Energy}}
            & \textbf{Mirror Descent Step Length} & $\{1.0e-3, 1.0e-2\}$ & $1.0e-3$ \\
            & & \textbf{LCQP Regularization} & $\{1.0e-3, 1.0e-2\}$ & $1.0e-2$ \\
            \cline{2-5}
            & \multirow{2}{*}{\textbf{SP}}
            & \textbf{Mirror Descent Step Length} & $\{1.0e-3, 1.0e-2\}$ & $1.0e-3$ \\
            & & \textbf{LCQP Regularization} & $\{1.0e-3, 1.0e-2\}$ & $1.0e-3$ \\
            \cline{2-5} 
            & \multirow{5}{*}{\textbf{MSE}}
            & \textbf{Mirror Descent Step Length} & $\{1.0e-3, 1.0e-2\}$ & $1.0e-3$ \\
            & & \textbf{$\mathbf{y}$ Step Length} & $\{1.0e-2, 1.0e-1\}$ & $1.0e-2$ \\
            & & \textbf{Moreau Parameter} & $\{1.0e-3, 1.0e-2, 1.0e-1\}$ & $1.0e-2$ \\
            & & \textbf{Energy Loss Coefficient} & $\{0, 1.0e-1, 1, 10 \}$ & $1$ \\
            & & \textbf{LCQP Regularization} & $\{1.0e-3, 1.0e-2\}$ & $1.0e-2$ \\
            & \multirow{5}{*}{\textbf{BCE}}
            & \textbf{Mirror Descent Step Length} & $\{1.0e-3, 1.0e-2\}$ & $1.0e-2$ \\
            & & \textbf{$\mathbf{y}$ Step Length} & $\{1.0e-2, 1.0e-1\}$ & $1.0e-1$ \\
            & & \textbf{Moreau Parameter} & $\{1.0e-3, 1.0e-2, 1.0e-1\}$ & $1.0e-3$ \\
            & & \textbf{Energy Loss Coefficient} & $\{0, 1.0e-1, 1, 10 \}$ & $0$ \\
            & & \textbf{LCQP Regularization} & $\{1.0e-3, 1.0e-2\}$ & $1.0e-3$ \\
        \hline
            \multirow{14}{*}{\textbf{Cora}} & \multirow{2}{*}{\textbf{Energy}}
            & \textbf{Mirror Descent Step Length} & $\{1.0e-3, 1.0e-2\}$ & $1.0e-3$ \\
            & & \textbf{LCQP Regularization} & $\{1.0e-3, 1.0e-2\}$ & $1.0e-2$ \\
            \cline{2-5}
            & \multirow{2}{*}{\textbf{SP}}
            & \textbf{Mirror Descent Step Length} & $\{1.0e-3, 1.0e-2\}$ & $1.0e-3$ \\
            & & \textbf{LCQP Regularization} & $\{1.0e-3, 1.0e-2\}$ & $1.0e-3$ \\
            \cline{2-5} 
            & \multirow{5}{*}{\textbf{MSE}}
            & \textbf{Mirror Descent Step Length} & $\{1.0e-3, 1.0e-2\}$ & $1.0e-2$ \\
            & & \textbf{$\mathbf{y}$ Step Length} & $\{1.0e-2, 1.0e-1\}$ & $1.0e-1$ \\
            & & \textbf{Moreau Parameter} & $\{1.0e-3, 1.0e-2, 1.0e-1\}$ & $1.0e-2$ \\
            & & \textbf{Energy Loss Coefficient} & $\{0, 1.0e-1, 1, 10 \}$ & $0.1$ \\
            & & \textbf{LCQP Regularization} & $\{1.0e-3, 1.0e-2\}$ & $1.0e-3$ \\
            & \multirow{5}{*}{\textbf{BCE}}
            & \textbf{Mirror Descent Step Length} & $\{1.0e-3, 1.0e-2\}$ & $1.0e-2$ \\
            & & \textbf{$\mathbf{y}$ Step Length} & $\{1.0e-2, 1.0e-1\}$ & $1.0e-1$ \\
            & & \textbf{Moreau Parameter} & $\{1.0e-3, 1.0e-2, 1.0e-1\}$ & $1.0e-2$ \\
            & & \textbf{Energy Loss Coefficient} & $\{0, 1.0e-1, 1, 10 \}$ & $0.1$ \\
            & & \textbf{LCQP Regularization} & $\{1.0e-3, 1.0e-2\}$ & $1.0e-3$ \\
        \hline
            \multirow{14}{*}{\textbf{DDI}} & \multirow{2}{*}{\textbf{Energy}}
            & \textbf{Mirror Descent Step Length} & $\{1.0e-3, 1.0e-2\}$ & $1.0e-3$ \\
            & & \textbf{LCQP Regularization} & $\{1.0e-3, 1.0e-2\}$ & $1.0e-2$ \\
            \cline{2-5}
            & \multirow{2}{*}{\textbf{SP}}
            & \textbf{Mirror Descent Step Length} & $\{1.0e-3, 1.0e-2\}$ & $1.0e-3$ \\
            & & \textbf{LCQP Regularization} & $\{1.0e-3, 1.0e-2\}$ & $1.0e-2$ \\
            \cline{2-5} 
            & \multirow{5}{*}{\textbf{MSE}}
            & \textbf{Mirror Descent Step Length} & $\{1.0e-3, 1.0e-2\}$ & $1.0e-3$ \\
            & & \textbf{$\mathbf{y}$ Step Length} & $\{1.0e-2, 1.0e-1\}$ & $1.0e-1$ \\
            & & \textbf{Moreau Parameter} & $\{1.0e-3, 1.0e-2, 1.0e-1\}$ & $1.0e-3$ \\
            & & \textbf{Energy Loss Coefficient} & $\{0, 1.0e-1, 1, 10 \}$ & $0.1$ \\
            & & \textbf{LCQP Regularization} & $\{1.0e-3, 1.0e-2\}$ & $1.0e-2$ \\
            & \multirow{5}{*}{\textbf{BCE}}
            & \textbf{Mirror Descent Step Length} & $\{1.0e-3, 1.0e-2\}$ & $1.0e-2$ \\
            & & \textbf{$\mathbf{y}$ Step Length} & $\{1.0e-2, 1.0e-1\}$ & $1.0e-2$ \\
            & & \textbf{Moreau Parameter} & $\{1.0e-3, 1.0e-2, 1.0e-1\}$ & $1.0e-2$ \\
            & & \textbf{Energy Loss Coefficient} & $\{0, 1.0e-1, 1, 10 \}$ & $0.1$ \\
            & & \textbf{LCQP Regularization} & $\{1.0e-3, 1.0e-2\}$ & $1.0e-2$ \\
        \hline
            \multirow{14}{*}{\textbf{Yelp}} & \multirow{2}{*}{\textbf{Energy}}
            & \textbf{Mirror Descent Step Length} & $\{1.0e-3, 1.0e-2\}$ & $1.0e-3$ \\
            & & \textbf{LCQP Regularization} & $\{1.0e-3, 1.0e-2\}$ & $1.0e-2$ \\
            \cline{2-5}
            & \multirow{2}{*}{\textbf{SP}}
            & \textbf{Mirror Descent Step Length} & $\{1.0e-3, 1.0e-2\}$ & $1.0e-3$ \\
            & & \textbf{LCQP Regularization} & $\{1.0e-3, 1.0e-2\}$ & $1.0e-2$ \\
            \cline{2-5} 
            & \multirow{5}{*}{\textbf{MSE}}
            & \textbf{Mirror Descent Step Length} & $\{1.0e-3, 1.0e-2\}$ & $1.0e-3$ \\
            & & \textbf{$\mathbf{y}$ Step Length} & $\{1.0e-2, 1.0e-1\}$ & $1.0e-2$ \\
            & & \textbf{Moreau Parameter} & $\{1.0e-3, 1.0e-2, 1.0e-1\}$ & $1.0e-1$ \\
            & & \textbf{Energy Loss Coefficient} & $\{0, 1.0e-1, 1, 10 \}$ & $10$ \\
            & & \textbf{LCQP Regularization} & $\{1.0e-3, 1.0e-2\}$ & $1.0e-2$ \\
            & \multirow{5}{*}{\textbf{BCE}}
            & \textbf{Mirror Descent Step Length} & $\{1.0e-3, 1.0e-2\}$ & $1.0e-3$ \\
            & & \textbf{$\mathbf{y}$ Step Length} & $\{1.0e-2, 1.0e-1\}$ & $1.0e-2$ \\
            & & \textbf{Moreau Parameter} & $\{1.0e-3, 1.0e-2, 1.0e-1\}$ & $1.0e-2$ \\
            & & \textbf{Energy Loss Coefficient} & $\{0, 1.0e-1, 1, 10 \}$ & $0.1$ \\
            & & \textbf{LCQP Regularization} & $\{1.0e-3, 1.0e-2\}$ & $1.0e-2$ \\
        \bottomrule
    \end{tabular}
    }
\end{table}

\textbf{HL-MRF Learning}
The LF D-BCD algorithm is used for inference in all experiments.
Moreover, the D-BCD algorithm is stopped when the primal-dual gap drops below $\delta = 1.0e-2$ CreateDebate, 4Forums, Epinions, Citeseer, Cora, and DDI while the primal-dual threshold is set to $\delta = 1.0e-1$ to adjust to the larger scale of the dataset.
For all learning losses, the learning algorithm is stopped when the training evaluation metric stops improving after $50$ epochs.
For the MSE and BCE losses trained with \algoref{alg:nesy_ebm_learning}, the final objective difference tolerance was set to $0.1$ for the smaller CreateDebate, 4Forums, and Epinions datasets and $1$ for Citeseer, Cora, DDI, and Yelp.
Moreover, the initial squared penalty coefficient is set to $2$ for all datasets.
The remaining hyperparameters are searched over the ranges specified in \tabref{tab:hlmrf_prediction_performance_hyperparameters}.
The hyperparameter value with the best performance metric on the first fold is selected.

\begin{table}[ht]
    \centering
    \caption{Hyperparameter ranges and final values for the deep HL-MRF learning prediction performance experiments on Citeseer and Cora.}
    \label{tab:deep_hl-mrf_citation_learning_hyperparameters}
    \scalebox{0.7}{
    \begin{tabular}{c|c|c||c|c}
        \toprule
            \textbf{Dataset} & \textbf{Loss} & \textbf{Parameter} & \textbf{Range} & \textbf{Final Value} \\
        \midrule
        \midrule
            \multirow{14}{*}{\textbf{Citeseer}} & \multirow{2}{*}{\textbf{Energy}} 
            & \textbf{Mirror Descent Step Length} & $\{1.0e-3, 1.0e-2\}$ & $1.0e-3$ \\
            & & \textbf{LCQP Regularization} & $\{1.0e-3\}$ & $1.0e-3$ \\
        \cline{2-5}
            & \multirow{2}{*}{\textbf{SP}} 
            & \textbf{Mirror Descent Step Length} & $\{1.0e-3, 1.0e-2\}$ & $1.0e-2$ \\
            & & \textbf{LCQP Regularization} & $\{1.0e-3\}$ & $1.0e-3$ \\
        \cline{2-5}
            & \multirow{5}{*}{\textbf{MSE}} 
            & \textbf{Mirror Descent Step Length} & $\{1.0e-3, 1.0e-2\}$ & $1.0e-2$ \\
            & & \textbf{$\mathbf{y}$ Step Length} & $\{1.0e-3, 1.0e-2\}$ & $1.0e-2$ \\
            & & \textbf{Moreau Parameter} & $\{1.0e-3, 1.0e-2, 1.0e-1\}$ & $1.0e-1$ \\
            & & \textbf{Energy Loss Coefficient} & $\{1.0e-1, 1, 10 \}$ & $1.0e-1$ \\
            & & \textbf{LCQP Regularization} & $\{1.0e-3\}$ & $1.0e-3$ \\
        \cline{2-5}
            & \multirow{5}{*}{\textbf{BCE}} 
            & \textbf{Mirror Descent Step Length} & $\{1.0e-3, 1.0e-2\}$ & $1.0e-2$ \\
            & & \textbf{$\mathbf{y}$ Step Length} & $\{1.0e-3, 1.0e-2\}$ & $1.0e-3$ \\
            & & \textbf{Moreau Parameter} & $\{1.0e-3, 1.0e-2, 1.0e-1\}$ & $1.0e-1$ \\
            & & \textbf{Energy Loss Coefficient} & $\{1.0e-1, 1, 10 \}$ & $1.0e-1$ \\
            & & \textbf{LCQP Regularization} & $\{1.0e-3\}$ & $1.0e-3$ \\
        \midrule
            \multirow{14}{*}{\textbf{Cora}} & \multirow{2}{*}{\textbf{Energy}} 
            & \textbf{Mirror Descent Step Length} & $\{1.0e-3, 1.0e-2\}$ & $1.0e-2$ \\
            & & \textbf{LCQP Regularization} & $\{1.0e-3\}$ & $1.0e-3$ \\
        \cline{2-5}
            & \multirow{2}{*}{\textbf{SP}} 
            & \textbf{Mirror Descent Step Length} & $\{1.0e-3, 1.0e-2\}$ & $1.0e-2$ \\
            & & \textbf{LCQP Regularization} & $\{1.0e-3\}$ & $1.0e-3$ \\
        \cline{2-5}
            & \multirow{5}{*}{\textbf{MSE}} 
            & \textbf{Mirror Descent Step Length} & $\{1.0e-3, 1.0e-2\}$ & $1.0e-2$ \\
            & & \textbf{$\mathbf{y}$ Step Length} & $\{1.0e-3, 1.0e-2\}$ & $1.0e-3$ \\
            & & \textbf{Moreau Parameter} & $\{1.0e-3, 1.0e-2, 1.0e-1\}$ & $1.0e-1$ \\
            & & \textbf{Energy Loss Coefficient} & $\{1.0e-1, 1, 10 \}$ & $1$ \\
            & & \textbf{LCQP Regularization} & $\{1.0e-3\}$ & $1.0e-3$ \\
        \cline{2-5}
            & \multirow{5}{*}{\textbf{BCE}} 
            & \textbf{Mirror Descent Step Length} & $\{1.0e-3, 1.0e-2\}$ & $1.0e-2$ \\
            & & \textbf{$\mathbf{y}$ Step Length} & $\{1.0e-3, 1.0e-2\}$ & $1.0e-2$ \\
            & & \textbf{Moreau Parameter} & $\{1.0e-3, 1.0e-2, 1.0e-1\}$ & $1.0e-1$ \\
            & & \textbf{Energy Loss Coefficient} & $\{1.0e-1, 1, 10 \}$ & $1.0e-1$ \\
            & & \textbf{LCQP Regularization} & $\{1.0e-3\}$ & $1.0e-3$ \\
        \bottomrule
    \end{tabular}
    }
\end{table}

\begin{table}[ht]
    \centering
    \caption{Hyperparameter ranges and final values for the deep HL-MRF learning prediction performance experiments on MNIST-Add datasets.}
    \label{tab:deep_hl-mrf_mnist_add_learning_hyperparameters}
    \scalebox{0.7}{
    \begin{tabular}{c|c|c||c|c}
        \toprule
            \textbf{Dataset} & \textbf{Loss} & \textbf{Parameter} & \textbf{Range} & \textbf{Final Value} \\
        \midrule
        \midrule
            \multirow{7}{*}{\textbf{MNIST-Add1}} & \multirow{2}{*}{\textbf{Energy}} 
            & \textbf{Mirror Descent Step Length} & $\{1.0e-14\}$ & $1.0e-14$ \\
            & & \textbf{Adam Step Length} & $\{1.0e-4, 1.0e-3\}$ & $1.0e-3$ \\
            & & \textbf{LCQP Regularization} & $\{1.0e-3\}$ & $1.0e-3$ \\
        \cline{2-5}
            & \multirow{5}{*}{\textbf{BCE}} 
            & \textbf{Mirror Descent Step Length} & $\{1.0e-14\}$ & $1.0e-14$ \\
            & & \textbf{Adam Step Length} & $\{1.0e-4, 1.0e-3\}$ & $1.0e-3$ \\
            & & \textbf{$\mathbf{y}$ Step Length} & $\{1.0e-3, 1.0e-2\}$ & $1.0e-3$ \\
            & & \textbf{Moreau Parameter} & $\{1.0e-3, 1.0e-2, 1.0e-1\}$ & $1.0e-2$ \\
            & & \textbf{Energy Loss Coefficient} & $\{1.0e-1, 1, 10 \}$ & $10$ \\
            & & \textbf{LCQP Regularization} & $\{1.0e-3\}$ & $1.0e-3$ \\
        \midrule
            \multirow{7}{*}{\textbf{MNIST-Add2}} & \multirow{2}{*}{\textbf{Energy}} 
            & \textbf{Mirror Descent Step Length} & $\{1.0e-14\}$ & $1.0e-14$ \\
            & & \textbf{Adam Step Length} & $\{1.0e-4, 1.0e-3\}$ & $1.0e-3$ \\
            & & \textbf{LCQP Regularization} & $\{1.0e-3\}$ & $1.0e-3$ \\
        \cline{2-5}
            & \multirow{5}{*}{\textbf{BCE}} 
            & \textbf{Mirror Descent Step Length} & $\{1.0e-14\}$ & $1.0e-14$ \\
            & & \textbf{Adam Step Length} & $\{1.0e-4, 1.0e-3\}$ & $1.0e-4$ \\
            & & \textbf{$\mathbf{y}$ Step Length} & $\{1.0e-3, 1.0e-2\}$ & $1.0e-3$ \\
            & & \textbf{Moreau Parameter} & $\{1.0e-3, 1.0e-2, 1.0e-1\}$ & $1.0e-3$ \\
            & & \textbf{Energy Loss Coefficient} & $\{1.0e-1, 1, 10 \}$ & $10$ \\
            & & \textbf{LCQP Regularization} & $\{1.0e-3\}$ & $1.0e-3$ \\
        \bottomrule
    \end{tabular}
    }
\end{table}

\textbf{Deep HL-MRF Learning}

For deep HL-MRF learning in the citation network and MNIST-Add evaluations reported in \tabref{tab:deep_hl_mrf_citation_prediction_performance} and \tabref{tab:deep_hl_mrf_mnist_add_prediction_performance}, respectively the validation set is used to determine when to stop the learning algorithms and what weights to use for final evaluations.
Specifically, after every learning step the model performance is measured on the validation data, and when $50$ consecutive steps finish without improvement, the learning algorithm is stopped.
For citation network datasets, the model obtaining the best validation metric averaged across all splits are used for final test evaluation.
For MNIST-Add datasets, the model obtaining the best validation metric on the first split is used for final test evaluation across all splits.
\tabref{tab:deep_hl-mrf_citation_learning_hyperparameters} and \tabref{tab:deep_hl-mrf_mnist_add_learning_hyperparameters} report the range of hyperparameters searched over and the final values resulting in the highest validation prediction performance for citation network datasets and MNIST-Add datasets, respectively.

\end{document}